\documentclass[lettersize,journal]{IEEEtran}
\usepackage{amsmath,amsfonts}
\usepackage{algorithmic}
\usepackage{algorithm}
\usepackage{array}
\usepackage{textcomp}
\usepackage{stfloats}
\usepackage{url}
\usepackage{verbatim}
\usepackage{cite}
\usepackage{blindtext}
\usepackage{multirow}
\usepackage{multicol}
\usepackage{graphicx}
\usepackage{epstopdf}
\usepackage{rotating}
\usepackage{makecell}

\usepackage{booktabs}
\usepackage{siunitx} 
\usepackage{mathtools}
\usepackage{subcaption}  
\usepackage{float}
\usepackage{tabularx}
\usepackage{tikz}       
\usepackage{forest}     
\usepackage[colorlinks=true,linkcolor=blue,citecolor=blue,urlcolor=blue]{hyperref}
\usepackage{orcidlink}

\usepackage{tabularx,booktabs,makecell,pifont,xcolor}

\newcolumntype{C}{>{\centering\arraybackslash}X}          
\renewcommand{\arraystretch}{1.2}                          
\newcolumntype{L}[1]{>{\raggedright\arraybackslash}p{#1}} 

\definecolor{armblue}{HTML}{1F77B4}
\definecolor{riscorange}{HTML}{FF7F0E}
\definecolor{hybridpurple}{HTML}{9467BD}
\newcommand{\Yes}{\raisebox{-0.1em}{\scalebox{1.8}{\textcolor{black}{\ding{51}}}}}
\newcommand{\Partial}{\raisebox{-0.1em}{\scalebox{1.8}{\textcolor{brown}{\ding{51}}}}}
\newcommand{\Arm}{\raisebox{-0.1em}{\scalebox{1.8}{\textcolor{armblue}{\ding{51}}}}}
\newcommand{\Risc}{\raisebox{-0.1em}{\scalebox{1.8}{\textcolor{riscorange}{\ding{51}}}}}
\newcommand{\Hybrid}{\raisebox{-0.1em}{\scalebox{1.8}{\textcolor{hybridpurple}{\ding{51}}}}}
\begin{document}

\title{Neural Network Quantization for Microcontrollers: A Comprehensive Survey of Methods, Platforms, and Applications}

\author{
    Hamza A. Abushahla\,\orcidlink{0009-0003-6247-6776}, \textit{Graduate Student Member, IEEE},  
    Dara Varam\,\orcidlink{0009-0004-7149-4842}, \textit{Graduate Student Member, IEEE},
    Ariel Justine N. Panopio\,\orcidlink{0009-0004-7149-4842}, \textit{Graduate Student Member, IEEE},
    and Mohamed I. AlHajri\,\orcidlink{0000-0003-1480-8054}, \textit{Senior Member, IEEE}%
    \thanks{
    \textit{(Corresponding Author: Mohamed I. AlHajri).}
    The authors are with the Department of Computer Science and Engineering, American University of Sharjah, Sharjah 26666, UAE (e-mail: b00090279@aus.edu; b00081313@aus.edu; b00088568@aus.edu; mialhajri@aus.edu).
    }
}


\maketitle

\begin{abstract}
The deployment of Quantized Neural Networks (QNNs) on resource-constrained edge devices, such as microcontrollers (MCUs), introduces fundamental challenges in balancing model performance, computational complexity, and memory constraints. Tiny Machine Learning (TinyML) addresses these issues by jointly advancing machine learning algorithms, hardware architectures, and software optimization techniques to enable deep neural network inference on embedded systems. This survey provides a hardware-oriented perspective on neural network quantization, systematically reviewing the quantization methods most relevant to MCUs and extreme-edge devices. Particular emphasis is placed on the critical trade-offs between model performance and the capabilities of MCU-class hardware, including memory hierarchies, numerical representations, and accelerator support. The survey further reviews contemporary MCU hardware platforms, including ARM-based and RISC-V-based designs, as well as MCUs integrating neural processing units (NPUs) for low-precision inference, together with the supporting software stacks. In addition, we analyze real-world deployments of quantized models on MCUs and consolidate the application domains in which such systems are used. Finally, we discuss open challenges and outline promising future directions toward scalable, energy-efficient, and sustainable AI deployment on edge devices.

\end{abstract}

\begin{IEEEkeywords}
Embedded Systems, Microcontrollers (MCUs), Neural Processing Units (NPUs), Edge Computing, Quantization, Reduced Precision, Deep Learning, TinyML, Sustainable AI
\end{IEEEkeywords}

\section{Introduction}

\IEEEPARstart{D}{eep} Neural Networks (DNNs) have achieved state-of-the-art performance across various domains including image classification \cite{he2016deep, lu2007survey, cai2017deep}, speech recognition \cite{nassif2019speech}, and object detection \cite{zaidi2022survey}. However, this exceptional performance comes at the cost of significantly increased computational and memory requirements \cite{gholami2022survey}. Modern DNN architectures often comprise millions, or even billions of parameters, necessitating billions of multiply-and-accumulate (MAC) operations during inference. This substantial computational workload imposes significant demands on both processing power and memory resources.

In parallel, the rapid growth of the Internet of Things (IoT) has significantly increased interest in resource-constrained embedded devices, particularly microcontroller units (MCUs). These MCUs are popular due to their low power requirements, high reliability, and ease of integration into diverse applications \cite{novac2021quantization}. Their integration with sensors, actuators, and networking capabilities enables sophisticated perception, intervention, and distributed intelligence on the edge.

Deploying complex DNNs on MCUs poses considerable challenges. Unlike graphics processing units (GPUs) or application-specific integrated circuit (ASIC) accelerators, MCUs typically operate within strict constraints: limited on-chip memory (often just hundreds of kilobytes), minimal computational resources, low clock speeds (typically 40 to 400 MHz), and stringent power budgets (in the order of milliwatts) \cite{capogrosso2024machine}. Furthermore, real-time responses must often be guaranteed \cite{lin2020mcunet}, particularly in safety-critical applications such as healthcare devices \cite{alshehri2020comprehensive}, autonomous vehicles \cite{yurtsever2020survey}, or industrial robotics \cite{soori2023artificial}. In such scenarios, inference latency is critical, as delayed decisions can lead to severe consequences such as compromised safety, system failures, or worse.

This has led to the emergence of the tiny machine learning (TinyML) paradigm, which enables machine learning (ML) models to run on edge devices with very limited compute and memory resources, typically consuming only a few milliwatts or less \cite{warden2019tinyml}. TinyML systems usually have memory footprints below 1 MB, often in the 64–256 KB range \cite{capogrosso2024machine}.

To meet the stringent resource constraints of these devices, model compression techniques have become essential for enabling DNN inference on such platforms. These methods aim to reduce parameter size, computational complexity, and memory consumption, thereby making neural networks deployable on MCUs. However, compression often introduces trade-offs, potentially degrading model performance, and thus finding the right balance between efficiency and accuracy remains a key challenge. Various strategies, including network pruning \cite{han2016deep}, low-rank decomposition \cite{phan2020stable}, lightweight network design, neural architecture search (NAS) \cite{ren2021comprehensive}, and knowledge distillation \cite{gou2021knowledge}, have been explored. One such method, which is the focus of this survey, is network quantization, where network architectures are preserved whilst the numerical representations of the parameters are replaced with low-memory alternatives.

Quantization leverages inherent redundancy within neural networks, enabling models to retain high accuracy despite reduced precision \cite{zhou2023towards}. Consequently, it has emerged as a key enabler of TinyML, and facilitating efficient inference acceleration widely adopted in industry. At scale, these efficiency gains translate into lower energy consumption and potentially reduced operational carbon footprint at the system level. This survey reviews state-of-the-art quantization techniques for efficient DNN deployment on MCUs, bridging quantization methods, hardware architectures, and software frameworks, providing a holistic guide for researchers and practitioners.

\subsection{Existing Surveys}
\begin{table*}[t]
\centering
\caption{
Scope comparison of surveys on quantization, TinyML, and edge AI. 
Ticks indicate {\scriptsize\Yes}\,surveyed and {\scriptsize\Partial}\,partially surveyed. 
Hardware ticks indicate {\scriptsize\Arm}\,ARM, {\scriptsize\Risc}\,RISC, and {\scriptsize\Hybrid}\,hybrid architectures.
}
\label{tab:survey-scope}
\scriptsize
\begin{tabularx}{\textwidth}{L{0.17\textwidth} c *{6}{C}}
\toprule
\textbf{Paper} & \textbf{Year} &
\textbf{Primary Quantization} &
\textbf{Advanced Quantization} &
\textbf{Numeric Representations} &
\textbf{Hardware Landscape} &
\textbf{Software Frameworks} &
\textbf{Applications} \\
\midrule
\makecell[l]{Gholami et al. \cite{gholami2022survey}} & 2022 & \Yes & \Yes &  & \Arm\ \Risc &  &  \\
\makecell[l]{Orășan et al. \cite{lucan2022brief}}    & 2022 &  &  &  & \Arm & \Yes & \Yes \\
\makecell[l]{Giordano et al. \cite{giordano2022survey}} & 2022 &  &  &  & \Arm\ \Risc\ \Hybrid &  &  \\
\makecell[l]{Saha et al. \cite{saha2022}}            & 2022 & \Yes & \Partial  &  &  & \Yes & \Yes \\
\makecell[l]{Ray \cite{ray2022review}}        & 2022 & \Yes & \Partial  &  & \Arm\ \Risc & \Yes & \Yes \\
\makecell[l]{Rokh et al. \cite{rokh2023comprehensive}} & 2023 & \Yes & \Yes &  &  &  &  \\
\makecell[l]{Akkad et al. \cite{akkad2023embedded}}  & 2023 &  &  &  & \Risc &  & \Yes \\
\makecell[l]{Liu et al. \cite{liu2024lightweight}}   & 2024 & \Yes & \Yes &  &  & \Yes &  \\
\makecell[l]{Liu et al. \cite{liu2024exploring}}  & 2024 &  &  &  & \Risc &  &  \\
\makecell[l]{Capogrosso et al. \cite{capogrosso2024machine}} & 2024 & \Yes &  &  & \Arm\ \Risc & \Yes &  \\
\makecell[l]{Liu et al. \cite{liu2025low}}           & 2025 & \Yes & \Yes & \Yes &  &  &  \\
\makecell[l]{Heydari and Mahmoud \cite{heydari2025tiny}}  & 2025 &  &  &  &  &  & \Yes \\
\makecell[l]{Wang and Jia \cite{wang2025optimizing}}  & 2025 & \Yes & \Partial &  &  & \Yes & \Yes \\
\textbf{Ours}                                        & 2025 & \Yes & \Yes & \Yes & \Arm\ \Risc\ \Hybrid & \Yes & \Yes \\
\bottomrule
\end{tabularx}
\end{table*}

Numerous recent surveys have addressed quantization, model compression, or TinyML deployment independently, yet each remains limited in scope. Our work addresses these gaps by taking a comprehensive and hardware-motivated perspective, focusing on the deployment of quantized neural networks (QNNs) on MCUs.

Several surveys, such as by Liu et al.~\cite{liu2025low} and Gholami et al.~\cite{gholami2022survey}, focus exclusively on software-level or algorithmic advancements in quantization. In their work, a comprehensive review of low-bit quantization algorithms and training techniques is provided across a variety of applications and domains. Other works include Rokh et al.~\cite{rokh2023comprehensive} evaluating quantization methods for image classification and Liu et al.~\cite{liu2024lightweight} presenting lightweight model design and hardware acceleration. Although extensive, these surveys give little attention to hardware constraints or concerns for practical deployment, not targeting MCUs or similar edge-based devices.

Conversely, hardware-focused surveys tend to omit model optimization strategies. Orășan et al. \cite{lucan2022brief} examine ARM Cortex-M based MCUs with DNN deployment examples, but without exploring quantization. Similarly, Liu et al. \cite{liu2024exploring}, Akkad et al. \cite{akkad2023embedded}, and Giordano et al. \cite{giordano2022survey} analyze RISC-V and hybrid platforms but provide limited discussion on model compression.

Broader TinyML surveys provide ecosystem-level overviews but lack technical depth in quantization. For instance, Saha et al.~\cite{saha2022} cover MCU hardware and deployment applications, while Heydari and Mahmoud \cite{heydari2025tiny} and Ray \cite{ray2022review} emphasize general domain-level use cases rather than quantization techniques.

Several studies attempt to bridge both the hardware and software aspects of the problem. For example, Capogrosso et al. \cite{capogrosso2024machine} survey model optimization and deployment frameworks, yet omit the discussion of more advanced quantization techniques. Wang and Jia \cite{wang2025optimizing} explore the entire landscape of working on-the-edge from data collection to hardware deployment. However, similar to \cite{capogrosso2024machine}, the work lacks focus on modern quantization strategies and specific MCU implementation. 

In contrast, our survey not only bridges quantization techniques, microcontroller hardware (including ARM, RISC-V, and hybrid NPUs), and software stacks for optimized inference, but also addresses a critical yet often overlooked hardware dimension: \textit{numerical representations}. Furthermore, we provide a consolidated review of diverse application domains in which different quantization methods and hardware platforms have been deployed. This integrated perspective fills a key gap in existing literature by delivering practical, actionable insights for deploying QNNs on embedded platforms. Table~\ref{tab:survey-scope} presents a comparison of our survey with the most relevant prior works on quantization, TinyML, and edge AI, covering quantization approaches (primary and advanced), numerical representations, hardware platforms, software frameworks, and applications.

\subsection{Contributions}
This survey makes the following key contributions:

\begin{enumerate} 
\item We provide a hardware-oriented perspective on neural network quantization, covering its motivations, core formulations, and practical design considerations, specifically in the context of MCU-class devices.

\item We present a structured taxonomy of state-of-the-art quantization methods, categorizing them based on key characteristics such as training approach, numerical precision and representation, and deployment constraints.

\item We survey recent advancements in hardware platforms, including ARM-based, RISC-V-based, and hybrid/NPU-enabled MCUs, and examine the supporting software frameworks, runtime libraries, and and deployment toolchains for low-precision inference.

\item We analyze real-world applications and use cases of QNN deployment on MCUs, offering a comparative discussion of quantization methods, deployment strategies, and platform suitability across various scenarios.

\item We identify key challenges in quantized model deployment on embedded hardware, such as accuracy degradation, mixed precision execution, and memory bottlenecks, and discuss emerging techniques aimed at mitigating these limitations.

\item We outline promising directions for future research, including distribution-aware training, quantization-friendly architecture design, full-stack co-optimization of hardware, software, and the treatment of sustainability as a system-level objective for energy-efficient and long-lived edge intelligence.
\end{enumerate}

\begin{figure*}[ht]
    \includegraphics[width=1.0\linewidth]{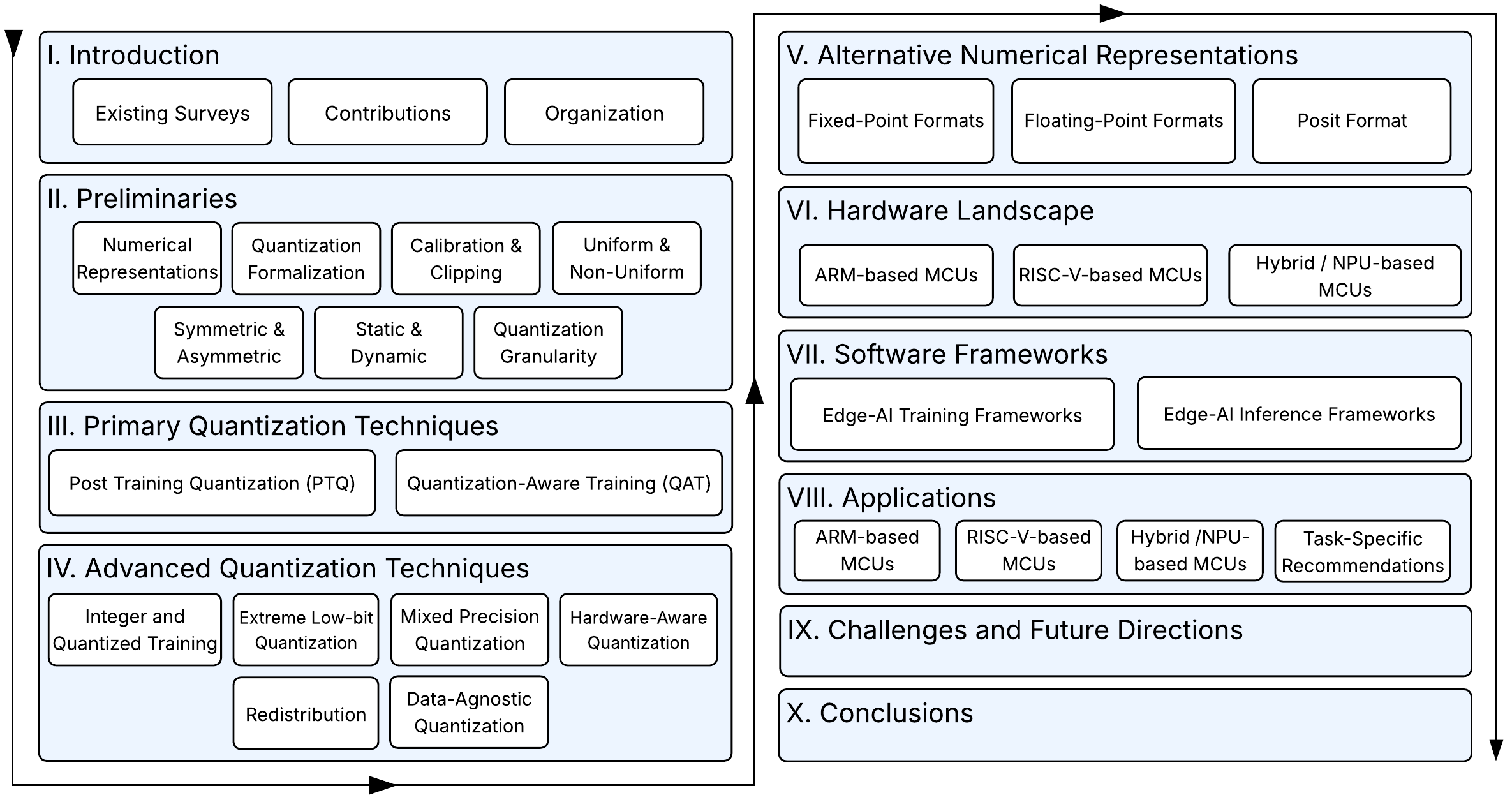}
    \caption{The taxonomy of the discussed topics in this survey.}
    \label{fig:taxonomy}
\end{figure*}

\subsection{Organization}

The rest of this survey is organized as follows: Section~\ref{sec:preliminaries} outlines the necessary background on numerical formats and quantization fundamentals. Section~\ref{sec:quant_techniques} reviews primary quantization methods. Section~\ref{sec:advanced} discusses advanced quantization techniques.
Section~\ref{sec:alternative_formats} explores alternative numerical representations.
Section~\ref{sec:hardware} surveys the microcontroller hardware landscape. Section~\ref{sec:software} reviews software stacks and deployment tools. Section~\ref{sec:applications} presents a comparative analysis of quantized deployment in real-world applications. Section~\ref{sec:challenges} highlights challenges and future directions. Finally, Section~\ref{sec:conclusion} concludes the survey. An overview of the taxonomy and structure of the topics discussed is shown in Fig.~\ref{fig:taxonomy}.

\section{Preliminaries}
\label{sec:preliminaries}

\subsection{Numerical Representations}
\label{sec:representations}

In digital hardware systems, numerical values must be represented using a finite set of bits. Since real numbers have infinite precision, their representation in hardware is necessarily an approximation. The process of encoding real numbers into a finite binary format is referred to as \textit{numerical representation}~\cite{gohil2021fixed, alsuhli2023number}, and it directly impacts the precision, memory requirements, and computational efficiency of DNNs.

Neural networks are essentially a set of parameters, such as weights, biases, and activations, which are stored in computer memory and used in arithmetic operations. Hence, the numerical representation chosen for these values affects not only the model’s precision and storage footprint, but also the speed and hardware cost of computation. From a hardware perspective, it influences both the silicon area required for arithmetic units and the energy consumed during computation.

In broad terms, there exist two classes of numerical representations that are commonly used in DNN systems: (1) floating-point formats and (2) fixed-point formats. Floating-point formats are standard in general-purpose computing platforms like CPUs and GPUs, especially during training, where higher precision is needed. Fixed-point formats, on the other hand, are typically used in embedded systems such as MCUs and digital signal processors, where energy efficiency, silicon area, and memory constraints are critical. 

\begin{figure}[htbp]
    \centering
    \includegraphics[width=0.45\textwidth]{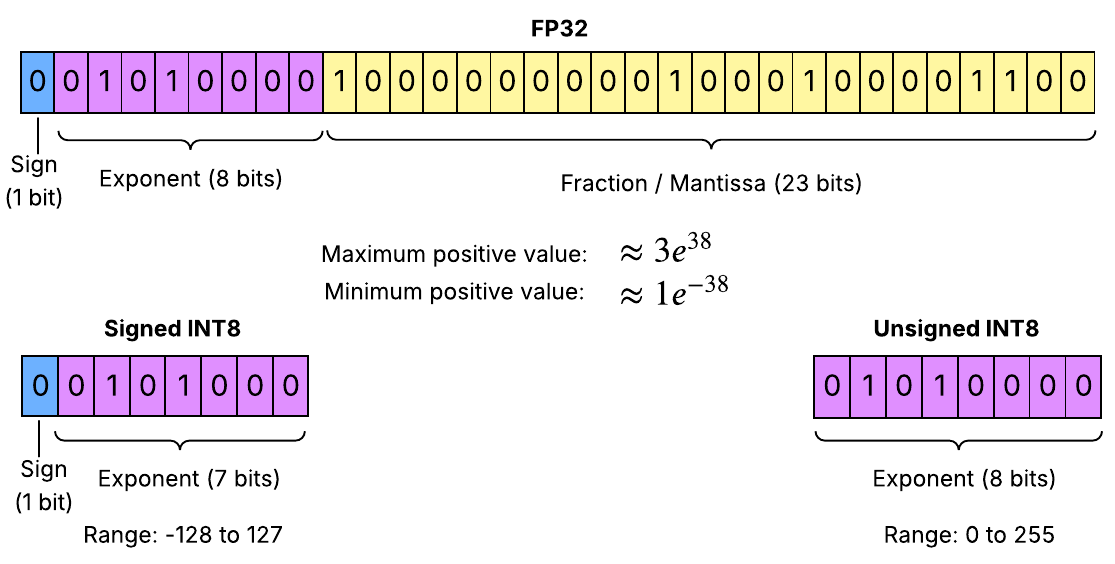}
    \caption{(Top) Representation of the FP32 format, (bottom left) signed INT8 format and (bottom right) unsigned INT8 format.}
     \label{fig:2}
\end{figure}

The most common floating-point format is FP32 (floating-point 32-bit), also known as single or full precision~\cite{rehm2021reduced}. Each FP32 value occupies 32 bits (4 bytes) and follows the IEEE 754 standard~\cite{kahan1996ieee}. A number $n$ in FP32 is represented as:
\begin{equation}
n = (-1)^s \times (1 + m) \times 2^{(e - 127)},
\end{equation}
\noindent where $s$ is the sign bit, $e$ is the exponent (with a bias of 127), and $m$ is the fractional part of the mantissa (also known as the significand), which is stored using 23 bits. The implicit leading 1 in the mantissa provides an effective precision of 24 bits. FP32 supports a wide dynamic range, approximately from \(10^{-38}\) to \(10^{38}\) (see Fig.~\ref{fig:2}, top), which typically exceeds the requirements for DNN inference~\cite{sze2017efficient}.

However, this precision comes at a cost. It results in higher memory usage and requires more complex arithmetic hardware. Moreover, the increased complexity of FP32 arithmetic typically requires a dedicated floating-point unit (FPU) to perform these calculations. The high power consumption and silicon area of FPUs limit their suitability for low-power embedded systems \cite{hassan2020design}.

On the other hand, fixed-point formats encode real numbers as scaled integers. A real number $n$ in fixed-point format is typically represented as:
\begin{equation}
n = \text{integer\_value} \times S,
\end{equation}
\noindent where $\text{integer\_value}$ is a signed or unsigned integer stored in hardware, and $S$ is a global scaling factor (power of two or arbitrary) that determines the position of the radix point. The separation between the integer and fractional parts is implicit and controlled by this scaling factor.

A major advantage of fixed-point formats is that they use simpler and more efficient integer arithmetic. Operations like addition, multiplication, and accumulation can be performed without the need for FPUs. As a result, fixed-point arithmetic is widely adopted in MCUs, digital signal processors, and low-power embedded platforms~\cite{przybyl2021fixed}.

A commonly used fixed-point format is the INT8, which uses only 1 byte per value. Signed INT8 values range from \([-128, 127]\) (see Fig.~\ref{fig:2}, bottom left), while unsigned values range from \([0, 255]\) (see Fig.~\ref{fig:2}, bottom right). Although fixed-point representations offer less dynamic range and lower precision than floating-point formats, they provide substantial benefits in terms of memory savings, computational speed, and energy efficiency.

The dynamic range of a numerical format refers to the ratio between the largest and smallest values it can represent. Fixed-point formats have a narrower dynamic range, making them more suitable for data with limited variation, such as activations or weights in well-calibrated layers. Despite this limitation, fixed-point formats remain the preferred choice for efficient inference on edge devices due to their practical advantages.

The choice of numerical representation has a direct impact on computational efficiency, particularly in deep neural network workloads where multiply-accumulate (MAC) operations dominate both training and inference costs~\cite{capra2020hardware}. Fixed-point arithmetic, especially INT8, offers substantial advantages over FP32 formats in this context. A MAC operation performed using INT8 consumes approximately $20\times$ less energy than with FP32~\cite{horowitz20141}, while addition operations can yield up to $30\times$ energy savings due to simpler circuit logic~\cite{wu2020integer}. The INT8 representation also reduces memory footprint by $4\times$, resulting in lower memory bandwidth and cache pressure~\cite{capra2020hardware}. Processing speed can improve significantly on optimized hardware, with typical acceleration factors ranging from $3.3\times$ for additions to $7.5\times$ for multiplications~\cite{wu2020integer}. These combined benefits make INT8 arithmetic highly effective for energy-efficient, low-latency inference in embedded and real-time systems. A comparative summary is presented in Table~\ref{table:int8_vs_fp32}.

\begin{table}[htbp]
\centering
\caption{Efficiency comparison between INT8 and FP32-based operations.}
\label{table:int8_vs_fp32}
\scriptsize
\setlength{\tabcolsep}{4pt}
\renewcommand{\arraystretch}{1.2}
\begin{tabular}{p{0.18\linewidth} p{0.18\linewidth} p{0.20\linewidth} p{0.15\linewidth}}
\toprule
\textbf{INT8-based Operation} &
\textbf{Energy Saving over FP32} &
\textbf{Memory Reduction over FP32} &
\textbf{Processing Speedup over FP32} \\
\midrule
Addition       & $30\times$ & $4\times$ & $3.3\times$ \\
Multiplication & $20\times$ & $4\times$ & $7.5\times$ \\
\bottomrule
\end{tabular}
\end{table}

Although other formats such as Posits~\cite{gustafson2017beating} and zero-skewed quantization~\cite{jacob2018quantization} have been explored to strike a better trade-off between range and efficiency, these alternatives are not widely adopted in real-world DNN applications. A more detailed discussion is provided in Section~\ref{sec:alternative_formats}.

With this foundation in numerical formats and their implications for inference, we now formalize the quantization process as the transformation of high-precision real values into compact low-precision representations that are suitable for efficient hardware execution.

\subsection{Formalization of Quantization}
\label{sec:quantization}

Consider a deep neural network (DNN) with \( L \) layers and learnable parameters \( \theta = \{W_1, W_2, ..., W_L\} \), applied to a supervised learning task. Each \( W_i \) denotes the weight matrix of the \( i^\text{th} \) layer. Given an input \( \mathbf{x} \in \mathbb{R}^{d} \), where \( d \) is the input dimensionality, the network computes predictions \( \hat{y} = f(\mathbf{x}; \theta) \) through a sequence of transformations. Assume that the model has been trained and the parameters \( \theta \) are stored in FP32 precision.

Quantization aims to reduce the numerical precision of both the model parameters and the intermediate activations, replacing floating-point values with low bit-width representations (e.g., INT8), while minimizing the loss in model performance (e.g., accuracy). This reduction is critical in scenarios constrained by memory, compute, or energy resources \cite{liu2025low}. The quantization task can be formulated as a constrained optimization problem:

\begin{equation}
\begin{aligned}
\min_{\hat{M}} \quad & \frac{1}{|D_{\text{test}}|} \sum_{(x, y) \in D_{\text{test}}} \mathcal{L}\left( \hat{M}(x), y \right), \\
\text{s.t.} \quad & \text{Storage}(\hat{M}) \leq C,
\end{aligned}
\end{equation}

\noindent where \( M \) is the original floating-point model and \( \hat{M} \) is its quantized counterpart. The dataset \( D_{\text{test}} \) is the test set with input-output pairs \( (x, y) \), and \( \mathcal{L}(\cdot, \cdot) \) denotes the task-specific loss function (e.g., cross-entropy). The constraint \( \text{Storage}(\hat{M}) \leq C\) enforces a limit on the memory footprint of the quantized model. In practical deployments, additional constraints—such as inference latency, throughput, or energy consumption—may also guide the quantization process depending on the target hardware or application.

To enable quantization, we define a quantization function \( q(\cdot)\) that maps a real-valued input to a quantized integer. This can be generally written as:
\begin{equation}\label{eq:quant}
    \mathbf{x}_{\text{int}} = q(\mathbf{x}; s, z) = \text{clamp} \left(\Big \lfloor \frac{\mathbf{x}}{s} \Big \rceil+ z~; q_{min}, q_{max} \right),
\end{equation}

\noindent where \( \mathbf{x} \) is the original value, \( \mathbf{x}_{\text{int}} \) is the quantized integer. The scale \( s \) defines the quantization resolution of the quantizer \cite{nagel2021white} and is typically computed as:
\begin{equation}
s = \frac{\text{max}(\mathbf{x}) - \text{min}(\mathbf{x})}{2^b - 1},
\end{equation}
\noindent where \( b \) is the bit-width of the integer representation (e.g., \( b = 8 \) for INT8). The zero-point \( z \) (also called quantization bias) is an optional integer parameter that shifts the quantized range so that real zero is mapped exactly \cite{jacob2018quantization}. 

The rounding operation \( \left\lfloor \cdot \right\rceil \) maps the scaled input to the nearest integer and introduces quantization error. This operation is non-differentiable, which has important implications for training quantized models, as will be discussed later.

The clamp$(\cdot)$ function restricts the result to lie within the target integer range \([q_{\text{min}}, q_{\text{max}}]\), preventing overflow and underflow. The clamping operation is formally defined as:
\begin{equation}
    \text{clamp}(x, a, b) =
    \begin{cases}
        a & \text{if } x < a, \\
        x & \text{if } a \leq x \leq b, \\
        b & \text{if } x > b.
    \end{cases}
    \label{eq:clip_function}
\end{equation}

To recover an approximate real-valued input from a quantized integer, a dequantization step is applied:
\begin{equation}
    \mathbf{x} \approx s \cdot (\mathbf{x}_{\text{int}} - z),
    \label{eq:dequantization_process}
\end{equation}

\noindent enabling downstream computation using approximate values that can be efficiently processed in integer arithmetic units.

In sum, quantization seeks to determine the optimal quantization parameters, namely, the bit-width \( b \), scale \( s \), and zero-point \( z \) that balance model accuracy with efficiency gains in memory, computation, and energy. In the following sections, we introduce various quantization techniques that realize this objective.

\subsection{Calibration and Clipping Range Selection}
\label{sec:calibration}

As noted earlier, the range \([q_{\text{min}}, q_{\text{max}}]\) represents the \textit{clipping range} of the quantizer. Improper choice of this range can lead to clipping errors or large quantization errors. The process of determining appropriate clipping bounds is referred to as \textit{calibration}. Below, we describe several commonly used calibration methods:

\subsubsection{Min-Max Calibration}
The most straightforward approach is to cover the full range of the input tensor \( \mathbf{x} \) by setting:
\begin{equation}
    q_{\text{min}} = \min(\mathbf{x}), \quad q_{\text{max}} = \max(\mathbf{x}).
\end{equation}

While simple, this method is highly sensitive to outliers, which can disproportionately expand the dynamic range and lead to poor quantization fidelity \cite{gholami2022survey}.

\subsubsection{Mean Squared Error (MSE)}
To mitigate the impact of outliers and reduce total quantization error, the clipping range can be selected to minimize the reconstruction error between the original and quantized-dequantized values. Specifically, the optimal range is defined as:
\begin{equation}
(q_{\text{min}}^{*}, q_{\text{max}}^{*}) = \underset{q_{\text{min}}, q_{\text{max}}}{\text{arg min}} \left\lVert \mathbf{x} - q^{-1}(q(\mathbf{x}; s, z); s, z) \right\rVert_{2}^{2},
\end{equation}

\noindent where \( q(\mathbf{x}; s, z) \) denotes the quantization function from \eqref{eq:quant}, \( q^{-1}(\cdot; s, z) \) is the corresponding dequantization operation, and \( \left\lVert \cdot \right\rVert_{2} \) is the Euclidean or L2 norm.

\subsubsection{Percentile-Based Calibration}

In this approach, the quantization range is selected based on the empirical distribution of the dataset by selecting specific percentiles to define the lower and upper bounds of the range \cite{liu2025low}. For example, using the 99th percentile method, \( q_{\text{min}} \) is set to the 1$^\text{st}$ percentile and \( q_{\text{max}} \) to the 99$^\text{th}$ percentile of the dataset. Variants using 99.9$^\text{th}$ or 99.99$^\text{th}$ percentiles are also used in literature \cite{abushahla2025cognitive, mckinstry2019discovering}. This method effectively excludes extreme outliers and tends to yield more robust quantization results across layers. 
For a comprehensive comparison of calibration techniques and their effects on model accuracy and deployment, we refer interested readers to~\cite{wu2020integer}.

\subsection{Uniform and Non-Uniform Quantization}
\label{sec:uniform_non-uniform}
\begin{figure}[htbp]
    \centering
    \includegraphics[width=0.45\textwidth]{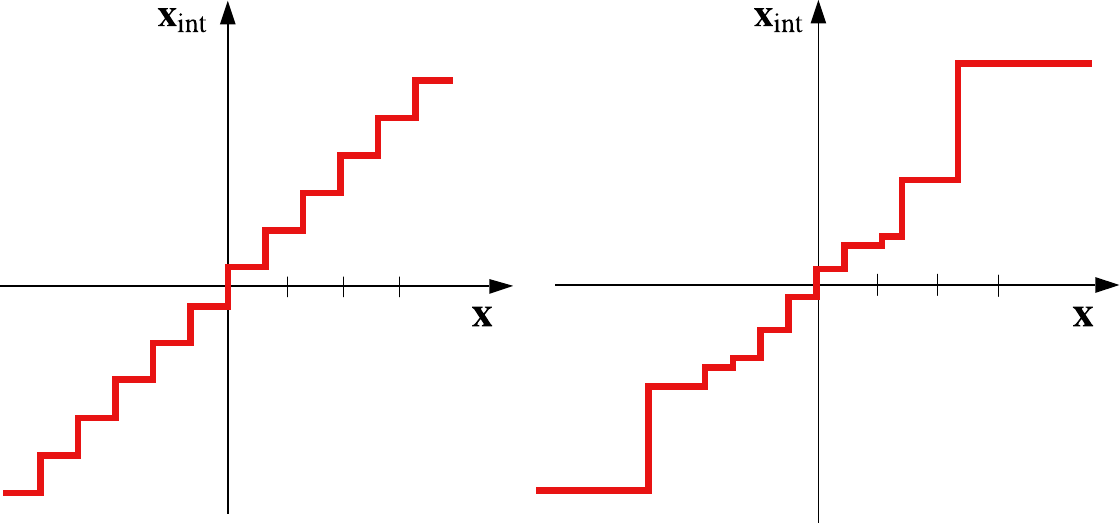}
    \caption{Uniform quantization (left) vs. Non-uniform quantization (right). Real values $\mathbf{x}$ are mapped into lower precision integer values $\mathbf{x}_{\text{int}}$. The distances between the quantized values (quantization levels) are the same in uniform quantization, whereas they can vary in non-uniform quantization.}
     \label{fig:Uniform_NonUniform}
\end{figure}

The quantization method defined in~\eqref{eq:quant} is known as \textit{uniform quantization}~\cite{nagel2021white}. In this approach, the quantization levels are \textit{evenly spaced}, meaning that each FP32 input is mapped to the nearest integer level using a fixed step size and rounding function \( \left\lfloor \cdot \right\rceil \). This results in a grid-like discretization of the input space, where values within the same quantization interval share the same quantized representation (see Fig.~\ref{fig:Uniform_NonUniform}, left).

Uniform quantization is widely adopted due to its efficient implementation using fixed-point arithmetic~\cite{nagel2021white}, which enables fast and low-power deployment on resource-constrained hardware such as mobile and embedded devices. However, it is important to note that the distributions of weights and activations in neural networks are often non-uniform~\cite{miyashita2016convolutional}. As such, applying uniform quantization may lead to suboptimal allocation of representational capacity, especially in low bit-width regimes.

This motivates the use of \textit{non-uniform quantization}, where the quantization levels are not evenly spaced (see Fig.~\ref{fig:Uniform_NonUniform}, right). These methods adapt the step sizes to the data distribution, typically assigning more resolution to high-density regions (i.e., near zero) and less to low-density regions. A general formulation is given by:
\begin{equation}
    \mathbf{x}_{\text{int}} = q_{\text{non-uniform}}(\mathbf{x}) = X_i \quad \text{if } \mathbf{x} \in [\Delta_i, \Delta_{i+1}),
\end{equation}
\noindent where \( \{ \Delta_i \} \) denotes variable-width partition intervals over the real-valued domain, and \( \{ X_i \} \subset \mathbb{Z} \) are the corresponding quantized integer levels. Unlike uniform quantization, neither the intervals \( \Delta_i \) nor the outputs \( X_i \) are required to be equidistant.

Several notable strategies have been proposed to implement non-uniform quantization. One prominent example is \textit{logarithmic quantization}, as used in LogNet~\cite{7953288}, where quantization levels are spaced exponentially. This allows for finer granularity near zero and is well-suited for data with a wide dynamic range. Another approach is \textit{vector quantization} (also referred to as weight sharing), which clusters weights using algorithms such as $k$-means and replaces each weight with a representative cluster centroid. This technique has been successfully applied in methods like Deep Compression~\cite{han2016deep} and incremental network quantization (INQ)~\cite{zhou2017incremental}.

However, non-uniform quantization comes with practical limitations. Unlike uniform schemes, it is generally incompatible with fast, fixed-point linear operations, requiring lookup tables or non-linear transformations. Consequently, it is less hardware-friendly and may not achieve the same acceleration benefits on platforms designed for linear quantization.

One notable highlight is the Nonuniform-to-Uniform quantization (N2UQ) \cite{liu2022nonuniform}, which combines the representational advantages of non-uniform quantization with the hardware-friendliness of uniform quantization. N2UQ learns flexible non-uniform input thresholds, but the outputs are mapped to uniformly spaced intervals. To address the challenge of training with learnable thresholds (where gradients are intractable using the conventional straight-through estimator (STE)), the authors introduce a generalized STE (G-STE), derived from the expectation of stochastic quantization. This enables backpropagation through nonuniform input thresholds while preserving deterministic outputs. N2UQ achieves state-of-the-art performance on benchmarking datasets with only 2-bit precision, maintaining high accuracy whilst remaining hardware-friendly.

In essence, while non-uniform quantization techniques can significantly improve accuracy in low-precision regimes by better matching the underlying data distribution, uniform quantization remains the dominant choice for deployment efficiency due to its simplicity and hardware compatibility.


\subsection{Symmetric and Asymmetric Quantization}

An important design choice in uniform quantization is whether to employ a \textit{symmetric} or \textit{asymmetric} scheme. This affects how the scale \( s \) and zero-point \( z \) are computed in \eqref{eq:quant}, and how well the quantization grid aligns with the data. The following subsections compare their formulations, use cases, and trade-offs.

\subsubsection{Symmetric Quantization}

\begin{figure*}[htbp]
    \centering
    \includegraphics[width=0.9\textwidth]{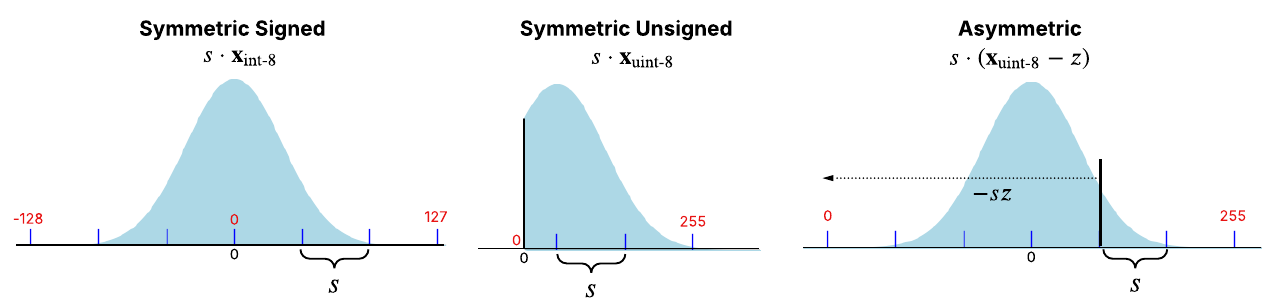}
    \caption{Illustration of different quantization schemes: symmetric signed (left), symmetric unsigned (center), and asymmetric (right). Red markers denote the INT8 quantization range. The black labeled “0” on the real axis indicates the true zero in floating-point space, while the red zero shows its corresponding integer location after quantization. In symmetric quantization, the true zero and the mapped zero-point align, whereas in the asymmetric case, the zero-point is shifted.}
    \label{fig:symmetric_asymmetric}
\end{figure*}

In symmetric quantization, it is assumed that the values follow a distribution centered around zero—commonly a Gaussian distribution. As a result, the minimum and maximum values are equal in magnitude but opposite in sign. The zero-point is fixed at zero, which simplifies hardware implementation and computation. Symmetric quantization can be further categorized into symmetric signed (see Fig.~\ref{fig:symmetric_asymmetric}, left) and symmetric unsigned quantization (see Fig.~\ref{fig:symmetric_asymmetric}, center).

\textbf{Symmetric Signed Quantization.} In this setting, values span both negative and positive ranges, making it particularly suitable for weight tensors. The scale \( s \) is computed based on the maximum absolute value in the tensor:
\begin{equation}
    s = \frac{\text{max}(|\mathbf{x}|)}{127},
    \label{eq:scale_symmetric_signed}
\end{equation}

\noindent where the quantized range corresponds to an 8-bit signed integer: \([-128, 127]\). This ensures that the largest absolute value in the tensor \( \mathbf{x} \) maps directly to the maximum representable integer value of 127. Given that the zero-point \( z \) is fixed at zero, the dequantization formula simplifies to:
\begin{equation}
    \mathbf{x} \approx s \cdot \mathbf{x}_{\text{int}} .
    \label{eq:dequantization_symmetric_signed}
\end{equation}

This method is efficient and particularly effective for data centered around zero, such as weights in neural networks. The fixed zero-point simplifies hardware implementation and is efficient for data with balanced positive and negative ranges. However, it may result in poor dynamic range utilization if the data is significantly skewed.

\textbf{Symmetric Unsigned Quantization.} 
Here, values are assumed to be strictly non-negative, with the zero-point still fixed at zero. This makes it ideal for activation tensors following ReLU operations. For an 8-bit unsigned integer range \([0, 255]\). The scale \( s \) is computed as:
\begin{equation}
    s = \frac{\text{max}(\mathbf{x})}{255}
    \label{eq:scale_symmetric_unsigned}.
\end{equation}

This method is well-suited for skewed, one-tailed distributions, where values are strictly non-negative (i.e., ranging from 0 to a positive maximum), but unsuitable when values include negative components~\cite{rehm2021reduced}.

\subsubsection{Asymmetric Quantization}
Asymmetric quantization removes the assumption that the data is centered around zero, enabling more flexible mapping for skewed or unbalanced distributions. It introduces a non-zero zero-point \( z \), which ensures that the true zero in the real-valued range is exactly representable in the quantized domain (see Fig.~\ref{fig:symmetric_asymmetric}, right). This is especially useful for activation tensors, which may span arbitrary ranges.

For an 8-bit representation (e.g., \([0, 255]\)), the scale and zero-point are computed as:
\begin{align}
    s &= \frac{\text{max}(\mathbf{x}) - \text{min}(\mathbf{x})}{255}, \label{eq:scale_asymmetric} \\
    z &= \left\lfloor -1 \times \frac{\text{min}(\mathbf{x})}{s} \right\rceil .\label{eq:zero_point_asymmetric}
\end{align}

Asymmetric quantization is ideal for activations and other data distributions that are not centered around zero, providing better accuracy for skewed or unbalanced data. The ability to adjust the zero-point introduces additional flexibility but also adds complexity to hardware implementations due to the variable zero-point.

\subsection{Static And Dynamic Quantization}

Another important aspect of quantization design is \textit{when} the clipping range $(q_{\text{min}}, q_{\text{max}})$ is determined. For weights, which remain fixed after training, this range is typically computed once and reused during inference. In contrast, activations vary with each input sample $\mathbf{x}$, making the choice between \textit{static} and \textit{dynamic} quantization particularly relevant~\cite{nagel2021white}.

\subsubsection{Static Quantization} Static quantization uses a fixed clipping range that is pre-computed, typically during a calibration phase, on a representative dataset. This approach avoids runtime overhead and is well-suited for deployment in latency-critical or resource-constrained environments. While it lacks adaptability to input variations, static quantization is widely adopted in practice due to its simplicity and efficiency.

The calibration process involves estimating activation ranges from sample inputs~\cite{jacob2018quantization, yao2021hawq}. Several methods have been proposed for determining optimal clipping thresholds (see Section~\ref{sec:calibration}), including minimizing the MSE between the original and quantized values~\cite{sung2015resiliency, shin2016fixed, choukroun2019low, zhao2019improving}. Other alternatives include entropy-based metrics~\cite{park2017weighted} and learning the clipping thresholds directly during training~\cite{li2019fully, choi2018pact, zhu2016trained, zhang2018lq}.

\subsubsection{Dynamic Quantization} In dynamic quantization, the scale factor and zero-point are \textit{dynamically} computed on-the-fly during inference using real-time statistics (e.g., min/max or percentiles) of the incoming activation values \cite{gholami2022survey}. This allows the quantizer to adapt precisely to the input distribution, often resulting in higher accuracy. However, the adaptivity comes at the cost of increased computational overhead due to the need to recompute quantization parameters for each input.

Thus, dynamic quantization adapts to each input by computing quantization parameters at runtime, achieving higher accuracy but at increased computational cost. Static quantization, on the other hand, fixes these parameters post-calibration, enabling efficient inference at the cost of some accuracy loss. For more adaptive strategies that bridge these trade-offs, see Section~\ref{subsubsection:adaptive}.

\subsection{Quantization Granularity}

\begin{figure}[ht]
    \centering
    \includegraphics[width=0.40\textwidth]{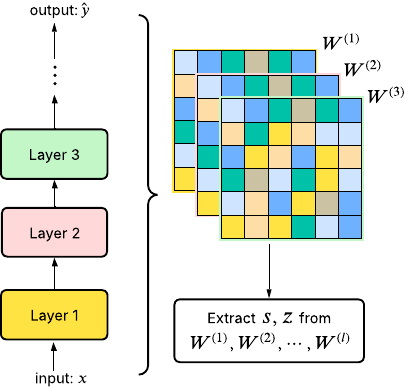}
    \caption{Simplified illustration of global quantization, where all weight matrices \( W^{(1)} \), \( W^{(2)} \), and \( W^{(3)} \) across different layers share the same quantization parameters.}
    \label{fig:global_quantization}
\end{figure}

\begin{figure}[ht]
    \centering
    \includegraphics[width=0.45\textwidth]{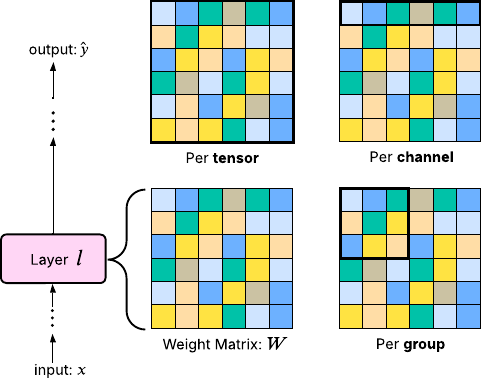}
     \caption{Simplified illustration of quantization granularity for the weight matrix $W$ of a given layer $l$. Quantization can be done per-tensor, per channel, or per group.}
    \label{fig:granularity}
\end{figure}

Quantization granularity refers to the level at which quantization parameters are applied within a neural network. This design choice directly impacts model accuracy, computational cost, and storage requirements. Coarser granularities simplify implementation but may compromise accuracy due to a less precise range of data across layer filters.

\textbf{Global:} Also referred to as \textit{network-wise} quantization, this approach applies the same quantization parameters, namely a single shared scale and zero-point, uniformly across all weight tensors in the network~\cite{jacob2018quantization, lin2016fixed}, as illustrated in Fig.~\ref{fig:global_quantization}. By quantizing all layers using the same dynamic range, this method simplifies hardware implementation, reduces metadata storage, and minimizes computational complexity. However, such coarse granularity often results in significant accuracy degradation due to the large variation in value distributions across layers \cite{lou2020autoq, choukroun2019low}.

Beyond global quantization, finer-grained strategies are commonly used and can be classified into the following levels, arranged from coarse to fine granularity (see Fig.~\ref{fig:granularity}).


\textbf{Per-tensor:} Also known as per-layer, this method applies a single set of quantization parameters to the entire tensor, so all elements share the same clipping range~\cite{krishnamoorthi2018quantizing}. It is the most common choice for \textit{activations} due to its simpler hardware implementation.


\textbf{Per-channel:} Also known as per-axis, this method assigns a distinct set of quantization parameters to each output channel, so every channel is scaled and clipped according to its own distribution. Compared to per-tensor (a single shared range), per-channel captures inter-channel variation, reducing quantization error and typically improving accuracy, while eliminating the need for global rescaling across channels. In practice, it is the standard choice for convolutional/linear \emph{weights}, where channel-wise distributions often differ substantially.

\textbf{Per-group:} This method assigns a unique quantization parameter to each group of channels within a layer. It can be helpful when the distribution of the parameters within a layer varies significantly. For instance, it has been used in Transformer models (e.g., Q-BERT~\cite{shen2020q}) with fully connected attention blocks~\cite{vaswani2017attention}. However, this approach inevitably comes with the extra cost of accounting for different scaling factors. All in all, it offers a middle ground between per-tensor and per-channel.

\textbf{Per-token:} Applied mostly to language models, this approach assigns quantization parameters at the token level (words or characters), offering fine-grained precision.


In summary, quantization granularity spans from global to per-tensor, per-channel, and per-group---and, in language models, per-token. Moving to finer granularity generally improves accuracy by better matching local distributions, at the cost of extra quantization parameters and kernel complexity. In practice, many CNN deployments pair per-channel for weights with per-tensor for activations, balancing accuracy with implementation simplicity. We refer interested readers to~\cite{garg2021confounding} for trade-offs associated with these design choices.

\section{Primary Quantization Techniques}
\label{sec:quant_techniques}

\begin{figure}[htbp]
    \centering
    \includegraphics[width=0.45\textwidth]{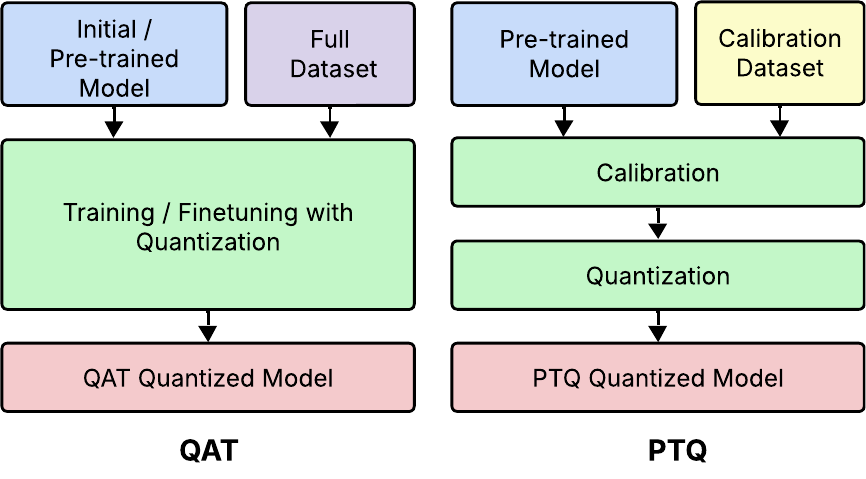}
    \caption{Comparison between QAT and PTQ. In QAT (left), quantization is simulated during training, either from scratch or as a fine-tuning phase on a pre-trained model, allowing the model to adapt its parameters to mitigate quantization-induced errors. In PTQ (right), a pre-trained model is calibrated using a small dataset to determine clipping ranges and scaling factors, after which the model is quantized without further training.}
     \label{fig:4}
\end{figure}

As for how quantization is applied, there generally exist two different approaches that are most commonly used, they are as depicted in Fig.~\ref{fig:4}.

\subsection{Post-Training Quantization}

Post-training quantization (PTQ) algorithms, as the name suggests, are post-processing techniques that take pre-trained FP32 models and quantize their parameters (e.g., weights and activations), effectively converting them into a fixed-point network without the need for the original training pipeline \cite{nagel2021white}. It is considered the simplest quantization technique and is especially useful for cases where data availability is limited (due to, e.g., privacy concerns or edge deployment) \cite{Stromberg1562946}. 

Depending on the availability of data, PTQ methods can be either data-driven or data-free. Data-driven PTQ uses a small calibration set to estimate the appropriate quantization parameters (e.g., scale and zero-point), while data-free approaches generate synthetic data~\cite{cai2020zeroq} or leverage model statistics~\cite{nagel2019data} to achieve calibration (see Section~\ref{sec:data-agnostic}). Once calibration is performed, the quantized model can be used directly for inference without any further modification to its weights or structure.

As no further training is needed, this approach is straightforward and computationally inexpensive, making it an attractive choice for deployment. Once a model is quantized, it can be deployed to perform low-precision inference and associated computations, without any adjustments to the parameters. However, this convenience often comes at the cost of reduced accuracy. Quantization introduces numerical errors in parameters and activations (including the input), which propagate through subsequent layers. The accumulation of these errors can lead to misclassifications at the network output, producing an accuracy drop relative to the full precision model. As the bit-width decreases (e.g., INT8 or lower), these quantization errors tend to grow, increasing the likelihood of accuracy degradation, particularly in deeper or more sensitive models~\cite{nagel2021white, novac2021quantization}.

Despite this limitation, PTQ is widely used in practice and suffices for many TinyML and edge AI applications, where the trade-off between accuracy and efficiency is acceptable. 

\subsection{Quantization-Aware Training}

Quantization-Aware Training (QAT) addresses the limitations of PTQ by simulating quantization effects during training. Unlike PTQ, where quantization noise is introduced after training, QAT incorporates quantization directly into the forward pass, allowing the model to learn parameters that are more resilient to quantization errors~\cite{jacob2018quantization}.

This is achieved by inserting ``fake'' quantization operations into the network, which simulate low-precision inference-time computation, such as INT8 arithmetic, during the forward pass of training. Specifically, both the inputs and weights are quantized to a fixed-point representation and then immediately dequantized back to floating-point. Although the underlying model parameters are stored and updated in FP32, this emulation allows the model to experience the quantization error that would occur during inference.  As a result, the introduced error becomes part of the overall loss that the optimization algorithm, through the loss function and backpropagation, actively tries to minimize. By exposing the network to INT8-like computations during training, the model learns weights that are inherently more robust to quantization effects~\cite{Stromberg1562946}. It is important to note that this quantization emulation is applied only during the forward pass; the backward pass, including gradient computation and parameter updates, remains in full precision (FP32).

\begin{figure*}[htbp]
    \centering
    \includegraphics[width=0.80\textwidth]{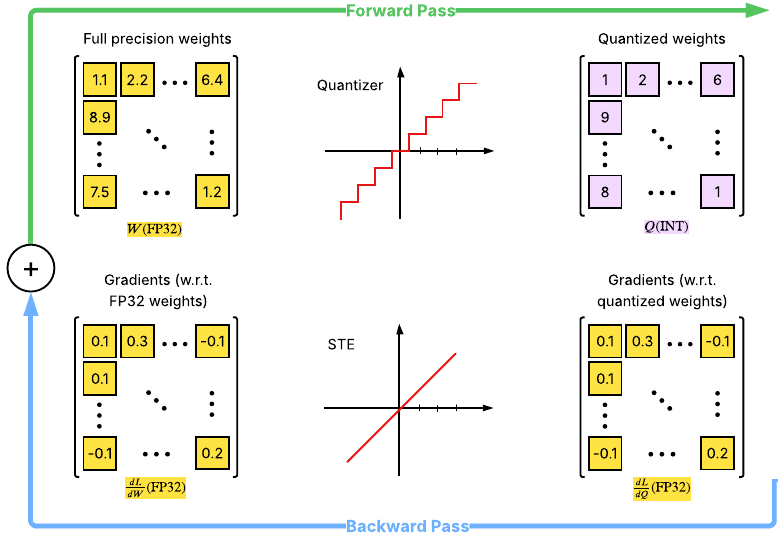}
    \caption{Illustration of the QAT procedure, including the use of the STE. The full precision weights are initially quantized in the forward pass. In the backward pass, the gradients are calculated w.r.t. the quantized weights using the STE.}
     \label{fig:QAT}
\end{figure*}

The backward pass remains in floating-point, but it must overcome a critical challenge: the quantization function, defined in~\eqref{eq:quant}, includes the non-differentiable rounding operation \( \lfloor \cdot \rceil \). This operation maps continuous values to discrete integers, preventing standard gradient-based optimization \cite{wu2020integer}. To address this challenge, QAT typically employs the STE, which approximates the gradient of the quantization function by treating it as the identity function within the clipping range and assigning zero gradients outside of it. This approach replaces the piecewise constant quantizer, whose gradient is zero almost everywhere, with a differentiable approximation that allows gradient flow during backpropagation~\cite{jacob2018quantization, bengio2013}, as illustrated in Fig.~\ref{fig:QAT}. The gradient approximation is given by:

\begin{equation}
    \frac{\partial \mathbf{x}_{\text{int}}}{\partial \mathbf{x}} \approx 
    \begin{cases}
        1, & q_{min} s \leq \mathbf{x} + z \leq q_{max} s, \\
        0, & \text{otherwise.}
    \end{cases}
\end{equation}

Recent research has explored alternatives to STE by introducing differentiable approximations to the quantization function. These include methods such as Differentiable Soft Quantization (DSQ)\cite{gong2019}, which uses a smooth function like a scaled \texttt{tanh} prior to quantization, or replacing the quantizer with a nonlinear differentiable function during training\cite{yang2019}.

\begin{figure}[htbp]
    \centering
    \begin{subfigure}{0.49\linewidth}
        \centering
        \includegraphics[width=\linewidth]{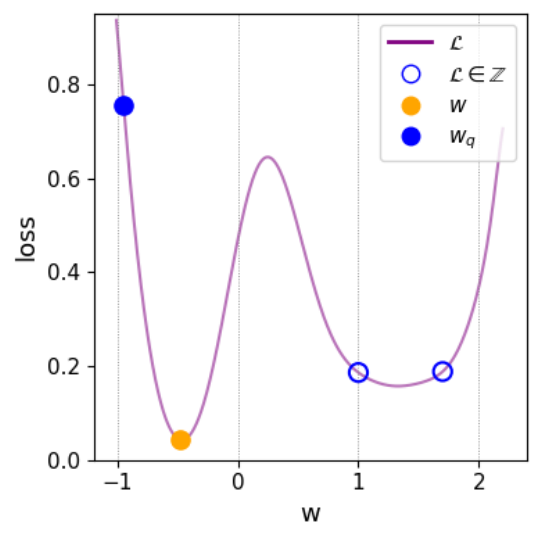}
        \caption{After PTQ}
        \label{fig:PTQ_Curve}
    \end{subfigure}
    \begin{subfigure}{0.49\linewidth}
        \centering
        \includegraphics[width=\linewidth]{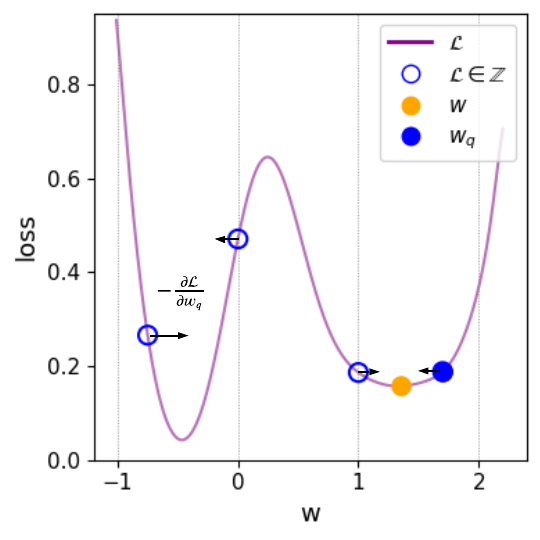}
        \caption{After QAT}
        \label{fig:QAT_Curve}
    \end{subfigure}
    \caption{Illustration of quantization effects on the loss landscape. The parameter \( w \) is quantized with a scale factor of 1. (a) PTQ: The model converges to a narrow minimum at \( w \approx-0.5\) (orange), which is then quantized to \( w_q = -1\) (blue), resulting in a significant increase in loss. (b) QAT: The model converges to a flatter region of the loss surface, where quantization to \( w_q \) introduces only a minor loss increase, demonstrating improved robustness to quantization.}

    \label{fig:QAT_PTQ_Curves}
\end{figure}

To better understand why QAT improves the robustness of quantized models, consider the example in Fig.~\ref{fig:QAT_PTQ_Curves}, which visualizes the impact of quantization on the loss landscape. DNNs are trained by minimizing a loss function \( L \), typically using stochastic gradient descent (SGD). During training, the gradient \( \frac{\partial L}{\partial w} \) is computed with respect to each parameter \( w \), and the weights are updated iteratively in the direction of the negative gradient until the model converges to a local or global minimum.

On the one hand, Fig.~\ref{fig:QAT_PTQ_Curves}(a) illustrates a one-dimensional loss curve for a model trained in full precision and later quantized using PTQ. Suppose the parameter converges to \( w \approx -0.5 \). Applying uniform quantization with scale factor $s=1$, this value is rounded to the nearest quantized level \( w_q = -1 \), according to \eqref{eq:quant}. This quantization step introduces a significant perturbation in the parameter value, resulting in a sharp increase in the loss. In this case, the model is said to have converged to a \textit{narrow minimum}, where small deviations in weight values lead to large variations in the loss function~\cite{wu2020integer}.

On the other hand, Fig.~\ref{fig:QAT_PTQ_Curves}(b) depicts training under QAT. Here, the forward pass simulates quantization during training, and gradients are computed with respect to the quantized weights using the STE. This approach ensures that the optimization process accounts for quantization effects throughout training. As a result, the model is guided toward \textit{flat} or \textit{wide} minima in the loss landscape, where small perturbations due to rounding have minimal impact on the loss value. These flatter regions provide better resilience to quantization, resulting in improved accuracy and generalization after deployment at low precision~\cite{wu2020integer}.

The underlying intuition is that in QAT, quantization noise is included as part of the training process, and thus the optimization algorithm actively seeks parameter configurations where this noise does not drastically alter model predictions. In contrast, PTQ does not account for quantization effects during training, and any deviation due to rounding is introduced after the model has converged, often leading to suboptimal post-quantization performance.

While QAT significantly improves robustness to quantization, it introduces additional training overhead due to the insertion of fake quantization operations during the forward pass. To reduce this computational burden, a widely adopted strategy is to apply QAT as a fine-tuning phase on top of a pre-trained FP32 model. This approach avoids introducing quantization noise early in training and allows the model to first learn rich and expressive representations in full precision.

Fine-tuning with QAT has been shown to converge more quickly and achieve higher post-quantization accuracy than training from scratch under low-precision constraints~\cite{wu2020integer}. Directly training with QAT exposes the model to quantization-induced perturbations from the outset, which can impair its ability to learn robust features. In contrast, beginning with a fully converged FP32 model provides a stable initialization, and the subsequent fine-tuning step helps the model adapt smoothly to quantized weights and activations. This two-stage training pipeline effectively mitigates performance degradation while preserving the accuracy benefits of full precision training.

Given its effectiveness in preserving model quality, QAT-based fine-tuning is widely favored in real-world deployments, particularly in resource-constrained environments where inference efficiency and predictive performance are both critical.

\section{Advanced Quantization Techniques}
\label{sec:advanced}

\begin{figure*}[!t]
  \centering
  \resizebox{\textwidth}{!}{%
    \tikzset{
      my node/.style={
        draw=black,
        thick,
        rounded corners=2,
        font=\footnotesize\rmfamily,
        align=center,
        inner sep=5pt,
      },
      non leaf/.style={ text width=3cm },
      leaf/.style={ text width=8cm },
    }
    \begin{forest}
      for tree={%
        my node,
        if n children=0{leaf}{non leaf},
        l sep+=5pt,
        grow'=east,
        edge={black, thick},
        parent anchor=east,
        child anchor=west,
        if n children=0{tier=last}{},
        edge path={
          \noexpand\path [draw, \forestoption{edge}] (!u.parent anchor) |- (.child anchor)\forestoption{edge label};
        },
        if={level()==1}{fill=yellow!25}{},
        if={level()==2}{fill=blue!7}{},
        if={level()>=3}{fill=gray!2}{},
        if={isodd(n_children())}{
          for children={
            if={equal(n,(n_children("!u")+1)/2)}{calign with current}{}
          }
        }{}
      },
      [Quantization, phantom
        [\ref{sec:advanced}. Advanced Quantization Techniques
          [\ref{sec:INT8_Training}. Integer and Quantized Training
            [{{AQT \cite{lewt2023aqt}, Zhu et al. \cite{zhu2020}, Octo \cite{zhou2021octo}, DDQS \& UDPS \cite{wang2023}}}]
          ]
          [\ref{sec:extreme-low-bit}. Extreme Low‑Bit Quantization
            [Binarization
              [{{BinaryConnect \cite{courbariaux2015binaryconnect}, BNNs \cite{hubara2018}, eBNNs \cite{mcdanel2017, Geiger2020}, XNOR‑Net \cite{rastegari2016xnor}}}]
            ]
            [Ternarization
              [{{TWNs \cite{lin2015neural, li2016ternary}, TBNs \cite{wan2018tbn}}}]
            ]
          ]
          [\ref{subsection:mixed-precision}. Mixed Precision Quantization
            [Rough Allocation
              [{{HAWQ \cite{hawq}, CherryQ \cite{cherryq}, REx \cite{rex}, OWQ \cite{owq}, HAWQ‑v2 \cite{hawqv2}, HTQ \cite{htq}, HAQ \cite{wang2019haq}}}]
            ]
            [Adaptive Allocation
              [{{NIPQ \cite{NIPQ} , MEBQAT \cite{meta-learning}, MetaMix \cite{kim2024metamix}, CADyQ~\cite{cadyq}, CABM \cite{cabm}, \\AdaBM~\cite{hong2024adabm}, CoQuant~\cite{coquant}, ALPS~\cite{langroudi2021alps}}}]
            ]
          ]
          [\ref{sec:hw-aware}. Hardware‑Aware Quantization
            [{{HAQ \cite{wang2019haq}, Rusci et al. \cite{rusci2020}, HAWQ-v3~\cite{yao2021hawq}, OHQ~\cite{huang2023chip}}}]
          ]
          [\ref{sec:redist}. Redistribution
            [Distribution Uniformization
              [{{GDRQ~\cite{yu2020low}, KURE~\cite{chmiel2020robust}, SqWQ~\cite{strom2022squashed}, EQ‑Net \cite{xu2023eq}, KurTail \cite{akhondzadeh2025kurtail}}}]
            ]
            [Distribution Reshaping
              [{{BR \cite{han2021improving}, $R^2$ Loss~\cite{kundu2023r2}}}]
            ]
            [Outlier Redistribution
              [{{SmoothQuant~\cite{xiao2023smoothquant}, OmniQuant~\cite{shao2024omniquant}, FQ-ViT~\cite{lin2021fq}, OS+~\cite{wei2023outlier}, DuQuant~\cite{lin2024duquant}, AdderQuant~\cite{nie2022redistribution}, MagR~\cite{zhang2024magr}}}]
            ]
            [Bias Redistribution
              [{{QuantSR~\cite{qin2023quantsr}, NoisyQuant~\cite{liu2023noisyquant}, RAOQ~\cite{zhang2024reshape}}}]
            ]
          ]
          [\ref{sec:data-agnostic}. Data‑Agnostic Free Quantization
            [{{DFQ~\cite{nagel2019data}, SQuant~\cite{squant}, REx~\cite{rex}, PNMQ~\cite{data-free-non-uniform}, UDFC~\cite{udfc}}}]
          ]
        ]
        [\ref{sec:alternative_formats}. Alternative Numerical \\Representations
          [\ref{sec:FP-based}. Floating‑Point Formats
            [{{FP16~\cite{kahan1996ieee}, BF16~\cite{guntoro2020next}, TF32 \cite{choquette2021nvidia}, CFloat8 \& CFloat16 \cite{tesla_dojo_2025}, MSFP~\cite{darvish2020pushing}, BSFP~\cite{lo2023block}, AdaptiveFloat~\cite{tambe2020algorithm}, DFloat~\cite{zhang202570}, ANT~\cite{guo2022ant}}}]
          ]
          [\ref{sec:FX-based}. Fixed‑Point Formats
            [{{Zero-Skew~\cite{jacob2018quantization}, F8Net~\cite{jin2022f8net}, Vs-Quant~\cite{dai2021vs}}}]
          ]
          [\ref{sec:posit}. Posit Format
            [{{Posit~\cite{gustafson2017beating, gustafson2}}}]
          ]
        ]
      ]
    \end{forest}
  }
  \caption{Taxonomy of quantization techniques and numerical formats.}
  \label{fig:quant-taxonomy}
\end{figure*}
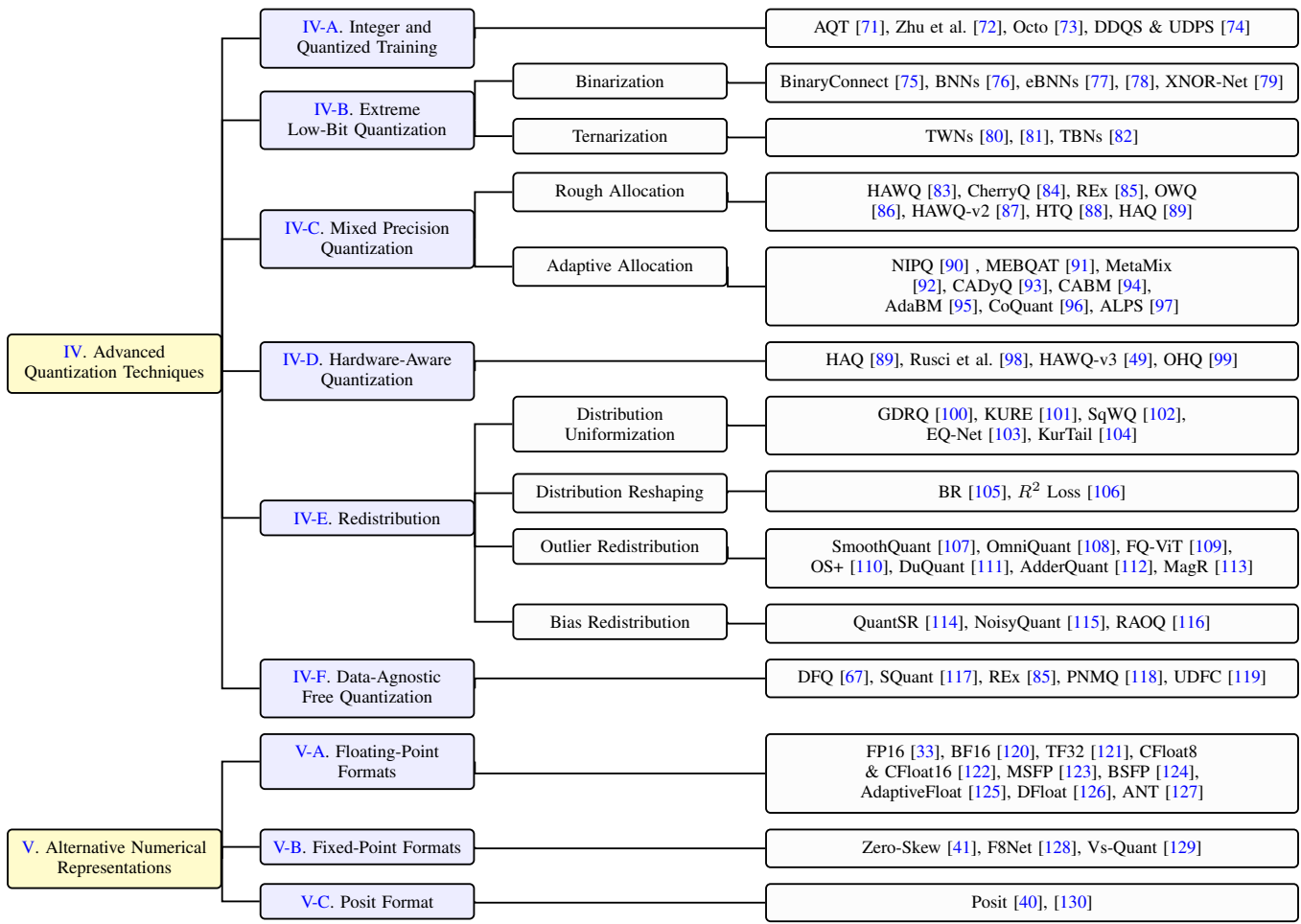

Beyond standard PTQ and QAT, recent research has introduced advancements aimed at improving the execution of DNNs on MCUs and other resource-constrained devices across multiple stages of the deployment pipeline. In the next two sections, Sections~\ref{sec:advanced} and~\ref{sec:alternative_formats}, we survey recent developments in both advanced quantization techniques, applied during training and inference, as well as novel or alternative numerical formats designed to reduce the error typically introduced by low-precision operations. Collectively, these approaches aim to improve quantization robustness, lower training overhead, and expand applicability to challenging deployment scenarios, thereby enabling an even broader range of applications. Fig.~\ref{fig:quant-taxonomy} provides an overview of the methods discussed, categorizing them into major groups based on their core methodological innovations. Notably, some works span multiple categories, reflecting their hybrid or cross-cutting contributions.

\subsection{Integer and Quantized Training}
\label{sec:INT8_Training}

A notable advancement in quantization techniques is Quantized Training (QT), which extends traditional QAT by quantizing both the forward and backward passes, including gradient calculations. This approach retains the benefits of QAT while also accelerating the training process itself. However, QT, particularly with backpropagation quantization, has mostly remained within research contexts due to its software complexity and hardware demands. To address these challenges, Google introduced Accurate Quantized Training (AQT) \cite{lewt2023aqt}, an open-source library in JAX that simplifies QT by providing pre-configured quantization functions and a flexible API for INT8 operations. Primarily designed for Google’s Tensor Processing Unit (TPU) v5e, AQT facilitates efficient INT8 training and inference, reducing training time with minimal accuracy loss. Although a formal research paper on AQT has not yet been published, early versions have been successfully utilized in Google's work with complex models such as transformers \cite{Abdolrashidi2021, zhang2022, ding2023, zhang2023}. While AQT is focused on high-performance hardware, its adaptable design suggests potential applicability to a broader range of hardware, including MCUs, depending on numerical compatibility \cite{lewt2023aqt}.

Furthermore, full INT8 training is emerging as a technique to enable models to train entirely in INT8, but requires substantial modifications to conventional training algorithms to accommodate integer-based gradient descent. Zhu et al. \cite{zhu2020} address the specific challenges of INT8 training for CNNs specifically by managing gradient quantization, as gradients often exhibit high variance and complex distributions, which can destabilize training. They propose two new techniques: Direction Sensitive Gradient Clipping and Deviation Counteractive Learning Rate Scaling to stabilize training by clipping gradients based on quantized-to-full-precision directional deviation and adjusting learning rates to counteract accumulated quantization errors. Their technique uses symmetric quantization, enabling efficient hardware implementation. Moreover, they deployed this approach on low-end Pascal GPUs, and their experimental results show a training time reduction of up to 22\% compared to traditional FP32 training, achieved without significant accuracy loss.

In addition, recent research suggests that the next generation of TinyML devices will likely need the capability to adapt deployed DL models to new data directly in the field \cite{Garofalo2022}. This marks a shift from the traditional paradigm of training models on powerful machines before deploying them on edge devices for inference-only tasks. Retraining models in datacenters using data collected on-field can be costly in terms of latency and power, and inconvenient from privacy and security perspectives \cite{Garofalo2022}. Consequently, there is a growing trend to enable models to adapt by learning from newly sensed data directly on-device, necessitating on-device training capabilities. INT8 training supports this shift by allowing true INT8 deployment and potential on-device or online training, benefiting paradigms such as unsupervised continual learning (CL) and distributed learning scenarios like federated learning (FL).

To support this, Octo \cite{zhou2023towards, zhou2021octo} was introduced as an open-source, lightweight, cross-platform framework aimed at fully supporting INT8 QAT while addressing the computational overhead of fake quantization in traditional QAT. Octo optimizes both forward and backward passes through two key techniques: (1) Loss-aware Compensation (LAC), which adaptively offsets quantization errors based on their impact on the model's loss, and (2) Parameterized Range Clipping (PRC), which controls gradient values dynamically to stabilize quantized gradients and reduce quantization noise. These optimizations streamline INT8-compatible operations throughout the training pipeline, reducing computational complexity while preserving model accuracy. Octo has been deployed on devices such as the NVIDIA Jetson Xavier, which features a dedicated neural processing chip. Although it has yet to be adapted for more constrained embedded devices, Octo's design provides a foundation for efficient on-device or online training in resource-limited settings.

In a similar direction, the authors in \cite{wang2023} propose a gradient distribution-aware INT8 quantization framework, introducing two innovative techniques: (1) the Data-aware Dynamic Segmentation Quantization (DDSQ) scheme and (2) the Update Direction Periodic Search (UDPS) strategy. The DDSQ scheme addresses the challenge of diverse gradient distributions by dynamically adjusting quantization parameters, effectively quantizing both small and large gradients to retain essential information and ensure stable convergence. Meanwhile, the UDPS strategy reduces computational overhead by periodically searching for optimal quantization parameters, thereby maintaining training stability without incurring excessive computational costs. This combined approach achieves minimal accuracy loss compared to floating-point training on models like ResNet, MobileNetV2, and LSTMs.

\subsection{Extreme Low-Bit Quantization}
\label{sec:extreme-low-bit}

While most quantization efforts have focused on reducing numerical precision to 8 bits (INT8), recent research has explored more aggressive strategies such as \textit{binarization} and \textit{ternarization}, where the quantized values are constrained to very small discrete sets. In binarization, weights and activations are typically represented using only two possible values, such as \(\{-1, +1\}\) or \(\{0, 1\}\), which can be encoded using 1 bit. Ternarization extends this idea by allowing a third value (usually 0), resulting in a ternary value set \(\{-1, 0, +1\}\), which can be represented with 2 bits. These techniques aim to push the boundaries of efficiency by minimizing both memory usage and computational cost, often enabling efficient implementation using bitwise operations.

Binarization refers to representing values using only one bit, typically mapping them to \(\{-1, +1\}\) or \(\{0, 1\}\). This approach reduces memory requirements by up to \(32\times\) compared to FP32 representations and enables the use of lightweight bitwise operations (e.g., XNOR and bit-count) in place of costly floating-point arithmetic. Despite these advantages, naive binarization often incurs severe accuracy degradation. To mitigate this, a broad array of methods have been developed~\cite{courbariaux2015binaryconnect, hubara2018, kim2016bitwise, kwon2020structured, wang2020apq, yang2020automatic, jin2020adabits, qin2020forward, zhuang2019structured, wang2019learning, liu2019circulant, zhu2019binary, xu2019main, he2019simultaneously, cai2017deep, guo2017network, juefei2017local, han2020training, qin2020bipointnet, bulat2020high, guo2021boolnet, razani2021adaptive}.

A foundational method in this line of work is BinaryConnect~\cite{courbariaux2015binaryconnect}, which constrains weights to \(\{-1, +1\}\) during training. Real-valued weights are maintained in memory and only binarized during the forward and backward passes using the sign function. Gradient propagation is enabled through the STE, which approximates the derivative of the non-differentiable sign function.

Building on this, Binarized Neural Networks (BNNs)~\cite{hubara2018} extend binarization to both weights and activations. Jointly binarizing weights and activations has the additional benefit of improved latency, since the costly ﬂoating-point matrix multiplications can be replaced with lightweight XNOR operations followed by bit-counting.  BNNs have demonstrated substantial speedups and memory reductions. For instance, studies have reported up to a \(19\times\) speedup and an \(8\times\) memory reduction \cite{bannink2021}. Additionally, Embedded Binarized Neural Networks (eBNNs) have brought BNNs to embedded platforms, such as the Intel Curie on an Arduino, enabling efficient operation on low-power MCUs \cite{mcdanel2017, Geiger2020}.

Another significant contribution is XNOR-Net proposed in~\cite{rastegari2016xnor}, which improves binarization accuracy by introducing a scaling factor \(\alpha\) such that each real-valued weight matrix \( W \) is approximated as \( W \approx \alpha B \), where \( B \in \{-1, +1\} \) is the binarized matrix. The optimal scaling factor is learned by solving:
\begin{equation}
    \alpha, B = \text{argmin} \left\| W -\alpha B\right\|^2.
\end{equation}

Ternarization, another low-bit approach, represents values using \(\{-1, 0, +1\}\), offering a compromise between binary and full precision formats. Inspired by the observation that many learned weights tend to cluster near zero, Ternary Weight Networks (TWNs)~\cite{lin2015neural, li2016ternary} propose quantizing weights to \(\{-w, 0, +w\}\), where \(w\) is a learned or computed scaling factor. Weights whose magnitudes fall below a predefined threshold are set to zero, effectively pruning small contributions while preserving the dominant ones. This ternary quantization enables substantial reductions in both model size and computation, as zero-valued weights eliminate MAC operations during inference.

Ternary-Binary Networks (TBNs)~\cite{wan2018tbn} extend this idea by combining ternary activations with binary weights, thereby achieving a favorable balance between representational capacity and computational efficiency. This hybrid scheme retains much of the performance advantage of binarization, while improving accuracy through the added expressiveness of ternary-valued activations.

Despite the clear benefits, extreme low-bit quantization comes with important limitations. First, the accuracy gap between full precision and sub-8-bit networks is still significant, particularly for complex tasks or deep architectures. Second, most conventional embedded platforms lack support for native sub-8-bit arithmetic, restricting the practical deployment of 1- and 2-bit networks. While techniques such as bit-packing can simulate low-bit operations, they often come with overheads that negate the theoretical gains.

However, this landscape is evolving. Recent MCUs equipped with specialized neural processing units (NPUs), such as the MAX78000 and MAX78002, offer native support for low-bit operations \cite{balbi2025embedded}. These platforms open the door to practical deployment of extreme low-bit quantization in real-world embedded systems. For more on such hardware, see Section~\ref{sec:NPU}.

\subsection{Mixed Precision Quantization}
\label{subsection:mixed-precision}
Within the quantization framework, the standard approach is to apply a uniform bit-width across an entire model. In other words, a model is quantized to $n$-bits homogeneously across all parameters. This approach introduces a trade-off: lower bit-width (e.g., 2 bits) achieves stronger compression of the model at the cost of increased quantization errors (and consequently, drops in performance). Higher bit-widths (e.g., 16-bit) reduce performance degradation, but may limit size and inference acceleration gains. Recently, more advanced quantization methods attempt to address this by finding optimal quantization levels per layer, giving rise to \textit{mixed precision} quantization. This prospect emerges based on the notion that each layer or activation for a model uniquely influences both quantization error, inference time, and overall performance. 

\begin{figure}[ht]
    \centering
    \includegraphics[width=1\linewidth]{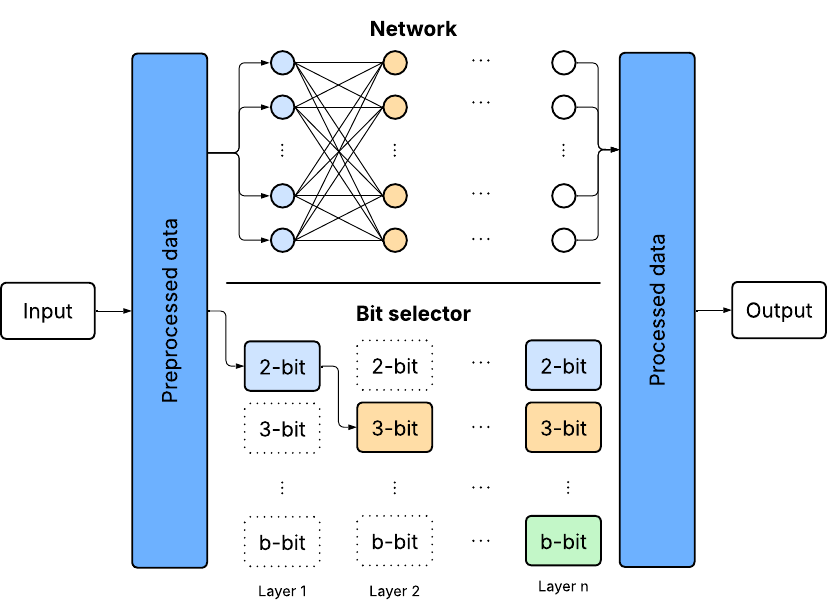}
    \caption{General structure of mixed precision-based frameworks. The network is trained such that each layer is allocated a certain bit-width, typically through some search-based methods of determining optimal bit-widths for performance.}
    \label{fig:mixed-precision-general}
\end{figure}

Mixed precision quantization methods assign variable bit-widths to different layers, channels, or activations based on a calculated ``importance'' factor. A straightforward approach to this would be to assign higher bit-widths to layers with higher importance and lower bit-widths for the less important parameters, effectively meeting at a decisive trade-off point between performance and compression \cite{mazumder2021, htq}. The primary challenge for mixed precision quantization is finding the optimal bit-width for each layer, which is a non-trivial problem due to the size of the search space. 
Broadly, mixed precision strategies can occur both during training, where the bit-widths are learned as the model trains, and post-training. The following subsections discuss the different methodologies that researchers have adopted to determine optimal bit-width allocations for mixed precision quantization. This is structured in two main methodological categories: \textbf{Rough Allocation}, where bit-width allocation is based on partitions or heuristic metrics, and \textbf{Adaptive Allocation}, which emerges from learning-based or differentiable methods that dynamically adapt allocation through model training. A general overview of how mixed precision works is presented in Fig. \ref{fig:mixed-precision-general}.

\subsubsection{Rough Allocation}
\label{subsubsection:rough-allocation}
As mentioned, rough allocation refers to broader partitioning of weight matrices, with each partition allocated a different bit-width. These methods typically fall into the post-training stage. 

To manage the exponential growth of the mixed precision search space with increasing network depth, heuristic-based methods have emerged that estimate each layer’s sensitivity to quantization. A prominent approach is the use of second-order information to guide bit-width assignment. In particular, the Hessian-Aware Quantization (HAWQ) \cite{hawq} framework exemplifies this direction by computing second-order Hessian traces to extract the largest eigenvalues of each layer's weight matrix, showing then that the layers with larger top Hessian eigenvalues are more sensitive. Accordingly, more sensitive layers are assigned higher precision. In a more focused manner, several studies have expanded the use of Hessian-based analyses to identify important weights, specifically for large language models (LLMs). These include works such as CherryQ \cite{cherryq}, which uses the Fisher Information matrix as a surrogate to the full Hessian. The motivation behind this is that computing the full Hessian for LLMs is expensive, given the large number of parameters. Using the Fisher Information matrix is a standard approximation under the assumption of negative log-likelihood loss. 

Other works, such as the proposed Outlier-Aware Quantization (OWQ) \cite{owq} prioritizes a subset of the weights for sensitivity calculations. Similarly, REx \cite{rex} identifies outliers in weight matrices which are quantized to binary values, whilst the remainder of the weights are quantized with higher bit-widths. Although primarily a data-free method (see Section~\ref{sec:data-agnostic}), their framework incorporates the use of mixed precision and is thus included here. This is similar to other works such as \cite{data-free-non-uniform}, which also uses mixed precision under the data-free quantization framework.

Works such as HAWQ-v2 \cite{hawqv2} directly extend the Hessian matrix calculations used in HAWQ to focus on the entire spectrum of the Hessian rather than just the top values. This method uses the average Hessian trajectory to calculate weight importances. This method also introduces better robustness and up to $100\times$ faster search compared to RL-based methods. 

Similarly, the HTQ framework \cite{htq} is designed to indicate perturbations in the performance after quantization. In contrast to HAQ \cite{wang2019haq}, which does not have an explicit importance parameter, HTQ quantifies layer importances before entering the search space to encourage further efficiency in the exploration. Building on the HAWQ methodology, HTQ uses a first-order quantity to estimate layer importances, which is over $1000\times$ faster to compute than the second-order Hessian traces used in HAWQ or similar Hessian-based methods, despite tracking weight sensitivity just as well. 

\subsubsection{Adaptive Allocation}

\label{subsubsection:adaptive}
Adaptive quantization allows for online learning of the bit-widths embedded within the training process itself. In these methods, bit-widths are allocated to layers based on each layer's effect on the loss of the model. Several studies work with novel loss functions that combine both task-specific loss and quantization parameters. Works such as NIPQ \cite{NIPQ} incorporate computational cost and memory consumption into the loss, adjusting the bit-width of each layer dynamically to minimize total loss. In a similar manner, \cite{data-free-non-uniform} incorporates a compression ratio into the loss function, stopping training only when the desired compression is met. 

\begin{figure}[htbp]
    \centering
    \includegraphics[width=0.9\linewidth]{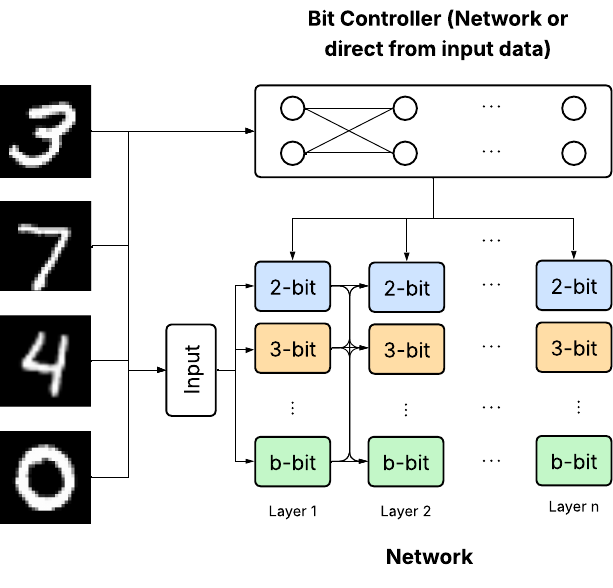}
    \caption{In adaptive allocation, the bits allocated to each layer are determined by a \textbf{Bit-Controller}, which can either be directly through the data itself, or through some degree of meta-training, where an additional network is trained on top of the existing network to ``learn'' the best bit-widths.}
    \label{fig:adaptive-bit-controller}
\end{figure}

Focusing more on the hardware deployment, several strategies optimize the chosen bit-widths based on available hardware budgets. Studies such as \cite{meta-learning} propose a meta-learning approach for QAT. In particular, the meta-learning-based bit-width-adaptive QAT (MEBQAT) framework allows for a single ``meta-trained'' network to be deployed at multiple precision levels without the need to retrain from scratch. This approach allows extended support for multiple bit-widths per layer with minimal loss in performance. To better understand the structure of these frameworks, we refer readers to Fig. \ref{fig:adaptive-bit-controller}. Similarly, MetaMix \cite{kim2024metamix} introduces a two-stage training process that uses a bit-meta-training phase. In this phase, the network is trained across multiple bit-width settings simultaneously to reach a meta-stable state with consistent activation distributions. In the second phase, the meta-trained weights are frozen, and the model learns per-layer bit-width probabilities on that stable base. 

Other works, such as CADyQ \cite{cadyq}, propose a content-aware dynamic quantization framework. This study demonstrated that bit-widths are learned and allocated per-layer and per-image region by training a small predictor network that outputs probabilities for different bit-widths for each activation tensor. At inference, the quantization setting with the highest predicted probability is chosen per layer. This content-driven bit assignment yields higher compression on simpler inputs while preserving accuracy on ``hard'' inputs, though it requires extensive quantization-aware retraining on full data. 

Building on this, CABM \cite{cabm} uses the learned predictor networks' probabilities to construct a lookup table that directly maps an input image's features to an optimal bit-allocation policy. This LUT-based approach refines the runtime bit selection and can better handle larger images, as demonstrated in super-resolution tasks. Like CADyQ, it performs heavy QAT on the full training set to obtain a mixed precision model, trading significant offline cost for improved inference efficiency on edge devices with varying content complexity. Further attempting to reduce retraining costs, AdaBM \cite{hong2024adabm} uses two mapping modules based on a small calibration set rather than the entire training data. One module maps input images to an image-wise bit-width factor and another maps each layer to a layer-wise sensitivity factor. AdaBM achieves comparable accuracy to \cite{cadyq} and \cite{cabm} with an accelerated bit-selection process that is $2000\times$ faster by avoiding full retraining, thereby making it a particularly appropriate framework for microcontroller-class devices.

Other approaches that use multiple models for optimization include the CoQuant framework \cite{coquant}. This work uses a high-precision teacher network to guide a lower-precision student model during training. Specifically, at each training step, certain blocks of the student network at lower bit-widths are dynamically swapped with the corresponding higher-precision blocks from a teacher to provide a form of knowledge distillation. By jointly optimizing all precisions with this selective teacher augmentation, CoQuant achieves a better balance: it significantly improves the accuracy of the lowest-bit layers compared to prior methods that simply treated all bit-widths equally. 

Tackling adaptive allocation from a different perspective, the ALPS framework \cite{langroudi2021alps} replaces FP arithmetic with ultra‑low‑precision generalized Posit numbers. This approach learns, for every layer, the two key Posit hyperparameters: maximum regime length $r$ and the exponent bias $s_c$ so they best match that layer’s weight/activation distribution. The algorithm formulates quantization as a compander model, derives an SQNR‑based upper bound on misclassification, and then chooses $r$ and $s_c$ by minimizing intra‑layer SQNR loss using simple statistics, such as mean and excess kurtosis, rather than an exponential mixed‑precision search, making the adaptation lightweight even for deep nets. For a more comprehensive overview of Adaptive quantization, we refer the reader to the survey by Rakka et al.~\cite{rakka2024review}. 

\subsection{Hardware-Aware Quantization}
\label{sec:hw-aware}

One of the key motivations behind quantization is to reduce inference latency and energy consumption. However, the benefits of quantization are highly dependent on the characteristics of the deployment hardware. Quantization does not yield uniform speed-ups across all platforms, as performance gains are influenced by factors such as memory bandwidth, cache hierarchy, on-chip memory size, and instruction support for low-precision arithmetic. Therefore, quantization strategies must be adapted to the constraints and capabilities of the target hardware to fully realize their efficiency benefits.

This need has led to the emergence of \textit{hardware-aware} quantization, where bit-width configurations are selected not only based on accuracy and model size but also considering hardware-specific metrics such as latency, throughput, and energy. A significant challenge in hardware-aware quantization arises from the enormous search space. Even in standard mixed precision quantization, where each layer in an $L$-layer DNN is assigned one of $S$ candidate bit-widths, the search space scales exponentially as $\mathcal{O}(S^L)$\cite{wang2019haq}. For instance, if $S = 8$ (i.e., candidate bit-widths from 1 to 8 bits) and $L = 50$ as in ResNet-50, the total number of possible bit-width assignments becomes $8^{50}$, which is a prohibitively large space for exhaustive search. The situation becomes even more complex when both weights and activations are quantized independently, increasing the space to $S^{2L} = 8^{100}$. When targeting $H$ hardware platforms, the total search space expands further to $\mathcal{O}(H \times 8^{2L})$\cite{huang2023chip}, underscoring the need for efficient search algorithms tailored to hardware constraints.

To address this, Wang et al.~\cite{wang2019haq} proposed the hardware-aware automated quantization (HAQ) framework, which uses reinforcement learning (RL) to automate the selection of per-layer bit-widths for weights and activations. A hardware simulator provides feedback to the RL agent about resource consumption, such as latency and energy, under various quantization schemes. Deployments on a Xilinx Zynq-7020 FPGA showed latency reductions of 1.4--1.95$\times$ and energy savings of 1.9$\times$ compared to uniform INT8 quantization. However, HAQ uses non-uniform quantization formats, which are incompatible with the integer-only inference required on MCUs.

To address MCU constraints, Rusci et al.~\cite{rusci2020} introduced a hardware-aware quantization approach using uniform quantization for both weights and activations, targeting platforms like the STM32H7 (512 KB RAM, 2 MB Flash). Their framework restricts the search space to integer-only operations and adheres to the memory and compute limits of typical edge devices. Using MobileNetV2, their approach achieved 68.4\% accuracy with mixed precision quantization—outperforming all 8-bit models that could fit within the same memory budget.

In another line of work, HAWQ-v3~\cite{yao2021hawq} explored deploying quantized models directly to hardware during search. Instead of relying solely on theoretical estimates like FLOPs, HAWQ-v3 measures actual per-layer latency on GPUs for different bit-widths and uses Integer Linear Programming (ILP) to find an optimal configuration under resource constraints such as model size or runtime. While this provides more accurate latency profiling than analytic models, the measurements are still affected by external software and system-level overhead, which limits their fidelity in reflecting true on-chip efficiency.

A notable recent advancement is On-Chip Hardware-Aware Quantization (OHQ)~\cite{huang2023chip}, which performs the entire quantization process directly on edge devices without requiring powerful external computing resources. OHQ introduces a mask-guided quantization estimation (MGQE) technique to assess the impact of bit-width choices on accuracy using only on-chip resources. The quantization search is performed on a dual-core ARM Cortex-A9 processor with 512MB of DDR3 memory, using bit-widths from 4 to 8. Unlike prior methods that rely on offline simulations on GPUs or clusters, OHQ avoids discrepancies between estimated and actual hardware performance and removes dependence on high-end machines. The results demonstrate that OHQ achieves state-of-the-art trade-offs in accuracy and efficiency, while being entirely self-contained and deployable on-device.

\subsection{Redistribution}
\label{sec:redist}

Quantization often introduces accuracy degradation due to information loss, particularly when dealing with non-uniform or heavy-tailed distributions of weights and activations. A central strategy to mitigate these effects involves redistributing these values in a manner that makes them more amenable to low-bit representation.

Redistribution techniques operate by adjusting the statistical structure of the data, either globally or locally, such that quantization error is reduced, outlier influence is minimized, and quantization bins are better utilized. The essence of redistribution lies in modifying the weight or activation distributions to either flatten them, compact their range, or smooth discontinuities, thereby preserving task performance under aggressive quantization.

The following subsections explore several redistribution strategies, beginning with full distribution uniformization, and continuing with more localized reshaping, outlier suppression, and bias-based shifts.

\begin{figure}[htbp]
    \centering
    \begin{subfigure}{0.8\linewidth}
        \centering
        \includegraphics[width=\linewidth]{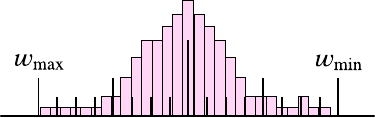}
        \caption{Original weight distribution of a model layer. The values follow a zero-centered, Gaussian-like distribution, ranging between \( w_{\min} \) and \( w_{\max} \).}
        \label{fig:Gaussian}
    \end{subfigure}
    
    \vspace{1em}
    
    \begin{subfigure}{0.8\linewidth}
        \centering
        \includegraphics[width=\linewidth]{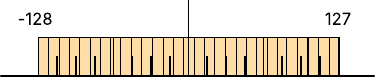}
        \caption{Weight distribution after uniformization. The weights are transformed to follow a flat uniform distribution over the target quantization range, i.e., \([-128, 127]\) in this case for 8-bit signed integers.}
        \label{fig:Uniform}
    \end{subfigure}
    
    \caption{Illustration of weight uniformization for quantization. (a) Original weights typically follow a Gaussian distribution. (b) After uniformization, the weights are redistributed uniformly to reduce quantization error.}
    \label{fig:Uniformization}
\end{figure}

\subsubsection{Distribution Uniformization}

Uniform quantization, favored for its hardware efficiency and implementation simplicity, is often suboptimal when applied to neural networks (see Section~\ref{sec:uniform_non-uniform}). This is largely due to the inherently non-uniform distributions of weights and activations, which commonly follow Gaussian- or Laplace-like shapes concentrated near zero. As a result, higher quantization bins are frequently underrepresented, leading to poor bit utilization and elevated quantization error~\cite{yu2020low, chmiel2020robust, li2024exploring}. In contrast, more uniformly distributed tensors have been shown to exhibit higher signal-to-noise ratios and reduced sensitivity to quantizer implementation details~\cite{chmiel2020robust}.

To mitigate this mismatch, recent studies have explored distribution uniformization techniques, which reshape the distributions of weights and activations toward a more uniform form during training. As illustrated in Fig.~\ref{fig:Uniformization}, the original Gaussian-like distribution (Fig.~\ref{fig:Gaussian}) is transformed into a uniform distribution (Fig.~\ref{fig:Uniform}), allowing quantization levels to be used more effectively and reducing quantization error. This reshaping facilitates more effective utilization of quantization levels and reduces information loss under low-bit quantization regimes.

For instance, Yu et al.~\cite{yu2020low} proposed the \textit{Group-based Distribution Reshaping Quantization} (GDRQ) framework, which integrates the Scale-Clip technique to reshape distributions and group-based quantization to allow localized quantization parameter adaptation. Their method achieves near state-of-the-art performance with as low as 2-bit weight and 4-bit activation quantization.

Alternatively, \textit{Kurtosis Regularization} (KURE)~\cite{chmiel2020robust} introduces an auxiliary loss term that penalizes the fourth central moment (kurtosis) of the model's weight and activation distributions. This regularization encourages tensor distributions to avoid heavy tails and instead adopt more controlled, light-tailed behaviors that are favorable for quantization. Specifically, KURE reshapes distributions to reduce sensitivity to extreme values, aiming to produce a \textit{bit-width-agnostic} model that performs robustly under different quantization policies.

The overall objective is a multi-objective loss function defined as:
\begin{equation}
\mathcal{L}_{\text{total}} = \mathcal{L}_{\text{task}} + \lambda \cdot \mathcal{L}_{K},
\end{equation}
where $\mathcal{L}_{K}$ denotes the kurtosis regularization term and $\lambda$ is a tunable hyperparameter that balances task accuracy against distribution regularity. The authors experiment with various $\lambda$ values to study this trade-off, and demonstrate that modest kurtosis penalization significantly improves quantization robustness in both PTQ and QAT. Notably, KURE achieves strong performance even under aggressive quantization (e.g., 3-bit), without the need for bit-specific re-training.

In Str{\"o}m et al.'s~\cite{strom2022squashed} \textit{Squashed Quantization} (SqWQ), they propose a training-time reparameterization technique to reshape the typically Gaussian-distributed weights into a more uniform distribution, better suited for linear quantization. Specifically, each weight is expressed as \( W = \tanh(P) \), where \( P \sim \mathcal{N}(0, \sigma^2) \), effectively squashing the weights into a bounded interval. This transformation reduces range outliers, leading to a more even utilization of quantization bins.

To preserve the target weight distribution during training, the authors introduce a regularization term that enforces the input parameters \(P\) to match a fixed Gaussian distribution. The target standard deviation \(\sigma_t\) is derived by minimizing the KL divergence between the squashed output and a uniform distribution. The result is a model with improved quantization efficiency, especially in low-bit settings (e.g., 2--3 bits), with minimal accuracy loss compared to traditional quantization methods.

EQ-Net \cite{xu2023eq} further generalizes distribution regularization across an elastic quantization space (varying bit-width, symmetry, granularity). They introduce the \textit{Weight Distribution Regularization Loss (WDR-Loss)} to penalize skewness and kurtosis. The multi-objective loss is linearly combined with task performance using tunable coefficients, providing flexibility in adapting to diverse quantization hardware.

Finally, KurTail \cite{akhondzadeh2025kurtail} applies a learnable kurtosis-based rotation to suppress heavy-tailed outliers in activations of large language models (LLMs). It uses layer-wise optimization and PTQ to reshape activation distributions, effectively enabling 4-bit quantization across weights, activations, and KV cache. Like others, it employs a lambda-weighted loss formulation to balance kurtosis minimization and task accuracy.

\subsubsection{Distribution Reshaping}
While distribution uniformization encourages flat, structureless weight profiles, an alternative strategy is to reshape distributions into forms more aligned with quantization behavior. These methods prioritize either enhancing bin-wise localization or limiting the spread of weights to better utilize low-bit quantization bins.

\textit{Bin Regularization (BR)}~\cite{han2021improving} approaches this by refining the internal structure of each quantization bin: it encourages weights within a bin to concentrate sharply around its quantized center. The method introduces an auxiliary loss that simultaneously minimizes the mean distance to the bin center and reduces within-bin variance. This effectively drives the distribution in each bin toward a Dirac delta-like peak. Like KURE, BR adopts a multi-objective linear loss formulation:
\begin{equation}
\mathcal{L}_{\text{total}} = \mathcal{L}_{CE} + \lambda \cdot \mathcal{L}_{BR},
\end{equation}
where $\mathcal{L}_{CE}$ is the standard cross-entropy loss and $\lambda$ balances task performance with quantization alignment. Despite outperforming KURE at lower bit-widths (e.g., 2–3 bits), BR introduces sensitivity to bin selection heuristics, which can affect model robustness.

\textit{Range Restriction Loss (R\textsuperscript{2})}~\cite{kundu2023r2} complements this idea from a different angle. Rather than emphasizing intra-bin compactness, R\textsuperscript{2} reduces the global weight range by explicitly penalizing outliers during pretraining. This leads to a narrower, denser weight distribution that remains non-uniform (e.g., a Gaussian mixture), thus retaining expressiveness while improving compatibility with downstream quantization. It serves as a preparatory step before PTQ or QAT and is introduced via an auxiliary loss:
\begin{equation}
\mathcal{L}_{\text{total}} = \mathcal{L}_{\text{task}} + \lambda \cdot \mathcal{L}_{R^2}.
\end{equation}
Three variants are proposed: (1) L1 R\textsuperscript{2} Loss, which simply penalizes the L1 norm of outlier magnitudes, (2) Margin R\textsuperscript{2} Loss, which defines a soft threshold beyond which weights are penalized, and (3) Soft-Min-Max R\textsuperscript{2} Loss, which replaces hard range enforcement with differentiable approximations to min and max operations. The choice of variant depends on whether the goal is symmetric quantization (favoring L1 or Margin) or compression-specific preconditioning (favoring Soft-Min-Max). Empirical results show significant gains in accuracy and quantization compatibility across 1–2 bit settings for ResNet and MobileNet architectures.

\subsubsection{Outlier Redistribution}\label{Outlier Processing}

In contrast to approaches that reshape entire distributions, a distinct line of research focuses on mitigating the disproportionate impact of outliers--extreme values that significantly degrade quantization performance. Such outliers, prevalent in both weights and activations, especially in large-scale models, can cause uniform quantizers to allocate resolution inefficiently, thereby reducing the fidelity of the bulk value representation.

To address this issue, SmoothQuant~\cite{xiao2023smoothquant} introduces a channel-wise scaling strategy applied to both activations and weights. For each channel, a positive smoothing factor is computed from activation statistics. Activations are divided by this factor, while the corresponding weights are multiplied by it, preserving the computation while shifting quantization difficulty from activations to weights.

Building on this principle, Outlier Suppression+ (OS+)~\cite{wei2023outlier} adds a per-channel offset and scaling to further reduce outlier effects, while OmniQuant\cite{shao2024omniquant} learns both parameters during training. FQ-ViT~\cite{lin2021fq} instead applies a channel-specific power-of-two scaling inside LayerNorm to dynamically adjust quantization ranges.

Despite its effectiveness, SmoothQuant does not redistribute outliers across non-outlier channels. Recent studies show that Hadamard transforms provide a more effective smoothing mechanism~\cite{ashkboos2024quarot,lin2024qserve,xi2023training,tseng2024quip}. Unlike channel-wise scaling, which only adjusts individual channels, Hadamard matrices perform orthogonal rotations that evenly spread outliers across all channels. This reduces their overall impact by increasing incoherence between weights and the Hessian matrix. Due to their orthogonality, the transformation preserves the original computation and allows the rotated weights to be preprocessed offline, avoiding runtime overhead.

DuQuant~\cite{lin2024duquant}  extends the rotation concept by targeting outlier-heavy channels, using block-wise rotations and zigzag permutations to balance variance across blocks, followed by an additional rotation to smooth activations. Other works employ \emph{clamping-based redistribution}. For example, AdderQuant~\cite{nie2022redistribution} limits weights to a fixed range and adds a bias to offset lost information.  MagR~\cite{zhang2024magr} adjusts weights in a preprocessing step by solving an optimization problem that trades off output accuracy against maximum weight magnitude, thereby systematically suppressing outliers while maintaining performance.

\subsubsection{Bias Redistribution}\label{Bias}

Complementary to scaling and rotation techniques, bias-based methods adjust the central tendency of data distributions to better align them with quantization intervals, typically through additive shifts. By repositioning the mean of the distribution, these methods can reduce quantization error and improve representational efficiency.

One representative approach is QuantSR~\cite{qin2023quantsr}, which introduces a learnable mean-shift parameter to dynamically reposition the distribution during quantization. This ensures that bin boundaries are better aligned with the actual data. In addition, QuantSR employs a transformation function to handle values far from the quantization center, further refining the representation of extreme values.

Building on the idea of modifying distributions before quantization, NoisyQuant~\cite{liu2023noisyquant} targets heavy-tailed activation distributions in vision transformers (ViTs). Instead of learning a shift, it adds a fixed uniform noise bias, which smooths the distribution and reduces the influence of outliers.

Extending this principle further, RAOQ~\cite{zhang2024reshape} introduces a learnable shift to move activation values away from zero, thereby increasing their variance and making more effective use of the quantization range. This is achieved through a clip-and-recenter operation that spreads values more evenly according to the bit precision, reducing the likelihood of activation clustering and improving quantization efficiency.

\begin{figure*}[htbp]
    \centering
    \includegraphics[width=0.80\textwidth]{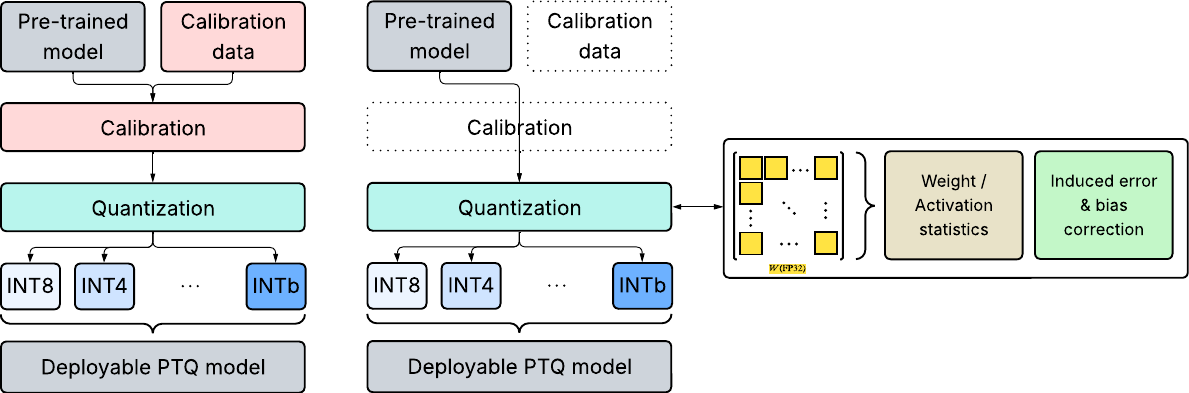}
    \caption{Illustration of Calibrated PTQ vs. Data-free PTQ. In data-free PTQ, the quantization is done based solely on extracted weight and activation statistics of the network itself, without any data ``guiding'' the scheme.}
     \label{fig:Data-Free}
\end{figure*}

\subsection{Data-Agnostic Quantization}
\label{sec:data-agnostic}

Typically, both QAT and PTQ approaches to quantization require some fine-tuning during the quantization process, usually in the form of data calibration. This is especially relevant for post-training frameworks. Calibration is essential for accurate, high-performing quantization because it allows for the collection of basic parameters such as the scaling factors, zero-points, minima and maxima of the distributions, and more. This is necessary for mapping continuous weights and activations to coarse, discrete integer values without causing degradations in performance. Since the distributions of the weights and activations are data dependent, the process of calibrating involves using a small representative set of the full data during a forward pass of the model. The collected statistics are then used to minimize the quantization error by choosing thresholds and grid values that retain the most important information. Without calibration, quantization can suffer from excessive rounding or clipping, especially in layers with outliers or skewed distributions, leading to poor model performance.

However, model calibration assumes complete access to the data, which is not necessarily the case. Data access may be restricted due to concerns regarding privacy or confidentiality. Consequently, data-free quantization methods have emerged in recent years as a powerful alternative.  These methods are particularly useful in on-edge deployment cycles, where data collection and access are typically decentralized. Fig.~\ref{fig:Data-Free} contrasts the standard calibrated PTQ process with data-free PTQ.

As mentioned, data-agnostic methods do not use any representative data for calibration, instead relying on exploiting model weights and activations directly to minimize quantization-induced errors. Works such as \cite{nagel2019data} introduce the Data-Free Quantization (DFQ) framework, which effectively quantizes models from full 32-bit floating-point precision down to INT-8 without any calibration or finetuning. Their methodology can be broken down into three parts, starting with cross-layer range equalization to balance weights across layers. Authors leverage the scale-equivariance property of activation functions (particularly ReLU) to ensure that all layers have similar ranges in terms of their outputs, eliminating any outliers that may occur. This may lead to larger biases within the model, which leads to the second stage of the framework. In this stage, DFQ ``absorbs'' part of these biases into the next layer to prevent large activation ranges, which would otherwise reduce quantization precision. Finally, the framework adds one final tier of correction through analytically estimating the expected output error using batch normalization parameters and adjusting layer biases. 

Other works, such as SQuant \cite{squant}, employ similar Hessian-based analyses as those presented in Section \ref{subsection:mixed-precision}. In their framework, diagonal Hessian approximations are used on a layer-by-layer basis. Their approximation splits the layer-wise error term into three mutually independent components aligned with the weight tensors' geometry: element-wise, kernel-wise, and output-channel-wise. For each independent component, they formulate a data-free discrete optimization step that minimizes a new metric called Constrained Absolute Sum of Error (CASE). This is a simple count of signed quantization errors bounded by $\pm0.5$ per element, $\pm0.5$ per kernel, and $\pm0.5$ per channel. On the other hand, hybrid methods such as REx \cite{rex} use mixed precision quantization (see Section \ref{subsubsection:rough-allocation}) whilst also quantizing residual errors that are accumulated during the quantization process itself. In this work, instead of opting for a single, low-bit representation, the authors focus on successive low-bit residuals that, when added together, reconstruct the full precision weights with provable error bounds. Since every step of the approximation relies directly on the model parameters themselves, the method is concretely data-agnostic. In particular, the REx methodology follows the following structure: In the first pass, standard PTQ is applied to the model and the weights are approximated through:
\begin{equation}
    R^{(1)} = Q^{-1}(Q(W))
\end{equation}
with $Q$ being the quantizer function, $Q^{-1}$ being the ``de-quantizer'' and $W$ being the weights. Then, this approximation is recursively expanded as: 
\begin{equation}
    R^{(k)} = Q^{-1}\left(Q(W - \sum_{k=1}^{K-1} R^{(k)})\right)
\end{equation}
Inference is done based only on $\sum_{k=1}^{K-1}R^{(k)}x$ rather than $Wx$, invoking the quantized weights and relying only on the weight matrices' statistics rather than on calibrated data. 

The Parametric Non-uniform Mixed precision Quantization (PNMQ) \cite{data-free-non-uniform} framework works analogously to REx, relying only on models' weights for quantization. The method uses a non-uniform quantization grid denoted as $G_p^n = \{x_i\}_{i=0}^{2^n-1}$. Each grid point depends on a non-uniformity scaling parameter, $p\in[1,2]$. For each layer, given a bit-width $n$, the method solves: 
\begin{equation}
\min_{s,p} \left\lVert 
W - s \cdot \left\lfloor \frac{W}{s} \right\rceil_p^n 
\right\rVert,
\end{equation}
with $W$ being the original weight tensor, $s$ being the quantization scale parameter, and $\lfloor\cdot\rceil^n_p$ rounding to the non-uniform grid. After determining the optimal $s$, $p$, and $n$ (mixed precision) for each layer's weights, the methodology from \cite{nagel2019data} is applied to minimize the quantization-induced output drift. Bai et al. \cite{udfc} also implement a data-free quantization scheme with their proposed Unified Data-Free Compression (UDFC) methodology, in which quantization-induced errors are compensated by combining other channels. Their formulation is based on the assumption of linear recoverability: Any ``damaged'' channel, either removed by structured pruning or coarsely quantized, can be approximated as a weighted linear combination of the remaining full‑precision channels in the same layer. Leveraging this, they derive a layer‑wise reconstruction form that rewrites the next layer’s convolutional kernels so that the signal originally carried by the lost channel is redistributed to its neighbors via optimal scale factors $s$. They then formulate a closed‑form, convex reconstruction error to recover lost information from compression.

For a more comprehensive overview of data-agnostic quantization, we refer the reader to the survey by Kim et al.~\cite{kim2025zero}.

\section{Alternative Numerical Representations}
\label{sec:alternative_formats}

While the techniques discussed so far focus on improving quantization within standard numerical formats, an orthogonal research direction investigates fundamentally different number systems. These \textit{alternative numerical representations} aim to overcome key limitations in dynamic range, precision allocation, hardware compatibility, and energy efficiency that constrain traditional fixed- and floating-point formats.

As mentioned in Section~\ref{sec:representations}, the core challenge is to approximate 32-bit floating-point numbers using fewer bits while maintaining numerical fidelity and enabling efficient hardware implementation. This trade-off becomes especially critical in constrained environments such as microcontrollers and edge devices. While formats like FP16 are widely supported by modern FPUs, they may not offer the optimal balance between accuracy, performance, and computational cost. Fixed-point and integer arithmetic, though hardware-efficient and simple, often suffer from a limited dynamic range and reduced expressivity—making them less suitable for deeper or more sensitive models. As a result, new number systems have emerged that better match the distributional characteristics of DNN weights and activations, aiming to improve both memory and compute efficiency.

In recent years, both academia and industry have actively explored new numeric formats. Major hardware vendors are investigating domain-specific representations tailored to their inference workloads~\cite{ma2021specializing, vogel2018efficient}. Notable examples include Google's BFloat16 (BF16)~\cite{guntoro2020next}, NVIDIA's TensorFloat-32 (TF32)~\cite{choquette2021nvidia}, Microsoft's Microsoft Floating Point (MSFP) \cite{darvish2020pushing} as well as Tesla's Configurable
Float8 (CFloat8) and Float16 (CFloat16) \cite{tesla_dojo_2025}. These formats revisit floating-point design to offer broader dynamic range or better precision scaling under hardware constraints.

The following subsections review advances across three major families: floating-point variants, improved fixed-point schemes, and tapered-accuracy representations such as Posit.

\subsection{Floating-Point Formats}
\label{sec:FP-based}

Traditional floating-point (FP) representation encodes real numbers using a sign bit, an exponent, and a mantissa. A general form of this representation can be written as:
\begin{align}
n=\operatorname{sign} \times 2^{\text {exponent value}} \times \text{mantissa value}.
\end{align}

The widely used FP32 format allocates 8 bits to the exponent and 23 bits to the mantissa, enabling a wide dynamic range suitable for a broad range of numerical tasks. However, this range often exceeds what is needed in DNNs, particularly during inference, leading to low information-per-bit density and unnecessary costs in power, area, and latency~\cite{leon2021improving}.

To mitigate this, reduced-precision variants have been introduced. FP16~\cite{kahan1996ieee}, for example, uses a 5-bit exponent and a 10-bit mantissa, while Google’s Brain Float 16 (BF16)\cite{intel2018bf16} retains the 8-bit exponent of FP32 but reduces the mantissa to 7 bits. NVIDIA’s TensorFloat-32 (TF32)\cite{choquette2021nvidia} also uses 8 exponent bits but limits the mantissa to 10 bits, resulting in a 19-bit format. These formats preserve the dynamic range required for DNN workloads while reducing bit-width, leading to faster computation and improved memory efficiency. They are now widely adopted across training accelerators, including those from Google, NVIDIA, and Intel.

Researchers have also explored the theoretical foundations of FP bit allocation. For an FP format with a total bit-width of \(N\), the bit-lengths of the mantissa (\(n_1\)) and exponent (\(n_2\)) can be optimized by analyzing propagation of quantization error. With one bit reserved for the sign, the relationship becomes \(n_1 = N - n_2 - 1\). By differentiating the quantization error function and setting the result to zero, the optimal allocation of bits between mantissa and exponent can be derived. NVIDIA’s TF32, for example, follows such a rationale, and optimal trade-offs between exponent and mantissa can be derived by minimizing expected error through analytical means~\cite{chmiel2020neural}.

More aggressive reductions have targeted inference use cases, including 8-bit floating-point formats~\cite{wu2021low, kang2018short}, which show that with proper QAT, competitive accuracy can be achieved. In practice, such formats are typically used for storing weights and activations, while higher-precision (e.g., FP32) is retained for accumulations and weight updates to avoid numerical degradation~\cite{wang2019bfloat16,kang2018short,narang2018mixed}.

In addition to reducing bit-width, several novel FP-based systems aim to enhance compression or adaptability. Microsoft’s MSFP~\cite{darvish2020pushing} format shares a single exponent across a group of values—known as a ``bounding box", allowing smaller per-value mantissas and enabling Single Instruction Multiple Data (SIMD)-style acceleration. Block and Subword-Scaling Floating-Point (BSFP)~\cite{lo2023block} builds upon this by representing each full precision weight as a combination of two low-bit sub-word vectors, each scaled independently by low-bit floating-point factors. This dual-scale mechanism captures both coarse and fine variations in weight distributions.

Other adaptive formats focus on structural optimizations. AdaptiveFloat~\cite{tambe2020algorithm} dynamically sets the exponent bias per tensor to minimize quantization error, while the adaptive numerical
data type (ANT)~\cite{guo2022ant} infers the exponent-mantissa boundary from the bitstream itself, based on the first occurrence of 1. Dynamic-Length Float (DFloat)~\cite{zhang202570} identifies redundancy in BF16’s exponent representation and applies Huffman coding to compress it losslessly, yielding highly compact encodings (e.g., 11-bit representations for LLM weights).

Together, these innovations represent a spectrum of floating-point alternatives that balance accuracy, compression, and hardware efficiency, significantly expanding the design space for quantization-aware neural networks.

\subsection{Fixed-Point Formats}
\label{sec:FX-based}
Fixed-point representations, most commonly realized as integer quantization (e.g, INT8), remain a practical choice for efficient DNN inference due to their hardware simplicity and low memory footprint. However, conventional low-precision fixed-point suffers from limited dynamic range and bias around zero, making it best suited for narrow-support distributions and often requiring careful calibration to avoid overflow or underflow. To overcome these limitations, several alternative fixed-point schemes have been proposed that better align numerical resolution with the statistics of the target data while retaining the efficiency of integer arithmetic.

A notable generalization of standard fixed-point is TFLite's Zero-Skew format~\cite{jacob2018quantization}, a uniform fixed-point representation with zero-centered bins. In this scheme, a real number $n$ is represented as $n = s(q - z)$ where $s$ is the \textit{skew} (scale factor), $z$ is the zero-point. The pair $(s,z)$ are the same quantization parameters defined in Section~\ref{sec:quantization}. This zero alignment shifts the quantization grid to match the distribution mean, reducing bias and improving accuracy for asymmetric data. 

Beyond grid alignment, adaptability in fractional precision offers further gains. F8Net~\cite{jin2022f8net} maintains a fixed bit-width but determines the position of the fractional boundary separately for each layer based on the statistical properties of its activations or weights. By analyzing the standard deviation of the target tensor, F8Net selects a precision setting that ensures the representable range is well matched to the layer’s value distribution. This per-layer adjustment minimizes saturation and underutilization of the numeric range, thereby reducing quantization error while preserving compatibility with efficient integer computation.

Complementary to per-layer adaptation, VS-Quant~\cite{dai2021vs} increases scale granularity through a hierarchical design that combines coarse per-channel scaling with lightweight per-vector factors. This finer partitioning locally reduces dynamic range, improving low bit-width accuracy, particularly in PTQ, while keeping metadata and computation overhead minimal. The vector-level granularity aligns with typical accelerator MAC tile sizes, preserving efficient fixed-point execution.

Collectively, these approaches extend the applicability of fixed-point quantization by addressing its core limitations through zero alignment, adaptive precision, and granular scaling, enabling accurate and efficient inference on highly resource-constrained hardware.

\subsection{Posit Format}
\label{sec:posit}

The \textit{Posit} number system, also known as Type III Universal Number (Unum)~\cite{gustafson2017beating, gustafson2, carmichael2019deep, lu2020evaluations}, has been proposed as a compact alternative to IEEE-754 floating-point (FLP) for efficient numerical representation in deep learning. It offers superior dynamic range and precision with the same or fewer bits than traditional FLP, using a format designed to allocate bits more effectively~\cite{cococcioni2019fast, romanov2021analysis}.

Posit numbers are parameterized by $(w, es)$, where $w$ is the total bit-width and $es$ is the number of exponent bits. Internally, a Posit number is composed of four fields: sign bit, variable-length regime, exponent (of fixed length $es$), and a variable-length mantissa. The regime field enables Posit's \textit{tapered accuracy}---a core feature where more precision is allocated to values close to 1 and less to extreme values~\cite{langroudi2021alps}. This is especially well-suited to deep neural networks, where parameters and activations typically follow zero-centered Gaussian-like distributions~\cite{lu2020evaluations}.

The numerical value of a real number $n$ is represented in the Posit format (by $\hat{n}$ ) as follows:
 \begin{equation} \label{Posit_format}
\hat{n} =(-1)^s \times u^k \times 2^e \times (1+\frac{m}{2^{ms}}),
\end{equation}
where \( s \) is the sign bit, \( e \) the exponent, \( m \) the mantissa, and \( u = 2^{2^{es}} \) is the so-called useed. The regime value \( k \) is computed based on the number of repeated bits in the regime field:
\begin{align}\label{kkk}
k =
\begin{cases}
-d, & \text{if } r = 0 \\
d-1, & \text{if } r = 1
\end{cases}
\end{align}
Unlike FLP, which maintains fixed-length fields, the Posit dynamically adjusts the mantissa length based on the magnitude of the number. This leads to higher relative precision near 1 and lower near the extremes. Because of this precision-profile and dynamic range efficiency, Posit is a strong candidate for low-precision DNN inference, especially in resource-constrained environments~\cite{carmichael2019deep, gustafson2}. Posit arithmetic has been shown to run effectively on MCUs with minimal accuracy loss~\cite{cococcioni2019fast} and is also being explored for approximate computing at the edge~\cite{guntoro2020next}.

For a comprehensive overview of emerging number representations supporting DNN architectures (including hybrid formats and Posit variants), we refer the reader to the survey by Alsuhli et al.~\cite{alsuhli2023number}. For additional insight into efficient computer arithmetic for edge computing, particularly hardware-oriented approximate computing, see the work by Guntoro et al.~\cite{guntoro2020next}.

\section{Hardware Landscape}
\label{sec:hardware}

When it comes to deploying DNNs on the edge, particularly within embedded systems, there exists a wide spectrum of hardware platforms with varying capabilities. These range from relatively powerful microcomputers or single-board computers, such as the Raspberry Pi family, which offer advanced processors and generous memory and storage, to ultra-low-power MCUs, which sit at the ``extreme edge'' due to their tight constraints on power, memory, and compute. In this survey, we focus specifically on MCUs. Below, we present the current MCU landscape, categorized by architecture: ARM-based, RISC-V-based, and recent developments in hybrid systems, including those that integrate NPUs or dedicated cores for DNN/CNN acceleration.

\subsection{ARM-based MCUs}
\label{sec:arm}

\begin{table*}[t]
    \centering
    \caption{Specifications of Common ARM‐based MCUs}
    \label{tab:mcu_specs}
    \vspace{-0.1cm}
    \footnotesize
    \renewcommand{\arraystretch}{1.5} 
    \begin{tabular}{lllllll}
        \toprule
        \textbf{Edge Device} & \textbf{MCU} & \textbf{CPU(s)} & \textbf{Clock} & \textbf{Flash} & \textbf{RAM} & \textbf{FPU} \\
        \midrule

        \makecell[l]{Adafruit Circuit Playground Bluefruit} & nRF52840 & Cortex-M4 & 64 MHz& 1 MB & 256 KB & N/A \\

        \makecell[l]{Arduino Nano 33 BLE Sense} & nRF52840 & Cortex-M4 & 64 MHz& 1 MB & 256 KB & N/A \\

        Raspberry Pi Pico & RP2040 & \makecell[l]{Cortex-M0+} & \makecell[l]{133 MHz} & 2 MB & 264 KB & \makecell[l]{N/A} \\

        SparkFun Edge & Ambiq Apollo3 B & Cortex-M4F & 48 MHz& 1 MB & 384 KB & Single-Precision \\

        ST Nucleo-L452RE & STM32L452RE & Cortex-M4F & 80 MHz& 512 KB & 160 KB & Single-Precision \\

        Sony Spresense & CXD5602 & Cortex-M4F & 156 MHz& 8 MB & 1.5 MB & Single-Precision \\

        STM32F746 Discovery Kit & STM32F746NG & Cortex-M7 & 216 MHz& 1 MB & 320 KB & Double-Precision \\

        OpenMV Cam H7 & STM32H743VI & Cortex-M7 & 480 MHz& 2 MB & 2 MB & Double-Precision \\

        OpenMV Cam H7 Plus & STM32H743II & Cortex-M7 & 480 MHz& \makecell[l]{2 MB,\\32 MB} & \makecell[l]{1 MB,\\32 MB} & Double-Precision \\

        Arduino Portenta H7 Lite & STM32H7 & Cortex-M7 & 480 MHz& 2 MB & 1 MB & Double-Precision \\

        Google Coral Dev Board Micro & NXP i.MX RT1176 & \makecell[l]{Cortex-M4F,\\Cortex-M7} & \makecell[l]{400 MHz,\\800 MHz} & 128 MB & 64 MB & \makecell[l]{Single-Precision,\\Double-Precision} \\
        \bottomrule
    \end{tabular}
\end{table*}

In the realm of available MCUs worldwide, the Arm Cortex-M Series, particularly the Cortex-M4, stands as the most popular processor \cite{novac2021quantization}. Billions of microcontrollers are built around this processor, making it an industry leader in 32-bit MCUs \cite{arm2021}. These cores are optimized for low-cost and energy-efficient integrated circuits. A Cortex-M4 microcontroller typically costs between 5–10 USD and can operate on a coin-cell battery for months, if not years \cite{saha2022}.

The Cortex-M4 is a 32-bit reduced instruction set computer (RISC) processor designed for energy efficiency, based on the ARMv7E-M instruction set architecture (ISA) \cite{cmsis}. It features a core register bank with 16 32-bit registers and a built-in digital signal processing (DSP) module that supports vectorized parallel operations, such as MAC and SIMD. It can handle 16/32-bit MAC and 8- or 16-bit SIMD arithmetic in one clock cycle. This is particularly important in DNNs, where MAC operations, multiplying inputs by weights and accumulating results to form neuron activations, are parallelized using SIMD to improve efficiency.

\begin{figure}[ht]
    \centering
    \includegraphics[width=0.40\textwidth]{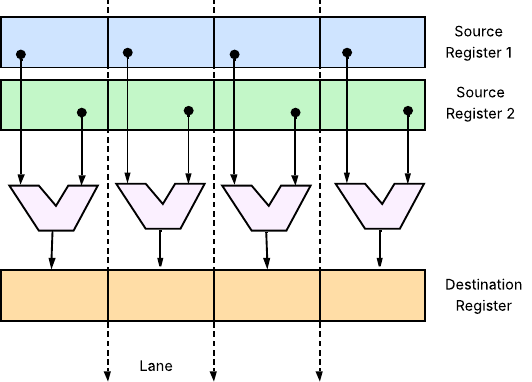}
    \caption{SIMD execution of packed data}
     \label{fig:6}
\end{figure}

In SIMD, data is stored in vector form, with each segment in a SIMD register referred to as a ``lane" (see Fig.~\ref{fig:6}). Each lane holds part of the vector data (e.g., 8-bit or 16-bit). A 32-bit register can thus be divided into four 8-bit lanes or two 16-bit lanes. SIMD instructions apply a single operation (e.g., Add, Multiply) to all lanes in parallel within one clock cycle; for example, an ADD instruction adds each pair of 8-bit values across lanes simultaneously. The degree of parallelism is limited by the register’s bit width. Organizing multiple values into lanes for full register utilization is called packing—for instance, combining four 8-bit values into one 32-bit register. After SIMD operations, data is often unpacked or rearranged to align for subsequent processing, ensuring consistent flow and efficient processor use.

Furthermore, the Cortex-M series includes models with varying power profiles, many lacking an FPU. In such cases, floating-point operations are emulated in software, creating significant overhead, as they require more processor cycles than integer operations. FPUs, when present, greatly accelerate these tasks; for example, the Cortex-M4F features a single-precision (32-bit) FPU with 32 dedicated registers, also addressable as 16 double-word registers. Table~\ref{tab:mcu_specs} lists representative Cortex-M devices, detailing processor type, clock frequency, memory, and FPU support.

Despite FPU availability in some models, integer operations—especially at lower bit-widths—remain faster and more memory-efficient, aligning with TinyML goals. For inference, Cortex processors commonly use the CMSIS-NN library~\cite{cmsis}, which leverages ARMv7E-M SIMD-like instructions to perform two 16-bit MAC operations in a single 32-bit accumulator per cycle. CMSIS-NN provides optimized kernels for convolution, activation, fully connected layers, pooling, and softmax, supporting fixed 8- and 16-bit types with power-of-two scaling~\cite{novac2021quantization}. However, it lacks native support for mixed precision or sub-byte values, limiting optimization for memory-constrained devices. Since TFLM~\cite{tensorflow} uses CMSIS-NN internally, it inherits the same constraints, including static memory allocation limitations (see Section~\ref{sec:challenges}).

To address these gaps, CMix-NN~\cite{capotondi2020cmix} was introduced as an open-source mixed precision framework for ARM Cortex-M, supporting 8-, 4-, and 2-bit weights and activations. It packs multiple low-precision values into 32-bit SIMD registers, sixteen 2-bit, eight 4-bit, or four 8-bit values, processed in parallel and unpacked as needed. Using CMix-NN, a MobileNetV1 was deployed on an STM32H7 (Cortex-M7) MCU, achieving 68\% accuracy at higher input resolutions.

However, a limitation of CMix-NN is its fixed SIMD packing layout, which can lead to partial lane utilization when tensor bit-widths misalign. MCU-MixQ~\cite{mcu-mixq} addresses this with a SIMD low bit-width convolution packing algorithm (SLBC) that adapts lane sizes to operation-specific bit-widths, plus a data reordering mechanism that reduces packing overhead. It also uses NAS to perform packing-aware quantization followed by QAT, aligning mixed precision configurations with MCU-optimized kernels. This hardware/software co-design yields a 1.5$\times$ speedup over CMix-NN on the same MCU.

While these software-level workarounds can enable some degree of mixed precision inference, true hardware support for dynamic and sub-byte bit-widths remains largely absent on Cortex-M–based MCUs. For this reason, recent efforts have turned toward rethinking the hardware itself, exploring both entirely new processor architectures and hybrid designs with integrated accelerators or specialized cores. We examine these emerging solutions in the sections that follow.

\subsection{RISC-V-based MCUs}

\begin{table*}[htbp!]
    \centering
    \caption{Specifications of surveyed RISC-V-based MCUs}
    \label{tab:riscv_mcu_specs}
    \vspace{-0.1cm}
    \footnotesize
    \renewcommand{\arraystretch}{1.5} 
    \begin{tabular}{llllll}
        \toprule
        \textbf{MCU} & \textbf{CPU} & \textbf{Flash} & \textbf{RAM} & \textbf{Clock} & \textbf{FPU} \\
        \midrule

        GAP8 & 8-core RISC-V & 20 MB & \makecell[l]{512 KB L2 SRAM, \\ 8 MB L3 RAM} & 250 MHz & N/A \\
        
        ESP32-C3 & Single-core RISC-V & 4 MB & \makecell[l]{400 KB SRAM, \\ 8 KB RTC SRAM} & 160 MHz& N/A \\
        
        ESP32-C6 & Single-core RISC-V & 8 MB & \makecell[l]{512 KB SRAM, \\ 16 KB RTC SRAM} & 160 MHz& N/A \\

        GAP9 & 9-core RISC-V & 2 MB & 1.6 MB L2 SRAM & 370 MHz& Single-precision \\
        
        ESP32-P4 & Dual-core RISC-V & 16 MB & 32 MB & 400 MHz& Single-precision \\
        
        BL616/BL618 & RV32IMAFCP & 8 MB & 8 MB PSRAM & 320 MHz& Single-precision \\
        
        CH32V003 & QingKe RISC-V2A & 16 KB & 2 KB SRAM & 48 MHz& Single-precision \\
    
        RP2350 & Dual-core RISC-V Hazard3 & 4 MB & 520 KB & 150 MHz& Single-precision \\
        
        \bottomrule
    \end{tabular}
\end{table*}

To address the limitations of ARM-based MCUs, specifically their proprietary nature and fixed ISAs, which constrain flexibility for specialized applications \cite{novac2021quantization}, the research community has increasingly turned toward RISC-V. This open-source and extensible ISA enables the design of highly customized hardware solutions tailored to specific needs. Managed by a non-profit foundation, RISC-V provides several notable advantages:

\begin{itemize}
\item Full hardware customization for specific application requirements (e.g., power, area, performance).
\item Reduced reliance on third-party licensing and intellectual property for DNN accelerators.
\item Compatibility, interoperability, and adherence to emerging standards across platforms.
\item Strong community support and long-term sustainability, including frameworks, compilers, and software stacks that facilitate deployment.
\end{itemize}

The base RISC-V implementation typically uses a 32-bit processor with a standard instruction set and optional extensions such as floating-point and vector instructions. Popular cores include RV12~\cite{rv12_datasheet}, E203~\cite{wu2020reconfigurable}, RI5CY~\cite{vrevca2020accelerating}, and Rocket~\cite{zhang2019risc}. A key development platform is the Parallel Ultra-Low-Power (PULP) platform~\cite{pulp}, from the University of Bologna and ETH Zurich, integrating multiple RISC-V cores, most notably the energy-efficient RI5CY, and forming the basis for high-performance, low-power accelerators.

Early PULP-based MCUs such as PULPino~\cite{traber2016pulpino} and PULPissimo~\cite{8640145} featured single-core designs, while later architectures like Trikarenos~\cite{rogenmoser2023trikarenos} adopted multi-core configurations. Commercial derivatives include the GAP8~\cite{flamand2018gap} and GAP9~\cite{gap9} from Greenwave Technologies, offering nine RISC-V cores with on-chip RAM and flash for robust edge AI workloads.

To support these architectures, the research community developed optimized software libraries. For example, the PULP-NN library \cite{pulpNN} enables deployment of quantized neural networks (QNNs) from 8-bit down to 1-bit. Later, Bruschi et al. \cite{bruschi2020} extended PULP-NN to enable mixed precision inference. However, these early implementations lacked low bit-width SIMD support, meaning quantization acted primarily as a form of memory compression without improving computational time or energy efficiency, since data had to be upcast to native hardware types, introducing overhead.

To address this gap, XpulpNN \cite{xpulpnn20} introduced a RISC-V ISA extension that added native low bit-width SIMD operations, enabling direct packed SIMD computation without unpacking, thereby maximizing SIMD lane utilization. Building on this, a mixed precision dot-product (Dotp) unit was designed \cite{xpulpnn21}, supporting single-cycle latency operations across 16- to 2-bit SIMD vectors. Integrated into a new PULP-based architecture, this unit delivered efficiency levels comparable to dedicated DNN inference accelerators, achieving up to three orders of magnitude better performance than state-of-the-art ARM Cortex-M MCUs. Similarly, Ottavi et al. \cite{ottavi2020mixed} proposed a mixed precision inference core (MPIC) and a mixed precision controller (MPC) built around a RISC-V processor, demonstrating two orders of magnitude better efficiency than existing commercial MCU solutions.

Mixed precision SIMD introduces overhead due to the added instruction complexity and memory access costs, which is exacerbated in multi-core systems. To mitigate this, Dustin~\cite{dustin} proposed a reconfigurable architecture switching between MIMD and Vector Lockstep Execution Mode (VLEM), reducing redundant fetch/decode stages. For on-device training, DARKSIDE~\cite{Garofalo2022} offers 2–32 bit support with a Tensor Product Engine for efficient matrix ops, while Huang~\cite{Huang} extends XpulpNN with mixed fixed-point training and inference, achieving up to 14.6$\times$ higher throughput via a specialized dataflow co-processor.

Beyond general-purpose accelerators, several works have proposed application-specific RISC-V-based DNN accelerators. For example, Zhang et al. \cite{zhang2019risc} developed a coprocessor integrated with the Rocket core to execute the convolutional layers of Darknet-19 \cite{redmon2018yolov3} and Tiny-YOLO \cite{redmon2018yolov3} entirely on-chip, achieving energy-efficient inference with a 400 ms end-to-end runtime and minimal FPGA resource usage. More recently, transformer-based models have begun to reach the MCU domain. For example, the SwiftTron accelerator \cite{marchisio2023swifttron} integrates an INT8 matrix multiplication unit with a dyadic requantization datapath to run ViT blocks entirely in integer arithmetic, achieving up to 28× speedup over an ARM Cortex-A (ARMv8-A) baseline while maintaining a sub-1 W power envelope.

While most of the RISC-V-based accelerators mentioned above, with the exception of GAP8 and GAP9, are research prototypes tested primarily on FPGAs, several commercial RISC-V-based MCUs are now available. Espressif’s ESP32-P4 \cite{esp32p4}, ESP32-C3 \cite{esp32c3}, and ESP32-C6 \cite{esp32c6} are popular wireless MCUs designed for IoT applications, providing Wi-Fi and Bluetooth connectivity with RISC-V cores. Bouffalo Lab’s BL616 and BL618 MCUs \cite{bl616}, also built for IoT, integrate RISC-V cores with low-power wireless capabilities. The WCH CH32V003 \cite{ch32v003} offers an ultra-low-cost, general-purpose RISC-V MCU designed for embedded control tasks. Finally, the Raspberry Pi Pico 2 features the RP2350 chip, which optionally integrates dual RISC-V Hazard3 processors for improved parallelism and flexibility \cite{rp2350}.

Table~\ref{tab:riscv_mcu_specs} details hardware specifications, including memory, performance, and feature sets of RISC-V-based MCUs and accelerators.

Essentially, these specialized designs demonstrate that with careful co-optimization of the ISA and microarchitecture, the open-source RISC-V ecosystem can achieve ASIC-like efficiency while maintaining flexibility, making it an increasingly attractive foundation for both general-purpose and application-specific deep learning workloads.

For a more comprehensive overview of RISC-V architectures and their role in embedded ML, we refer the reader to the surveys by Akkad et al.~\cite{akkad2023embedded} and Liu et al.~\cite{liu2024exploring}.

\subsection{Hybrid / NPU-augmented MCUs}
\label{sec:NPU}

\begin{table*}[htbp]
    \centering
    \caption{Specifications of surveyed $\mu$NPU-based MCUs and Accelerators}
    \label{tab:npu_mcu_specs}
    \vspace{-0.1cm}
    \footnotesize
    \begin{tabular}{llllllll}
        \toprule
        \textbf{MCU} & \textbf{CPU(s)} & \textbf{NPU}  & \textbf{Flash} & \textbf{RAM} & \textbf{Clock} & \textbf{GOPs} & \textbf{Bit Precision}\\ 
        \midrule
         
         \makecell[l]{MAX78000} &  \makecell[l]{Cortex-M4 \\ RISC-V} & CNN-Accelerator & 512 KB & \makecell[l]{512 KB NPU \\ 128 KB CPU} & \makecell[l]{100 MHz \\ 60 MHz} & 30 & 1, 2, 4, 8  \\ 

        \makecell[l]{MAX78002} &  \makecell[l]{Cortex-M4 \\ RISC-V} & CNN-Accelerator & 2.5 MB & \makecell[l]{1.3 MB NPU \\ 384 KB CPU} & \makecell[l]{120 MHz \\ 60 MHz} & 30 & 1, 2, 4, 8  \\ 

        GAP8 & 8-core RISC-V & HWCE & 20 MB & \makecell[l]{512 KB L2 SRAM, \\ 8 MB L3 RAM} & 250 MHz & 8-10 & 4, 8, 16 \\

         GAP9 & 9-core RISC-V & NE16 & 2 MB & 1.6 MB L2 SRAM & 270 MHz& 50 & 2, 3, 4, 5, 6, 7, 8 \\
         
         \makecell[l]{HX6538-WE2} & Cortex-M55 & Ethos-U55 & 16 MB & \makecell[l]{2 MB SRAM \\ 512 KB TCM} & \makecell[l]{400 MHz} & 50 & 8, 16\\ 
         
         \makecell[l]{NXP-MCXN947} & Dual-core Cortex-M33 & eIQ Neutron & 2 MB & 512 KB & 150 MHz & 4.8 & 8 \\ 
         
         \makecell[l]{STM32N6} & Cortex-M55 & ST Neural-ART & 2 MB & \makecell[l]{4.2 MB SRAM \\ 128 KB TCM} & 800 MHz & 600 & 8
         \\

        \bottomrule
    \end{tabular}
\end{table*}

In recent years, hybrid microcontroller platforms that combine traditional CPUs with dedicated neural processing units (NPUs) have emerged as a promising class of devices, addressing many limitations of earlier MCU-based systems. These platforms typically integrate an ARM Cortex-M processor alongside a RISC-V-based NPU or custom neural network accelerator, enabling efficient on-device DL for ultra-low-power edge applications.

While conventional neural accelerators such as GPUs and TPUs provide high throughput, they are impractical for embedded contexts due to power and size constraints~\cite{kim2023hardware, lin2020mcunet}. Conversely, standalone MCUs lack the compute capacity for real-time neural workloads, leaving a hardware gap.

To address this, microcontroller-scale NPUs (commonly referred to as $\mu$NPUs) have been developed as lightweight, energy-efficient accelerators tailored to neural operations~\cite{millar2025benchmarking}. Tightly integrated into MCU platforms, $\mu$NPUs speed up compute-intensive operations like matrix multiplication and convolution while maintaining ultra-low-power consumption~\cite{manor2022custom, wang2022bed, song2024tada}. Performance is typically measured in Giga-operations per second (GOPS).

Among neural network architectures, CNNs are the most widely deployed on edge devices, owing to their inherent parallelism and tolerance for approximate computing~\cite{sekanina2021neural}. They can be further compressed through quantization, pruning, and similar techniques to reduce power, memory, and latency requirements with minimal performance degradation~\cite{akkad2023embedded}. As a result, $\mu$NPUs are primarily designed as CNN accelerators.

A prominent example of commercial $\mu$NPU platforms is the Maxim Integrated MAX78000 series, designed for ultra-low-power edge AI applications~\cite{balbi2025embedded}. These devices integrate a low-power ARM Cortex-M4 core, a RISC-V controller, and a proprietary CNN accelerator delivering up to 30 GOPS. Notably, the accelerator supports ultra-low bit-width weights down to 1-bit, enabling dense memory packing and bit-wise acceleration. The newer MAX78002 expands on this with increased weight memory and enhanced compute capabilities. Both are well-suited for battery-powered deployments and remain among the most extensively documented platforms in this space.

Another leading example is the GAP series from GreenWaves Technologies, which target edge AI applications using clusters of low-power RISC-V cores. The GAP8 and GAP9 feature proprietary CNN accelerators: the Hardware Convolution Engine (HWCE) and 16-channel Neural Engine (NE16), respectively. These platforms offer strong performance under tight power constraints and support mixed-precision inference with varying bit-widths.

STMicroelectronics offers the STM32N6, a high-performance MCU featuring an ARM Cortex-M55 core and the ST Neural-ART Accelerator~\cite{bartoli2025benchmarking}. The convolutional NPU achieves up to 600 GOPS at 1 GHz, supports INT8 weights and activations with per-channel quantization, and defers higher-precision operations to the CPU. It includes 4.2 MB of SRAM and interfaces for external memory.

Another example is the Himax HX6538-WE2, built around an ARM Cortex-M55 and an Ethos-U55 NPU~\cite{akkad2023embedded}. It delivers up to 50 GOPS, with 512 KB TCM, 2 MB SRAM, and 16 MB flash. Designed for more complex models, it offers greater capacity at a higher power budget.

These hybrid MCU-NPU platforms represent a clear shift in embedded AI, moving beyond handcrafted firmware toward deployable neural models that run efficiently on ultra-constrained hardware. Table~\ref{tab:npu_mcu_specs} details the hardware specifications, including memory, performance, and feature sets of the discussed MCU-NPU platforms. For a detailed overview of $\mu$NPU architectures, benchmarks, and deployment case studies, we refer readers to the survey by Millar et al.~\cite{millar2025benchmarking}.

\section{Software Frameworks}
In this section, we present frameworks for lightweight model training and inference from a software perspective, as well as methods for accelerating models using hardware-based approaches.
\label{sec:software}

\subsection{Edge‑AI Training Frameworks}

For much of the recent history of the field of deep learning, TensorFlow has been the preferred framework for production-level deployment in real-world applications, while PyTorch gained popularity primarily within the research community. Over time, this distinction has become increasingly blurred, with PyTorch now being widely adopted across both research and production domains.  However, models developed in either framework are often too large and resource-intensive for direct deployment on edge devices such as mobile phones or microcontrollers, where memory, power, and latency are constrained.

To address this, both have introduced lightweight deployment solutions: TensorFlow Lite (now LiteRT) and PyTorch Mobile (succeeded by PyTorch Edge’s \textit{ExecuTorch}). These frameworks convert and optimize models for efficient on-device inference on resource-constrained hardware.

TensorFlow Lite converts full-scale models into the compact \texttt{.tflite} format and provides a dedicated runtime interpreter. It now offers limited on-device training for personalization and federated learning, and its LiteRT rebrand maintains backward compatibility while adding enhanced hardware delegation and a unified runtime.

In parallel, PyTorch Edge’s ExecuTorch is a lightweight C++ runtime for mobile, embedded, and MCU platforms. It improves on PyTorch Mobile with an efficient \texttt{torch.export} pipeline, dynamic memory planning, and hardware delegation via XNNPACK, CoreML, Vulkan, and vendor APIs. Designed for tight integration with the PyTorch ecosystem, ExecuTorch enables fine-grained optimization for target devices, currently supporting inference only. Already deployed in products such as Meta Quest and Ray-Ban Smart Glasses, it is gaining momentum with backing from Arm, Apple, Qualcomm, and Unity.

On the optimization front, both frameworks provide robust quantization toolchains. TensorFlow’s Model Optimization Toolkit (TFMOT) supports PTQ, QAT, pruning, and clustering; PyTorch’s \texttt{torch.ao.quantization} offers dynamic/static PTQ, weight-only quantization, and QAT, with Torch-TensorRT integration. ExecuTorch introduces advanced workflows, including 4-bit GPTQ-style quantization, for efficient on-device LLM deployment.

Beyond general-purpose frameworks, specialized toolchains have emerged for deploying neural networks on ultra-low-power microcontrollers. A notable example is Analog Devices’ MAX78000, supported by the PyTorch-based \href{https://github.com/analogdevicesinc/ai8x-training}{\texttt{ai8x-training}} repository~\cite{balbi2025embedded} for INT8 QAT with deployment-aware constraints, converting trained models to C code via \href{https://github.com/analogdevicesinc/ai8x-synthesis}{\texttt{ai8x-synthesis}}. Another notable advance is Fully Quantized Training (FQT)~\cite{deutel2024device}, enabling complete 8-bit integer training—including forward and backward passes—on ARM Cortex-M MCUs. FQT uses a unified quantization scheme, dynamic parameter updates, minibatching, and sparse gradient updates for up to $8.7\times$ faster training, validated on both vision and time-series tasks.

Despite growing momentum for PyTorch-based solutions, TensorFlow Lite and LiteRT remain dominant due to their mature optimization ecosystems and adoption. Nevertheless, the rise of PyTorch Edge and ExecuTorch offers compelling benefits for developers seeking close integration with PyTorch pipelines and modern quantization/delegation features. Meanwhile, \texttt{ai8x-training} highlights PyTorch’s versatility for MCU deployment, and FQT points toward the future of on-device training even on the most constrained platforms.

\subsection{Edge-AI Inference Frameworks}

Directly porting ML models designed for high-end edge devices is not suitable for MCUs \cite{saha2022}. While inference is the primary goal of deploying ML models on these devices, the deployment process itself is an essential precursor to ensure that the models are optimized to perform effectively within the resource constraints of MCUs.

After a DNN has been trained and optionally quantized, it is deployed onto a microcontroller for inference. The deployment process begins by porting the best-performing model to a TinyML software suite, which first encodes the model weights into a format (typically an array of bytes) that can be efficiently processed by the embedded hardware. The inference program is then generated by structuring it according to the DNN's topology, ensuring that the network's structure and operations are properly replicated on the target MCU, with operator optimizations also applied. Finally, the inference program, along with the encoded weights, is compiled into embedded C code and flashed to the MCU's flash memory for inference \cite{novac2021quantization, saha2022}. 

Deployment choices, such as quantization format and model size, directly influence the efficiency and accuracy of on-device inference. Therefore, deploying complex models onto such constrained devices often requires extensive software and hardware optimizations to achieve feasible performance while maintaining inference accuracy. Moreover, some of these software frameworks offer inference engines for resource management and model execution during deployment. We will discuss some of these frameworks below.

\textbf{TensorFlow Lite for Microcontrollers (TFLM)} \cite{tensorflow}: Is a specialized version of TensorFlow Lite (TFLite) tailored for running TinyML models on embedded systems, particularly on Cortex-M and ESP32-based MCUs. TFLM is composed of three main components: first, an operator resolver that links only the essential operations to the model binary file; second, a preallocated contiguous memory stack, known as the "arena," used for initialization and storing runtime variables, which prevents dynamic expansion to avoid memory fragmentation issues; and third, an interpreter (inference engine) that resolves the network graph at runtime, allocates the arena, and performs runtime calculations. However, TFLM lacks support for on-device training. 

\textbf{STM32Cube.AI}\cite{xcube_ai}: Is a software suite provided by STMicroelectronics, specifically designed for deploying deep neural networks (DNNs) on STM32 Cortex-M-based MCUs. It facilitates the deployment of trained DNN models from popular frameworks such as Keras and TensorFlow Lite. The primary advantage of \texttt{STM32Cube.AI} is its seamless compatibility with ST hardware, allowing for efficient runtime memory optimization. However, this tight integration results in reduced flexibility, as it is optimized explicitly for STM32 architectures, which limits its applicability to other hardware platforms.

\textbf{Edge Impulse} \cite{edgeImpulse}: Is a comprehensive end-to-end deployment solution for TinyML devices, encompassing everything from data collection using IoT devices to feature extraction, model training on a cloud platform, and ultimately model deployment and optimization for TinyML devices. Edge Impulse employs the interpreter-less Edge Optimized Neural (EON) compiler for model deployment, while also providing support for TFLM. The EON compiler directly compiles the model into C/C++ source code, eliminating the need to store unused ML operators and thereby optimizing resource utilization, albeit at the cost of reduced portability. 

\textbf{MinUn} \cite{minun}: Is a new framework jointly developed by Microsoft Research in India, ETH Zurich, and UC Berkeley, designed explicitly for TinyML applications. It is a comprehensive framework that primarily targets Arm MCUs and addresses many of the limitations posed by current frameworks, such as TFLM. MinUn essentially performs PTQ with mixed precision, determining the optimal bit-width for each tensor through a proposed algorithm. Additionally, it implements a smart memory management technique that dynamically allocates memory for different tensor sizes and supports fixed-point arithmetic at runtime. This approach optimizes bit-width assignment to ensure minimal memory usage while preserving accuracy, and the dynamic memory management helps mitigate memory fragmentation issues.

\textbf{ai8x-synthesis} \cite{balbi2025embedded}: Is the deployment toolchain developed by Analog Devices for its ultra-low-power MAX78000 microcontroller. It works in conjunction with the PyTorch-based \texttt{ai8x-training} environment and is designed to convert quantization-aware trained models into highly optimized embedded C code. By leveraging architectural knowledge, such as SRAM partitioning and MAC unit scheduling, \texttt{ai8x-synthesis} ensures that inference runs within the strict resource constraints of the MAX78000. Unlike general-purpose interpreters, \texttt{ai8x-synthesis} produces static binaries tailored specifically to the accelerator hardware for extremely efficient inference on microcontroller platforms.

\textbf{ONNX Runtime} \cite{onnxruntime}: Is a cross-platform ML inference engine developed by Microsoft, designed to execute models in the Open Neural Network Exchange (ONNX) format efficiently on a wide range of hardware. It supports models exported from popular training frameworks, including PyTorch, TensorFlow, and scikit-learn, enabling deployment with minimal code changes. The framework utilizes graph-level optimizations and a modular backend system, known as execution providers, to offload subgraphs to hardware-specific accelerators, including TensorRT (NVIDIA GPUs), OpenVINO (Intel), and XNNPACK (ARM CPUs). Although primarily designed for high-performance inference with support for quantization and mixed precision, it can also be customized for edge deployment by selectively including only the necessary operators.

\textbf{TinyMaix} \cite{tinymaix}:
Is an ultra-lightweight neural network inference library for microcontrollers, released by Sipeed for TinyML applications. Developed as a minimalist C library, TinyMaix is compact enough to run basic CNN models on resource-constrained MCUs like the Arduino UNO, by using static memory allocation and highly efficient code. It supports INT8, FP16, and FP32 models, providing conversion tools to compress pre-trained models into a custom lightweight binary format or C header files for deployment. Furthermore, Sipeed also created a platform for convenient \href{https://maixhub.com/welcome}{online model training} to expedite AI model development for deployment.

\textbf{Neural Network on Microcontroller (NNoM)} \cite{nnom}: Is a higher-level inference library tailored for deploying complex deep learning models on microcontrollers. NNoM provides an abstraction that manages the neural network’s graph structure, memory usage, and execution flow internally, enabling developers to take a model designed in high-level frameworks and deploy it to an MCU in a C-based model representation with minimal manual effort. The library runs on a pure C backend by default, with no external dependencies, ensuring broad compatibility across platforms. It can optionally utilize optimized kernels, such as Arm’s CMSIS-NN, for a noticeable speedup on Cortex-M cores. NNoM also facilitates quantized inference, including per-channel quantization for convolution layers in its latest version, allowing 8-bit fixed-point models to run faster on MCU hardware.

\textbf{TinyNeuralNetwork (TinyNN)} \cite{tinynn}: Is a lightweight neural network inference framework developed by Alibaba for deploying deep learning models on resource-constrained microcontrollers. TinyNN converts PyTorch models, including those trained with QAT, into 8-bit TFLM models optimized for INT8 inference, thereby supporting a range of MCU platforms, including ARM Cortex-M4 and RISC-V architectures.

\textbf{GAPflow} \cite{gapflow}: Is an end-to-end deployment toolchain for running TinyML models on GreenWaves Technologies' GAP8 and GAP9 platforms. This SDK includes two key components: \href{https://github.com/GreenWaves-Technologies/gap_sdk/tree/master/tools/nntool}{nntool}, which handles model parsing, quantization, and graph transformations, and Autotiler, which generates optimized C code for the GAP microcontroller architecture. On the one hand, nntool applies INT8 PTQ to compress models for model deployment. Additionally, it supports mixed precision deployment, allowing some layers to utilize higher precision formats, such as bfloat16. On the other hand, Autotiler optimizes memory access and computation by breaking large layers into tiles that fit within the microcontroller's SRAM, and orchestrates memory transfers to and from on-chip and off-chip memory using double- or triple-buffered DMA. This method minimizes idle time and ensures that the multi-core RISC-V architecture of the GAP8 and GAP9 is kept saturated, achieving high inference throughput with low latency under tight energy constraints.

\textbf{HEEPstor} \cite{palacios2025heepstor}: Is an open-hardware co-design framework developed at EPFL for deploying PyTorch models on RISC-V-based MCUs with custom accelerators. HEEPstor performs symmetric per-tensor post-training INT8 weight quantization while retaining the original floating-point precision for activations, offering a hybrid quantization approach. Then, it generates general matrix multiplication (GEMM)-optimized C++ inference code using \texttt{im2col}~\cite{chellapilla2006high}, tiling, and custom memory management. Furthermore, it supports core PyTorch layers and runs on a systolic array tightly integrated into the X-HEEP MCU. Thus, its modular design enables rapid hardware-software exploration while maintaining model accuracy and performance on ultra-low-power platforms.

\begin{figure*}[ht]
    \centering
    \includegraphics[width=0.90\textwidth]{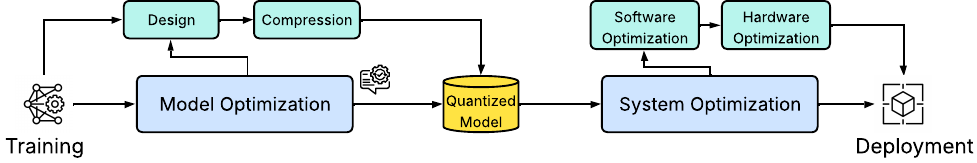}
    \caption{An overview of edge deployment. The figure illustrates a general pipeline spanning from model training and optimization to the quantized container and ultimately deployment on the edge.}
     \label{fig:deployment_cycle}
\end{figure*}

\section{Applications}
\label{sec:applications}

The culmination of the previously discussed quantization techniques, software frameworks, and hardware capabilities results in model deployment on the edge. Several studies have been dedicated to deploying TinyML models for various applications, spread across healthcare, agriculture, object detection, remote sensing, and many more. Generally, the process of deployment on the edge can be summarized as shown in Fig. \ref{fig:deployment_cycle}. We start with the model itself, which is trained and optimized for quantization using the methodologies discussed in earlier sections. Once the quantized model is available, it is optimized for the particular hardware system in use before being fully ready for deployment. 

In this section, we provide a systematic overview of the three hardware families discussed in this survey, detailing recent application-based studies across multiple domains. The section is divided based on the hardware family (as outlined in Section~\ref{sec:hardware}), with each section including a discussion of the relevant works, accompanied by tables summarizing the standout works. Such a division is necessary for two main reasons: (1) It is aligned with the hardware-focused theme of the paper, and (2) it provides the interested researcher with an extensive idea or ``feel'' of the recent progress in the field in terms of pushing the boundaries for hardware applications.

The analysis focuses on the quantization techniques employed and the quantization-based metrics suitable for evaluation. The chosen studies in this section have been selected based on the following important considerations: 
\begin{enumerate}
    \item Only genuine MCU deployments are considered, in which authors explicitly exclude higher-end embedded platforms such as Raspberry Pi boards. In particular, we exclude any studies that may mischaracterize such capable systems as MCUs. Thus, the presented works are a strong reflection of the strict resource constraints that MCU-class devices pose in recent years.
    \item Since performance metrics vary significantly across different application domains, we clearly specify the exact metrics used for comparisons (e.g., accuracy, F1-score, etc.).
    \item In cases where multiple datasets or model variants were evaluated, we exclusively report the results from quantized and deployed model versions, considering the range of worst- to best-performing models.
    \item Metrics are reported consistently in their original units as found in the source studies. 
    \item For memory consumption, we systematically use model size, which corresponds to the amount of memory needed to store model parameters in the flash memory of the MCU.  
\end{enumerate}

\subsection{ARM-based MCUs}

As discussed in Section~\ref{sec:arm}, MCUs based on the ARM processor family are among the most widely used platforms across embedded computing, supporting a vast array of applications. Prior work has explored ARM-based MCU deployments in diverse domains such as object detection \cite{moosmann2024flexible, moosmann2023tinyissimoyolo, dabbous2024benchmarking, deutel2023mu}, speech processing \cite{ulkar2021ultra, vsljubura2024deep, abbas2023keyword, zeng2022sub}, human activity recognition (HAR) \cite{rafee2025decision, kang2024device, zhou2025efficient, mach2024uwb, mishra2025use}, environmental monitoring \cite{flores2025enhanced, cerioli2025efficient, dabbous2024benchmarking, chen2024efficient, mellit2025tinyml}, agriculture \cite{rakib2024lightweight, sathyanarayanan2024quantized, krayden2025tinyml, crupi2024deep}, healthcare \cite{benoit2024analyzing, rostami2024real, yang2025real, gambhir2024microcontroller, kasnesis2025replacing, krasteva2025implementing, mazinani2024deep, zanetti2020robust, borah2024blockchain}, education \cite{abushahla2025real}, and networking \cite{chehade2025energy}.

Several works focus on privacy-preserving vision applications at the edge, including the egocentric privacy-preserving intelligent camera (EPIC) \cite{li2025extra} and XimSwap \cite{ancilotto2024ximswap}, which perform real-time video anonymization on smart cameras and edge-based vision sensors. These studies apply quantization techniques to reduce computational load and inference latency, achieving performance comparable to larger models while maintaining efficiency on constrained devices. Abushahla et al. \cite{abushahla2025cognitive} explored low-latency wideband spectrum sensing for cognitive radio using the Sony Spresense MCU, adapting state-of-the-art models with QAT to significantly reduce hardware requirements. Related efforts, such as \cite{hu2025tin}, also address edge-based spectrum sensing. In the area of model architectures, Burrello et al. \cite{burrello2021microcontroller} introduced the TinyTransformer library, providing optimized execution kernels for transformer-based models on low-power MCUs, showing that even complex models can be efficiently deployed at the edge.

Table~\ref{tab:quantized_arm_apps} lists selected examples that highlight the range of applications implemented on ARM-based MCUs, most of which rely on conventional quantization techniques, primarily INT8 QAT and PTQ.

Beyond these standard approaches, a number of studies have pushed the limits by exploring advanced or extreme quantization methods (as presented in Section~\ref{sec:advanced}). Sakr et al. \cite{sakr2022memory} and Borah et al. \cite{borah2024blockchain}, for example, implemented ultra-low-bit models, including binarized neural networks, by applying software-level and ISA-level optimizations that enable efficient deployment on microcontrollers. Zeng et al. \cite{zeng2022sub} employed a two-stage QAT pipeline, where the first stage used squashed quantization \cite{strom2022squashed} to transform the weight distribution toward a uniform distribution (in the statistical sense), reducing skewness and extreme values, followed by mixed precision quantization (8-, 5-, and 4-bit) for efficient compression and deployment. Cassimon et al. \cite{cassimon2025designing} applied kurtosis regularization from the KURE framework \cite{chmiel2020robust} to actively reduce the peakedness and tails of the weight distribution, promoting a near-uniform weight distribution in a model trained for active fire detection using multispectral satellite imagery, and subsequently applied INT8 PTQ for deployment on the Google Coral Micro. Misra et al. \cite{misra2025latency} extended this idea by combining kurtosis regularization from EQ-Net \cite{xu2023eq} with adaptive mixed precision QAT, where the quantization bit-width was dynamically selected based on input complexity. Their models were deployed on the Raspberry Pi Pico, demonstrating low-latency, energy-efficient inference across varying precision levels.

It is worth emphasizing that these are selected examples and do not represent the full breadth of possible deployments achievable with ARM-based MCUs.

\begin{table*}[htbp]
    \centering
\caption{Selected papers using quantization for deployment on ARM-based MCUs.}
\label{tab:quantized_arm_apps}
    
    \scriptsize
    \setlength{\tabcolsep}{3pt} 
    \renewcommand{\arraystretch}{1.2} 
    \begin{tabular}{lllp{2.2cm}lp{2cm}p{1.8cm}p{1.8cm}p{2cm}}
        \toprule
        \textbf{Study} & \textbf{Category} & \textbf{Quantization} & \textbf{Device(s)} & \textbf{Framework(s)} & \textbf{Performance} & \textbf{Power / Energy} & \textbf{Latency} & \textbf{Memory} \\
        \midrule
        \cite{ulkar2021ultra} & Speech & INT8 QAT & \makecell[l]{ARM Cortex M4F} & PyTorch & 96.3\% Acc. & 11200 $\mu$J &  905 ms & 419.8 KB \\
        \cite{moosmann2023tinyissimoyolo} & Object Detection & INT8 QAT & \makecell[l]{STM32H7A3, \\ STM32L4R9, \\ Apollo4b} & TFLM & 43.5--75.4\% mAP & \makecell[l]{41.8 mJ, \\ 102 mJ, 6.08 mJ} & \makecell[l]{359 ms, \\ 996 ms, 540 ms} & 350--422 KB \\
        \cite{kang2024device} & HAR & INT8 PTQ & STM32F756ZG & PyTorch & 91.04--98.29\% Acc. & \makecell[l]{4.50--8.77 mJ, \\ 3.5e-2--6.3e-2 mJ} & \makecell[l]{20.70--39.76 ms, \\ 1.11--1.93 ms} & 37.8--44.7 KB \\
        \cite{chehade2025energy} & Networking & INT8 PTQ & \makecell[l]{STM32F746G, \\ Nucleo-F401RE} & TFLM, STM32Cube.AI & 96.59\% Acc. & 7.86 mJ, 29.10 mJ & 31.43 ms, 115.40 ms & 353 KB \\
        \cite{cerioli2025efficient} & Environment & INT8 PTQ & \makecell[l]{STM32H747XI \\ nRF52840 \\ ESP32-PICO-D4} & \makecell[l]{ONNX Runtime, \\ NeuralCasting, \\ TensorFlow Lite} & 98.5--99.1\% Acc. & \makecell[l]{0.6 W, 0.1 W, \\ 2.5 W} & \makecell[l]{34--246 $\mu$s, \\ 34--252 $\mu$s, \\ 308--2001 $\mu$s, \\ 40--262 $\mu$s} & 0.14--0.58 MB \\
        \cite{abushahla2025real} & Education & INT8 QAT & \makecell[l]{Sony Spresense, \\ OpenMV Cam H7, \\ H7 Plus} & TFLM & 93.60--98.73\% Acc. & \makecell[l]{494--884 mW, \\ 1238--1609 mW} & \makecell[l]{0.63 ms, 0.08 ms, \\ 2.88--12.84 ms} & 2.9 MB, 0.34 MB \\
        \cite{abushahla2025cognitive} & Spectrum Sensing & INT8 QAT & Sony Spresense & TFLM & \makecell[l]{92.88--99.09\%, \\ 70.55--99.09\% F1} & \makecell[l]{52--54 mW, \\ 44--52 mW} & \makecell[l]{5.54--37.37 ms, \\ 1.18--5.20 ms} & \makecell[l]{23.8--72.6 KB, \\ 12.8--19.6 KB} \\
        \cite{zhou2025efficient} & HAR & UINT8 PTQ & \makecell[l]{Arduino Nano 33\\BLE Sense Rev2} & \makecell[l]{TFLM, \\ EdgeImpulse} & 97.09\% Acc. & 0.021 W & 21 ms & 189.6 KB \\
        \cite{dabbous2024benchmarking} & Environment & INT8 QAT & \makecell[l]{STM32H7, \\ Arduino Nano 33\\BLE Sense Rev2} & \makecell[l]{TFLM, STM32Cube.AI} & 99.63\% Acc. & 227.86 mW, 1.43 mW, 9 mW & 30.25 ms, 522.15 ms, 1.45 ms & -- \\
        \cite{rostami2024real} & Healthcare & \makecell[l]{INT8 QAT, \\ PQAT, PTQ} & \makecell[l]{STM32H743iit6, \\ STM32H750vbt6} & TFLM & 84.78--87.76\% Acc. & 168 mA, 189 mA & 6.23 s, 6.30 s & 0.86 MB \\
        \bottomrule
    \end{tabular}
\end{table*}

\subsection{RISC-V-based MCUs}

Similar to ARM-based systems, RISC-V-based MCUs are gaining significant traction across diverse application domains. Prior work has explored RISC-V MCU deployments in object detection \cite{testa2025real, moosmann2024ultra, moosmann2024flexible}, vision tasks \cite{rashid2024tinyvqa, perlo2024software, zemlyanikin2019512kib, scherer2022widevision}, healthcare \cite{kasnesis2025replacing, busia2025endoscopy, mai2024end, itani2025wireless}, human activity recognition (HAR) \cite{daghero2022human, kang2024device, zimmerman2024fully}, speech processing \cite{cioflan2024device}, agriculture \cite{rashid2025hac}, and drone applications including navigation and obstacle avoidance \cite{chacun2024dronebandit, zhou2023solving, scarciglia2025map, lamberti2023bio, navardi2022e2edgeai, zimmerman2024fully, zheng2024monocular, hulens2022autonomous, crupi2024deep, crupi2025efficient, palossi2021fully}.

Several studies also explore unique and cutting-edge use cases. Burrello et al. \cite{burrello2021microcontroller}, for example, demonstrated the efficient execution of transformer-based models on RISC-V-based MCUs for ultra-low-power deployments, introducing the TinyTransformer library with optimized execution kernels. Busia et al. \cite{busia2024tiny} deployed a tiny ViT model for arrhythmia diagnosis, quantizing to 8 bits and running the model on the GAP9. In a similar vein, Rashid and Mohsenin \cite{rashid2024tinym} introduced TinyM$^2$Net-V3, a framework for memory-aware compressed multimodal deep neural networks that integrates multiple complementary data modalities. Their approach combines model compression techniques, including knowledge distillation and low bit-width quantization, with memory hierarchy–aware strategies to fit models into constrained memory levels, thereby reducing latency and improving energy efficiency.

Other research has focused on improving reliability and robustness. De Melo et al.~\cite{de2025microfi} introduced MicroFI, a framework that uses software-level fault injection during DNN inference—targeting memory and CPU registers, to assess error resilience under safety-critical conditions on MCUs.

In the wearable technology domain, Itani et al. \cite{itani2025wireless} developed NeuralAids, a hearing aid device with real-time speech enhancement and denoising capabilities. The system integrates a compact, battery-powered AI accelerator based on the GAP9 MCU and leverages mixed precision quantization and QAT to meet strict real-time and power constraints. Similarly, Ingolfsson et al. \cite{ingolfsson2025wearable} introduced BioGAP, an EEG-based wearable system for speech imagery decoding (specifically vowel imagination) that is optimized for embedded speech classification. BioGAP, also built on the GAP9, explores continual learning strategies for adaptive user-specific performance.

Table~\ref{tab:quantized_riscv_apps} lists selected works that demonstrate the diversity of applications deployed on RISC-V-based MCUs, most of which rely on conventional quantization strategies such as INT8 PTQ.

Beyond these standard approaches, several studies have explored advanced and extreme quantization techniques. Rutishauser et al. \cite{rutishauser2024xtern} and Cerutti et al. \cite{cerutti2020sound} implemented binary and ternary neural networks, pushing quantization down to extreme low-bit levels. Risso et al. \cite{risso2022channel} applied per-channel mixed precision quantization at sub-byte granularity (8-, 4-, and 2-bit), enabling adaptive neural networks on the MPIC \cite{ottavi2020mixed}. Daghero et al. \cite{daghero2022human} similarly employed sub-byte and mixed precision quantization, deploying their models on the PULPissimo MCU. Carlucci~\cite{carlucci2024optimization} and Nagar and Engineer~\cite{nagar2025energy} explored mixed precision quantization techniques for deployment on the GAP8 and GAP9 platforms, respectively, demonstrating how fine-grained bit-width control can enhance performance and energy efficiency on these RISC-V-based edge devices.

\begin{table*}[htbp]
    \centering
    \caption{Selected papers using quantization for deployment on RISC-V MCUs.}
    \label{tab:quantized_riscv_apps}

    \scriptsize
    \setlength{\tabcolsep}{4pt} 
    \renewcommand{\arraystretch}{1.2}
    \begin{tabular}{lp{1.7cm}p{1.4cm}p{1.3cm}p{2.0cm}p{3cm}p{1.7cm}p{1.7cm}p{2cm}}
        \toprule
        \textbf{Study} & \textbf{Category} & \textbf{Quantization} & \textbf{Device(s)} & \textbf{Framework(s)} & \textbf{Performance} & \textbf{Power / Energy} & \textbf{Latency} & \textbf{Memory} \\
        \midrule
        \cite{moosmann2024ultra} & Object Detection & INT8 PTQ & GAP9 & PyTorch & 14--49\% mAP & 54 mW & 56.45 ms & $<$1 MB \\
        \cite{busia2025endoscopy} & Healthcare & INT8 PTQ & GAP9 & PyTorch & 98.5\% Acc. & 30.6 mW & 61 ms & 750 KB \\
        \cite{itani2025wireless} & Healthcare & QAT Mixed & GAP9 & -- & 8.19 $\pm$ 3.38 SISDRi (dB) & 71.64 mW & 5.54 ms & 298.8 KB \\
        \cite{kang2024device} & HAR & INT8 PTQ & GAP9 & PyTorch & 91.04--98.29\% Acc. & 35--62 $\mu$J & 1.11--1.93 ms & 37.8--44.7 KB \\ 
        \cite{zemlyanikin2019512kib} & Vision & INT16 PTQ & GAP8 & \makecell[l]{PyTorch, \\ GAP8 Autotiler} & 0.9300--0.9633 Acc. & -- & $<$1 s & 1.5 MiB \\
        \cite{zhou2023solving} & Drones & INT8 PTQ & GAP8 & \makecell[l]{TFLite, \\ GAP8 Autotiler} & \makecell[l]{0.988 Acc., \\ 0.572 Succ. Score} & 130 mW & 28 FPS & 292 KB \\
        \cite{rashid2025hac} & Agriculture & INT8 PTQ & GAP8 & TFLM & 95\% Acc. & 378 mW & 37.6 ms & 49.6 KB \\
        \cite{rashid2023hac} & HAR & INT8 PTQ & GAP8 & \makecell[l]{GAPFlow, \\ TFLite} & 95.6\% Acc. & 307.6 mW & 49.2 ms & 58 KB \\
        \cite{crupi2025efficient} & Drones & \makecell[l]{INT8, \\ FP16 PTQ} & GAP9 & nntool & 0.79 mAP, 0.80 mAP & 34--41 mW & 147--462 ms & 1.8--3.6 MB \\
        \bottomrule
    \end{tabular}
\end{table*}

\subsection{Hybrid / NPU-augmented MCUs}

More recently, hybrid and NPU-based MCUs are gaining traction due to the advanced capabilities they offer, including orders of magnitude faster inference times and lower power consumption. Like the previous ones, works have considered these devices in a multitude of applications including object detection \cite{millar2025benchmarking, okman20223u, fiala2023tpdnet, moosmann2024flexible, moosmann2023tinyissimoyolo, wang2022bed, tran2025facets}, human activity recognition \cite{plozza2022real, mahmud2024actsonic, yen2023keep, mnif2024ultra, yen2023keep}, audio \cite{balbi2024ultra}, wearable devices \cite{gong2023collaborative, gong2025synergy}, environment-related applications \cite{ingaleshwar2024wildlife, dabbous2024benchmarking}, healthcare \cite{van2023real, nguyen2024low, ibrahim2024end}, vision \cite{gong2024dex, yen2023keep}, anomaly detection \cite{lightbody2022host}, speech \cite{jakuvs2025implementing, ulkar2021ultra, yen2023keep}, and DSP \cite{liu2025research}.

Table~\ref{tab:quantized_npu_apps} lists some selected works that represent the range of applications deployed on NPU-based MCUs. 

Although these hybrid and NPU-augmented devices explicitly support extreme low-bit and extreme precision techniques, none of the works we surveyed utilized such capabilities as of late---instead, defaulting to INT8 quantization. Therefore, we will only be exploring unique and exciting applications here.

Much of the unique applications we identified involved some form of novel latency reduction technique for NN inference by leveraging the AI accelerator. For example, in an effort to reduce the number of expensive memory fetch operations, the authors Vaddeboina et al. \cite{vaddeboina2024pagori} built upon their previous work \cite{vaddeboina2023parallel} to develop a hardware-efficient weight decompression mechanism, leveraging a neural network accelerator. Using their decoder, they observed that 8-weight decoding significantly outperforms 4-weight decoding in latency. In a similar vein, Jeon et al. \cite{jeon2025tinymem} aim to reduce the weight loading time and the end-to-end latency of multi-DNN inference by introducing \textit{2D Weight Memory Coordination}. Such a system can achieve up to 5$\times$ the gains in throughput for multi-DNN inference through weight preservation, preloading, and packing within the AI accelerator.

The other two papers we will highlight may not have introduced novel ways to minimize latency, but we believe that they are important for a resource-constrained direction for robotics \cite{ruegg2023kp2dtiny} and motion estimation \cite{buyuksolak2023ai}, respectively. As an example in the domain of spatial intelligence, R{\"u}egg et al. \cite{ruegg2023kp2dtiny} modify the neural network for keypoint detection, called KeyPointNet \cite{tang2019neural}, to make it more suitable for MCU deployment. This is achieved by creating two variants: one to decrease model size and another to improve latency. Onto the realm of positional intelligence, B{\"u}y{\"u}ksolak and G{\"u}ne{\c{s}}~\cite{buyuksolak2023ai} propose a neural network called Tiny Visual Odometry, TinyVO, for real-time, low-power visual odometry to either offer as an alternative to a Global Navigation Satellite System or an Inertial Navigation System (INS) or work in conjunction with them. 

\begin{table*}[htbp]
    \centering
    \caption{Selected papers using quantization for deployment on NPU-based MCUs.}
    \label{tab:quantized_npu_apps}

    \scriptsize
    \setlength{\tabcolsep}{3pt}
    \renewcommand{\arraystretch}{1.2}
    \begin{tabular}{lllp{1.6cm}lp{2.6cm}p{1.8cm}p{1.8cm}p{2cm}}
        \toprule
        \textbf{Study} & \textbf{Category} & \textbf{Quantization} & \textbf{Device(s)} & \textbf{Framework(s)} & \textbf{Performance} & \textbf{Power / Energy} & \textbf{Latency} & \textbf{Memory} \\
        \midrule
        \cite{balbi2024ultra} & Audio & INT8 PTQ & MAX78000 & \makecell[l]{PyTorch, \\ ai8x-tools} & 94.87\% Acc. & 5 $\mu$J & 104 $\mu$s & $<$432 KB \\
        \cite{liu2025research} & DSP & INT8 QAT & MAX78000 & \makecell[l]{PyTorch, \\ ai8x-tools} & 96.9--99.0\% Acc. & -- & 640--1600 $\mu$s & -- \\
        \cite{ingaleshwar2024wildlife} & Environment & INT8 PTQ & MAX78000 & \makecell[l]{PyTorch, \\ ai8x-tools} & 79.67--86.53\% F1 & 0.885--4.275 mJ & $\sim$4--27.5 ms & $\sim$100--460 KB \\
        \cite{van2023real} & Healthcare & INT8 QAT & MAX78002 & \makecell[l]{PyTorch, \\ ai8x-synthesis} & 94.6\% AUC & 18 mW & 248 $\mu$s & 25.156 KB \\
        \cite{gong2024dex} & Vision & INT8 QAT & \makecell[l]{MAX78000, \\ MAX78002} & \makecell[l]{PyTorch, \\ ai8x-tools} & 19.8--62.0\% Acc. & 0.14--0.4 mJ & 2--13 ms & 171.2--1330.7 KB \\
        \cite{lightbody2022host} & Anomaly Detection & INT8 QAT & MAX78000 & PyTorch & 87.19--99.95\% Acc. & 15 mW & 2556 $\mu$s & 55.9 KB \\
        \cite{tran2025facets} & Object Detection & INT8 QAT & MAX78000 & Once-for-All NAS & 88.1 mAP & 311 $\mu$J & 9.9 ms & 391 KB \\
        \cite{jakuvs2025implementing} & Speech & INT8 QAT & NXP MCXN94 & TFLite & 97.06\% Acc. & -- & 3847 $\mu$s & 30.6 KB \\
        \cite{ibrahim2024end} & Healthcare & INT8 QAT & Ethos-U55 & PyTorch & 94.25\% Acc. & 20 pJ & 5 ms & $<$32 KB \\
        \cite{moosmann2023tinyissimoyolo} & Object Detection & INT8 QAT & MAX78000 & \makecell[l]{PyTorch, \\ ai8x-tools} & 43.5--75.4\% mAP & \makecell[l]{0.19 mJ} & \makecell[l]{5.5 ms} & 350--422 KB \\
        \cite{ulkar2021ultra} & Speech & INT8 QAT & MAX78000 & \makecell[l]{PyTorch, \\ ai8x-tools} & 96.3\% Acc. & 251 $\mu$J & 3.5 ms & 419.8 KB \\
        \cite{dabbous2024benchmarking} & Environment & INT8 QAT & MAX78000 & \makecell[l]{TFLM, \\ PyTorch, \\ai8x-tools} & 99.63\% Acc. & 227.86 mW, 1.43 mW, 9 mW & 30.25 ms, 522.15 ms, 1.45 ms & -- \\
        \bottomrule
    \end{tabular}
\end{table*}

\subsection{Task-Specific Recommendations}
\label{sec:results-summary}

To summarize our applications section, we outline hardware-application mappings, quantization strategies, and optimization guidelines for different deployment scenarios.

\textbf{Hardware Selection.}  
When matching hardware to application requirements, three broad scenarios emerge. (1) RISC-V (e.g., GAP8/9) devices are best suited for reaction-critical, parallelizable perception loops---such as drones or real-time human activity recognition---where the combination of clustered cores, DMA, and tiling pipelines helps amortize memory traffic \cite{zhou2023solving, palossi2021fully}. ARM Cortex-M remains a strong choice for modest-depth models under moderate latency constraints, such as environmental sensing, where mature toolchains like TFLM or STM32Cube.AI enable rapid deployment with minimal engineering overhead \cite{kang2024device, dabbous2024benchmarking}. For higher-throughput or multi-model workloads, including multi-task vision, keyword spotting ensembles, or adaptive compression, NPUs and hybrid MCUs excel by leveraging accelerator dataflows to offset controller overhead \cite{balbi2024ultra, tran2025facets, van2023real}.  

\textbf{Quantization Strategy.}  
Quantization choices should be guided by profiling and model sensitivity. In most cases, uniform INT8---applied via PTQ or QAT---offers the best balance between performance and accuracy. Where profiling indicates sensitivity, selective use of FP16 for boundary layers or accumulators, or mixed sub-byte precision guided by per-layer signal-to-quantization-noise ratios, can yield improvements \cite{itani2025wireless, crupi2025efficient, misra2025latency}. Extreme low-bit formats, such as binary or ternary, should be reserved for situations where flash memory or bandwidth is the primary constraint. In such cases, distribution shaping and regularization are essential to mitigate accuracy degradation \cite{rutishauser2024xtern, sakr2022memory, zeng2022sub}.  

\textbf{Optimization and Energy Strategy.}  
Deployment decisions must also consider persistence and energy constraints. Toolchain co-design, using autotiling and memory-aware compilation frameworks such as Autotiler, ai8x-tools, or nntool, can externalize tiling and scheduling, freeing model designers to focus on architectural exploration \cite{zhou2023solving, balbi2024ultra, crupi2025efficient}. Energy strategy should be aligned with the application’s operational profile: for always-on tasks, prioritize minimizing energy per informative inference through bit-width tuning and sparsity-friendly kernels; for episodic clinical or safety-critical tasks, allocate resources to robustness measures such as regularization or adaptive precision, avoiding unnecessary aggressive compression \cite{itani2025wireless, busia2025endoscopy}.

\section{Challenges and Future Directions}
\label{sec:challenges}

As quantization continues to evolve, it is important to consider current challenges and potential future directions. This section provides an overview of areas that future researchers can explore to advance the field, breaking them down by sub-category of edge-friendly deployment, with a particular focus on hardware and software landscapes.

\subsection{Limited Sub-8-Bit and QAT Support}
Mainstream frameworks such as TensorFlow and PyTorch currently lack native support for quantization below 8 bits. As a result, researchers often turn to lesser-known toolchains (e.g., MAX-based frameworks) or custom implementations, which typically have limited documentation, non-standard workflows, and reduced reproducibility. This hinders adoption in adjacent applications and prevents building on standardized foundations. While 8-bit quantization remains the norm for MCU-class devices, more aggressive approaches see limited use primarily due to insufficient framework support.

Future work should extend popular frameworks to support sub-8-bit quantization, particularly for extreme edge devices, given that 4-bit and even binary models can be effective in many scenarios (see Section~\ref{sec:extreme-low-bit}). Integrating QAT at these low precisions would further standardize workflows for latency- and energy-sensitive applications. Community-driven initiatives could also define a compact, quantization-aware model exchange format (e.g., extending ONNX) that encodes precision, scale, and operation-level metadata to promote interoperability and reduce engineering overhead.

\subsection{Operator Support in Quantization Toolchains}
Current quantization frameworks often lack consistent operator coverage, leading to incompatibilities, especially when quantizing network parameters. For instance, TensorFlow’s QAT pipeline requires manual annotations and replacements for many layer operations, reducing the portability of QAT-tuned models and complicating deployment on resource-constrained devices. Addressing this gap will require a unified operator support standard across libraries, complemented by validation tools that ensure full ``quantizability" and provide safe replacements or workarounds where native support is absent.

\subsection{Static Memory Allocation Constraints and Mixed Precision}
Most embedded frameworks, such as TFLM~\cite{tensorflow}, rely on static memory allocation via a fixed-size tensor arena. While this simplifies management, it restricts the deployment of models with variable tensor shapes or mixed precision layers. Mixed precision quantization, which was shown to improve efficiency without compromising performance~\cite{mazumder2021, htq} (see Section~\ref{subsection:mixed-precision}), is rarely supported due to this limitation, as both hardware and toolchains are tied to fixed allocation schemes. Although recent work explores more flexible approaches~\cite{jeon2025tinymem}, adoption remains limited.

To address both the concerns regarding mixed precision quantization and static memory allocation, we recommend that future researchers work towards dynamic memory virtualization at the software level, and extend the support of MCU-class devices to work with mixed precision model parameters as well. This would be essential to support more complex models, flexible precision settings, and efficient multi-model deployment on memory-constrained hardware. 

\subsection{On-Device Training}
While inference on MCUs has grown into a practical standard, enabling on-device training remains a major challenge due to tight memory, energy, and compute restrictions. This limitation hinders applications that require continual learning, federated learning, or user-specific personalization. Several recent efforts have been made to address these challenges \cite{deutel2024device, kang2024device}, yet once again, widespread adoption is still not at a mature stage. 

To enable on-device training, frameworks must support quantized gradient descent, efficient memory reallocation, and selective fine-tuning strategies. Independently, each one of these components has existed within the field for extended periods of time, but future research can work on proposing robust frameworks for continual and federated learning directly on MCUs. This stretches its use beyond just quantization, but also to truly adaptive, privacy-preserving intelligence at the extreme edge.

\subsection{Extended Support for Emerging Architectures}

Modern architectures, such as transformers and diffusion models, are gaining traction at the edge but remain largely unsupported in current quantization frameworks due to their non-convolutional operations and complex memory patterns.

Recent work shows that these models can be adapted for MCU-class hardware: Deeploy~\cite{scherer2024deeploy} enables efficient deployment of quantized small language models on heterogeneous MCUs via compiler-level tiling and memory-aware scheduling; Zheng~\cite{zheng2025diffusion} address quantization challenges in diffusion models with optimizations for upsampling and denoising blocks; and domain-specific designs such as Tiny Transformer\cite{busia2024tiny} illustrate how structural modifications improve quantization compatibility for biomedical tasks.

To fully integrate such architectures into embedded AI workflows, quantization toolchains must extend their support to non-CNN models, including those with attention-aware operators, advanced memory scheduling, and mixed precision flexibility. Recent advances form a foundation, but broader integration is needed to close this gap.

\subsection{Support for Alternative Numerical Formats}

Conventional 8-bit and fixed-point formats limit flexibility across diverse edge workloads. Emerging formats like Posit offer improved dynamic range but lack mainstream hardware support. For example, POLARON \cite{lokhande2025polaron} introduces a trans-precision framework that adaptively switches between floating-point, fixed-point, and Posit representations. Its reconfigurable hardware and precision-aware controller enable runtime tuning based on workload sensitivity. Integrating trans-precision support into quantization toolchains and hardware platforms could enable more adaptive and efficient inference under tight resource constraints.

\subsection{Hardware-Aware Quantization Co-Design}
We continue to see quantization strategies proposed at a considerable scale. However, it should be observed that quantization strategies that ignore hardware constraints often yield insufficient performance within the deployment cycle. Especially for MCU-class devices, the co-optimization of quantization parameters with the underlying architecture would allow researchers to fully leverage the limited memory and compute resources. This calls for the emergence of better, robust hardware-aware quantization and NAS tailored to specific MCU families. Even with recent work done in this domain~\cite{palacios2025heepstor, fayyazi2025marco, king2025micronas}, we recommend further attention to the development of native hardware-aware optimization techniques.

\subsection{Sustainability and Environmental Impact}
Although quantization is typically motivated by efficiency (latency, memory, and energy), its implications increasingly intersect with sustainability as TinyML systems scale. At the device level, lower precision can reduce operational energy by decreasing compute cost and, often more importantly, memory traffic, which in turn can extend battery lifetime and enable always-on sensing without frequent charging or replacement. However, sustainability arguments remain difficult to make rigorously because the energy footprint of an embedded ML application is rarely dominated by neural inference alone. Pre- and post-processing, sensor acquisition, and communication can materially shift the energy profile, meaning that model-only efficiency claims may not translate to system-level sustainability benefits. To put it simply, an embedded ML pipeline often includes substantial non-neural-network-based costs which have to be carefully considered in terms of is overall environmental impact.

Recent work has begun to explicitly position compression and low-bit quantization explicitly through a sustainability lens~\cite{rashid2024tinym, rashid2025hac}. However, to make sustainability a defensible and reproducible future direction for quantization on microcontrollers, we recommend that future work (i) treat sustainability as a first-class objective in quantization by explicitly optimizing accuracy--latency--memory alongside energy and, where feasible, carbon-aware metrics, (ii) report both end-to-end and phase-separated energy/latency so that quantization gains are not overstated or masked by pipeline overheads~\cite{bartoli2025benchmarking}, (iii) evaluate system-level trade-offs jointly (compute versus communication, update frequency, and duty-cycle effects) rather than relying on model-only efficiency, and (iv) extend quantization toolchains to better support deployment regimes aligned with sustainability constraints, including lifetime-aware scheduling, energy-budgeted inference policies, and maintenance strategies that minimize costly re-training or frequent device-side updates.

\section{Conclusion}
\label{sec:conclusion}
This survey provided a hardware-oriented perspective on neural network quantization for MCU-class edge devices, synthesizing quantization methods, hardware platforms, software stacks, and deployment practices through the lens of practical MCU constraints. By systematically reviewing quantization techniques most relevant to MCUs and extreme-edge platforms, we highlighted the critical trade-offs between model accuracy, computational complexity, memory footprint, and numerical representations as a hardware design factor, alongside accelerator support and memory hierarchies. From a deployment standpoint, we examined the three major families of MCU platforms, namely ARM-based, RISC-V-based, and NPU-augmented designs, together with the software stacks actively used in the literature to enable TinyML. We further consolidated real-world applications and practitioner-oriented insights across domains such as audio, vision, healthcare, robotics, and environmental sensing, and identified persistent challenges that shape current and future MCU deployments, including the role of quantization as a potential enabler of sustainable AI when evaluated at the system level.

Several cross-cutting patterns tend to emerge. Firstly, INT8 quantization (both PTQ and QAT) remains the practical go-to for edge deployment. The most durable gains materialize through per-channel weight quantization, bias correction, cross-layer equalization, and quantization-aware finetuning. In sub-8-bit regions, we mainly see a gain under tight energy budgets, especially for MCU-based architectures that support extreme low bits. Moreover, we observe a shift in priority within different application domains. Healthcare and similar domains tend towards accuracy and performance, given the sensitivity of the field, whilst applications such as HAR and drones prioritize end-to-end latency and energy consumption. Reflecting these constraints and priorities, deployments are most consistent and reproducible on NPU-augmented MCUs (e.g., Ethos-U/STM32N6, MAX78000/78002, GAP8/9) when paired with their vendor toolchains, which constrain operator sets and tensor layouts, allocate static memory arenas, and target low-bit inference, making the latency, energy, and accuracy trade-offs more predictable.

From a software perspective, it is evident that frameworks are maturing, but still nascent. For example, TFLM’s static arenas, uneven operator coverage, and limited sub-8-bit support complicate portability and mixed precision execution. With the rise of PyTorch-based solutions (e.g., ExecuTorch), more PyTorch-based frameworks (e.g., ai8x-training/ai8x-synthesis) will become the norm. Moreover, other emerging solutions such as dynamic memory virtualization and exporter-compiler co-design point to a healthier path where accuracy, latency, energy, and memory are jointly optimized rather than being posed as trade-offs.

Looking ahead, the following key directions are pivotal to the advancement of quantization-based deployment. Firstly, the mainstream adoption of sub-8-bit quantization and mixed precision appears as the natural next step, as several works have highlighted the benefits of this particular direction. When coupled with quantization-aware strategies, we anticipate exponential advancements in the performance of quantized models once the hardware landscape readily supports them. Furthermore, enabling robust on-device training for continual or federated learning is on an upward trend, and its combination with online quantization strategies can also yield gains in energy and latency without sacrificing much performance.  Beyond model-level efficiency, future work should increasingly treat sustainability as a system-level objective by jointly optimizing accuracy, latency, memory, and energy, and by evaluating quantization benefits using end-to-end measurements. Addressing system-wide trade-offs, including compute, communication, update frequency, and device lifetime, as well as extending quantization toolchains to support energy-aware deployment policies, will be essential for realizing trustworthy, low-latency, and environmentally sustainable intelligence at the extreme edge.

\bibliographystyle{IEEEtran}  
\bibliography{references}

\begin{IEEEbiography}[{\includegraphics[width=1in,height=1.25in,clip,keepaspectratio]{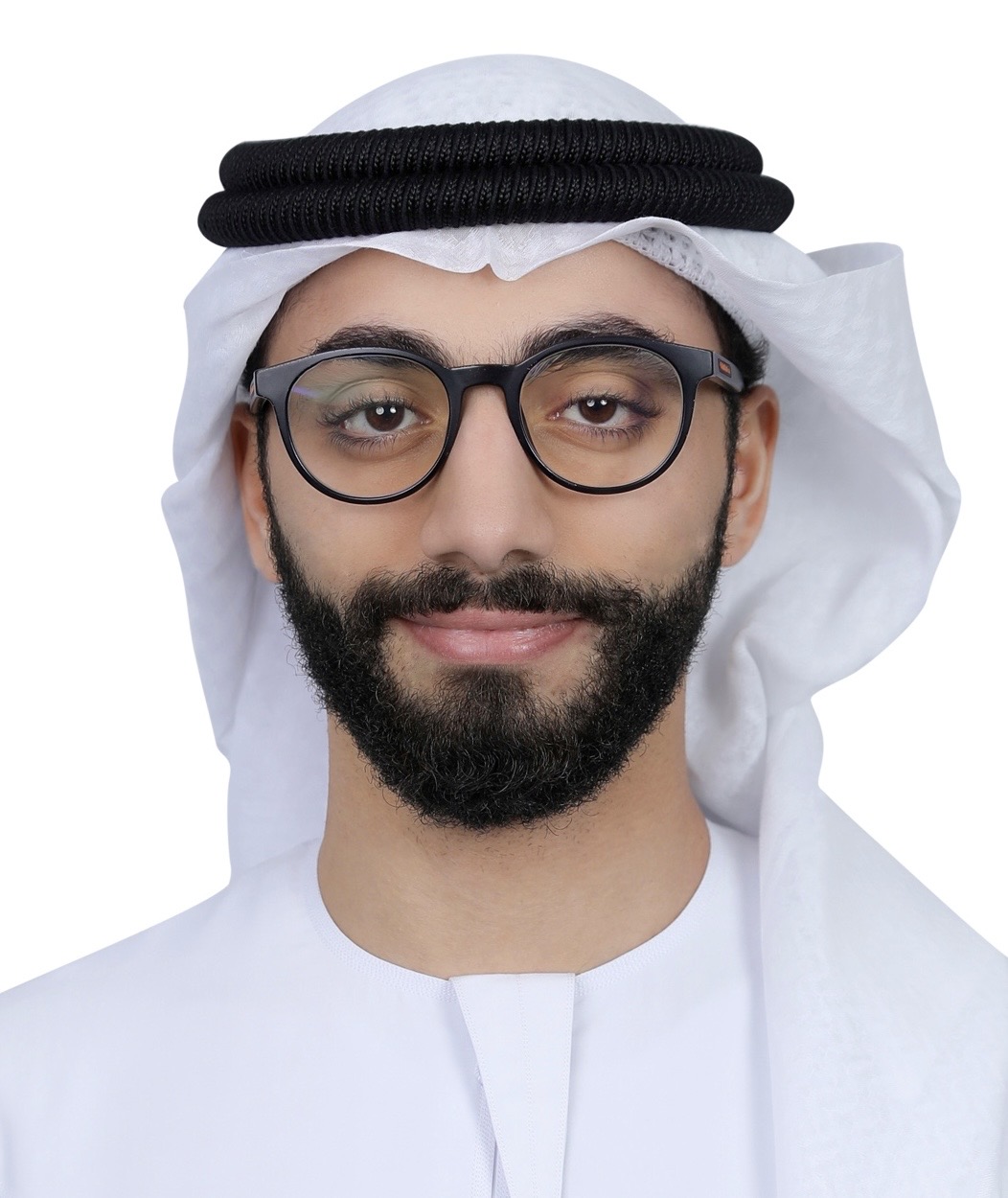}}]{Hamza A. Abushahla} (Graduate Student Member, IEEE) received the B.Sc. degree in Computer Engineering from the American University of Sharjah (AUS), U.A.E., in 2024, where he is currently pursuing the M.Sc. degree in Machine Learning. He is a Graduate Teaching and Research Assistant with the Department of Computer Science and Engineering at AUS. He is also a Visiting Researcher with the Advanced Research and Innovation Center (ARIC) at Khalifa University, Abu Dhabi, UAE. His research interests include quantization and compression of neural networks, microcontrollers, computer security, and robotics. He is a member of the IEEE Eta Kappa Nu Honors Society. 
\end{IEEEbiography}
\vspace{-1.5\baselineskip}

\begin{IEEEbiography}[{\includegraphics[width=1in,height=1.25in,clip,keepaspectratio]{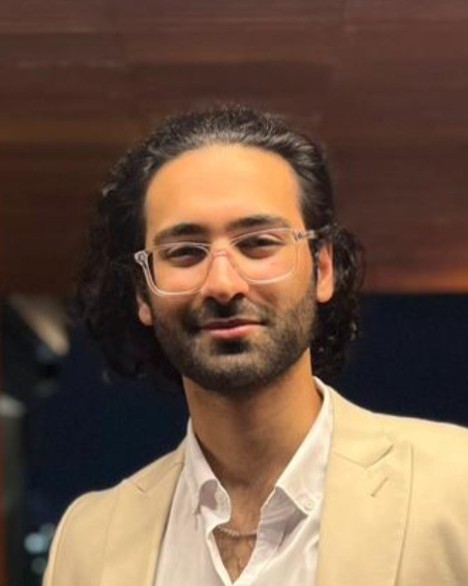}}]{Dara Varam} (Graduate Student Member, IEEE) received the dual B.Sc. in computer engineering and mathematics with a minor in data science from the American University of Sharjah, U.A.E, in 2023. He is currently pursuing the M.Sc. in Machine Learning at the same institution. He has been a research assistant with the Department of Computer Science and Engineering since 2021. Dara is also part of the Massachusetts Institute of Technology's Senseable City Laboratory as a visiting student. His research interests include deep learning, quantization and compression of neural networks, computer vision and optics. He is a member of the IEEE Eta Kappa Nu Honors Society and the Engineering Honors Society at the American University of Sharjah. 
\end{IEEEbiography}
\vspace{-1.5\baselineskip}

\begin{IEEEbiography}[{\includegraphics[width=1in,height=1.25in,clip,keepaspectratio]{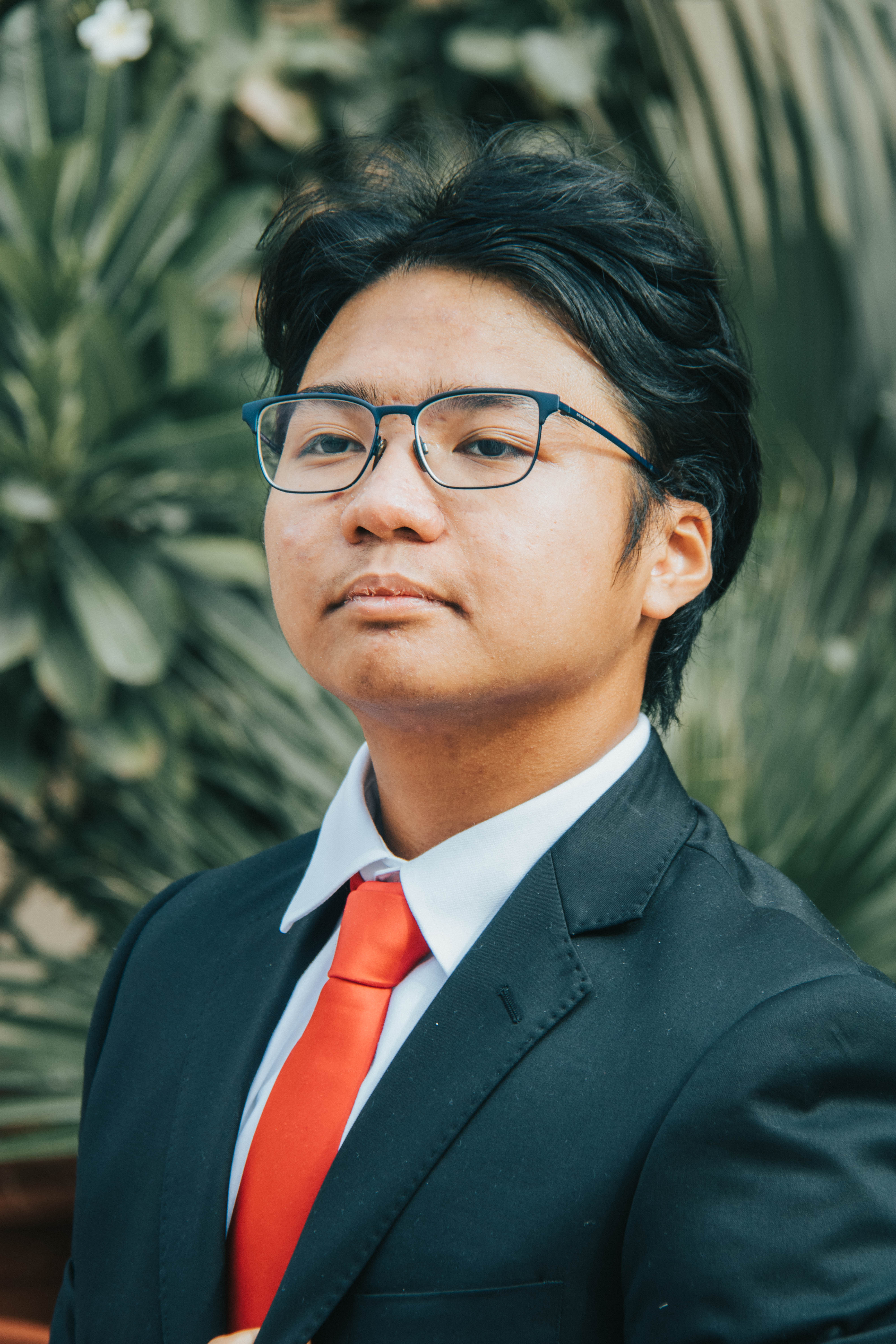}}]{Ariel J. N. Panopio} (Graduate Student Member, IEEE) received the B.Sc. degree in Computer Science with a minor in Data Science from the American University of Sharjah, United Arab Emirates, in 2024, where he is currently pursuing the M.Sc. degree in Machine Learning. He is also a Graduate Teaching and Research Assistant with the American University of Sharjah. His research interests include natural language processing, generative artificial intelligence, and computer security. He is a member of the IEEE Eta Kappa Nu Honors Society and the Tau Beta Pi Engineering Honor Society at the American University of Sharjah. 
\end{IEEEbiography}
\vspace{-1.5\baselineskip}

\begin{IEEEbiography}[{\includegraphics[width=1in,height=1.25in,clip,keepaspectratio]{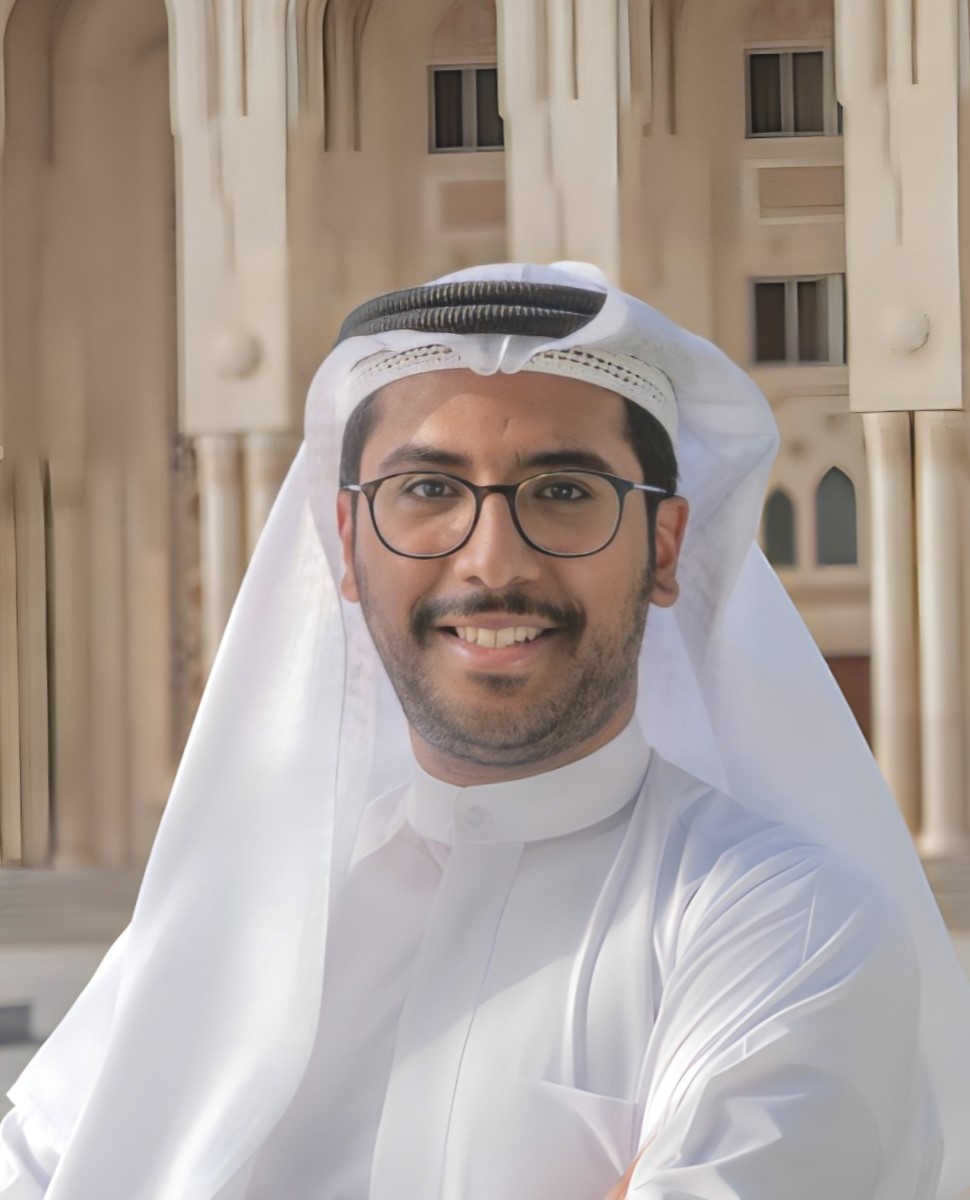}}]{Mohamed AlHajri} (Senior Member, IEEE)
received the B.Sc. degree (with Highest Hons.)
in electrical engineering from Khalifa University, UAE, in 2015, and the M.Sc. and Ph.D. degrees in electrical engineering and computer science, with a minor in applied mathematics from the Massachusetts Institute of Technology, Cambridge, MA, USA, in 2018 and 2023, respectively. He is currently an Assistant Professor of Computer Science and Engineering with the College of Engineering, American University of Sharjah, UAE. His research interests lie at the intersection of information theory, machine learning, and wireless communications, focusing on the analysis and design of low-complexity inference algorithms and real-time processing
of large datasets from complex systems. 
\end{IEEEbiography}

\end{document}